\documentclass[10pt,journal,compsoc]{IEEEtran}
\ifCLASSOPTIONcompsoc
\usepackage[nocompress]{cite}
\usepackage{amsmath,amsfonts}
\usepackage{algorithmicx}

\usepackage{algpseudocode}
\usepackage{array}
\usepackage{amsmath}
\usepackage{amsmath, bm}
\usepackage{threeparttable}
\usepackage{enumitem}
\usepackage[caption=false,font=normalsize,labelfont=sf,textfont=sf]{subfig}
\usepackage{booktabs}
\usepackage{tabularray}
\usepackage{bbding}
\usepackage{dsfont}
\usepackage{color} 

\usepackage[ruled,linesnumbered]{algorithm2e}
\usepackage[pagebackref=true,breaklinks=true,colorlinks=true,bookmarks=false,linkcolor=blue,anchorcolor=black,citecolor=blue, urlcolor=blue]{hyperref}

\usepackage[table]{xcolor}
\usepackage{textcomp}
\usepackage{pifont}
\usepackage{stfloats}
\usepackage{enumerate}
\usepackage{url}
\usepackage{verbatim}
\usepackage{graphicx}
\usepackage{cite}
\usepackage{bbm}
\usepackage{multirow}
\else
\usepackage{cite}
\fi
\ifCLASSINFOpdf
\else
\fi
\begin{document}
%





\title{RingMoE: Mixture-of-Modality-Experts Multi-Modal Foundation Models for Universal Remote Sensing Image Interpretation}
%
%
%
%

\author{

Hanbo~Bi,~
Yingchao~Feng,
Boyuan~Tong,
Mengyu~Wang,
Haichen~Yu,
Yongqiang~Mao,
Hao~Chang,
Wenhui~Diao,
Peijin~Wang,
Yue~Yu,~\IEEEmembership{Member,~IEEE,}
Hanyang~Peng,
Yehong~Zhang,
Kun~Fu,~\IEEEmembership{Member,~IEEE,}
and~Xian~Sun,~\IEEEmembership{Senior Member,~IEEE}


\IEEEcompsocitemizethanks{
\IEEEcompsocthanksitem H. Bi, B. T, M. W, H. Y, H. C, K. Fu, and X. Sun are with the Aerospace Information Research Institute, Chinese Academy of Sciences, Beijing 100190, China, also with the School of Electronic, Electrical and Communication Engineering, University of Chinese Academy of Sciences, Beijing 100190, China, also with the University of Chinese Academy of Sciences, Beijing 100190, China, and also with the Key Laboratory of Target Cognition and Application Technology(TCAT), Aerospace Information Research Institute, Chinese Academy of Sciences, Beijing 100094, China.
\IEEEcompsocthanksitem Y. Feng, W. Diao, and P. Wang are with the Aerospace Information Research Institute, Chinese Academy of Sciences, Beijing 100190, China, and also with the Key Laboratory of Target Cognition and Application Technology(TCAT), Aerospace Information Research Institute, Chinese Academy of Sciences, Beijing 100094, China.
\IEEEcompsocthanksitem Y. Mao is with the Department of Electronic Engineering, Tsinghua University, Beijing 100084, China.
\IEEEcompsocthanksitem Y. Yu, H. Peng, and Y. Zhang are with the Peng Cheng Laboratory, Shenzhen 518066, China.
\IEEEcompsocthanksitem Corresponding authors: Y. Feng (e-mail: fengyc@aircas.ac.cn).

}
}

%
%

\markboth{Journal of \LaTeX\ Class Files,~Vol.~14, No.~8, August~2015}%
{Shell \MakeLowercase{\textit{et al.}}: Bare Demo of IEEEtran.cls for Computer Society Journals}
%



\IEEEtitleabstractindextext{%
\begin{abstract}
The rapid advancement of foundation models has revolutionized visual representation learning in a self-supervised manner. However, their application in remote sensing (RS) remains constrained by a fundamental gap: existing models predominantly handle single or limited modalities, overlooking the inherently multi-modal nature of RS observations. Optical, synthetic aperture radar (SAR), and multi-spectral data offer complementary insights that significantly reduce the inherent ambiguity and uncertainty in single-source analysis. To bridge this gap, we introduce RingMoE, a unified multi-modal RS foundation model with 14.7 billion parameters, pre-trained on 400 million multi-modal RS images from nine satellites. RingMoE incorporates three key innovations: (1) A hierarchical Mixture-of-Experts (MoE) architecture comprising modal-specialized, collaborative, and shared experts, effectively modeling intra-modal knowledge while capturing cross-modal dependencies to mitigate conflicts between modal representations; (2) Physics-informed self-supervised learning, explicitly embedding sensor-specific radiometric characteristics into the pre-training objectives; (3) Dynamic expert pruning, enabling adaptive model compression from 14.7B to 1B parameters while maintaining performance, facilitating efficient deployment in Earth observation applications. Evaluated across 23 benchmarks spanning six key RS tasks (i.e., classification, detection, segmentation, tracking, change detection, and depth estimation), RingMoE outperforms existing foundation models and sets new SOTAs, demonstrating remarkable adaptability from single-modal to multi-modal scenarios. Beyond theoretical progress, it has been deployed and trialed in multiple sectors, including emergency response, land management, marine sciences, and urban planning.

\end{abstract}

\begin{IEEEkeywords}
Foundation Model, Self-supervised Learning, Mixture-of-Expert, Multi-Modal, Remote Sensing
\end{IEEEkeywords}}

\maketitle

\IEEEdisplaynontitleabstractindextext

%
\IEEEpeerreviewmaketitle

\IEEEraisesectionheading{\section{Introduction}\label{sec:introduction}}

\begin{figure}[t]
\setlength{\abovecaptionskip}{0pt}
\setlength{\belowcaptionskip}{0pt}
\centering
\includegraphics[width=1.0\linewidth]{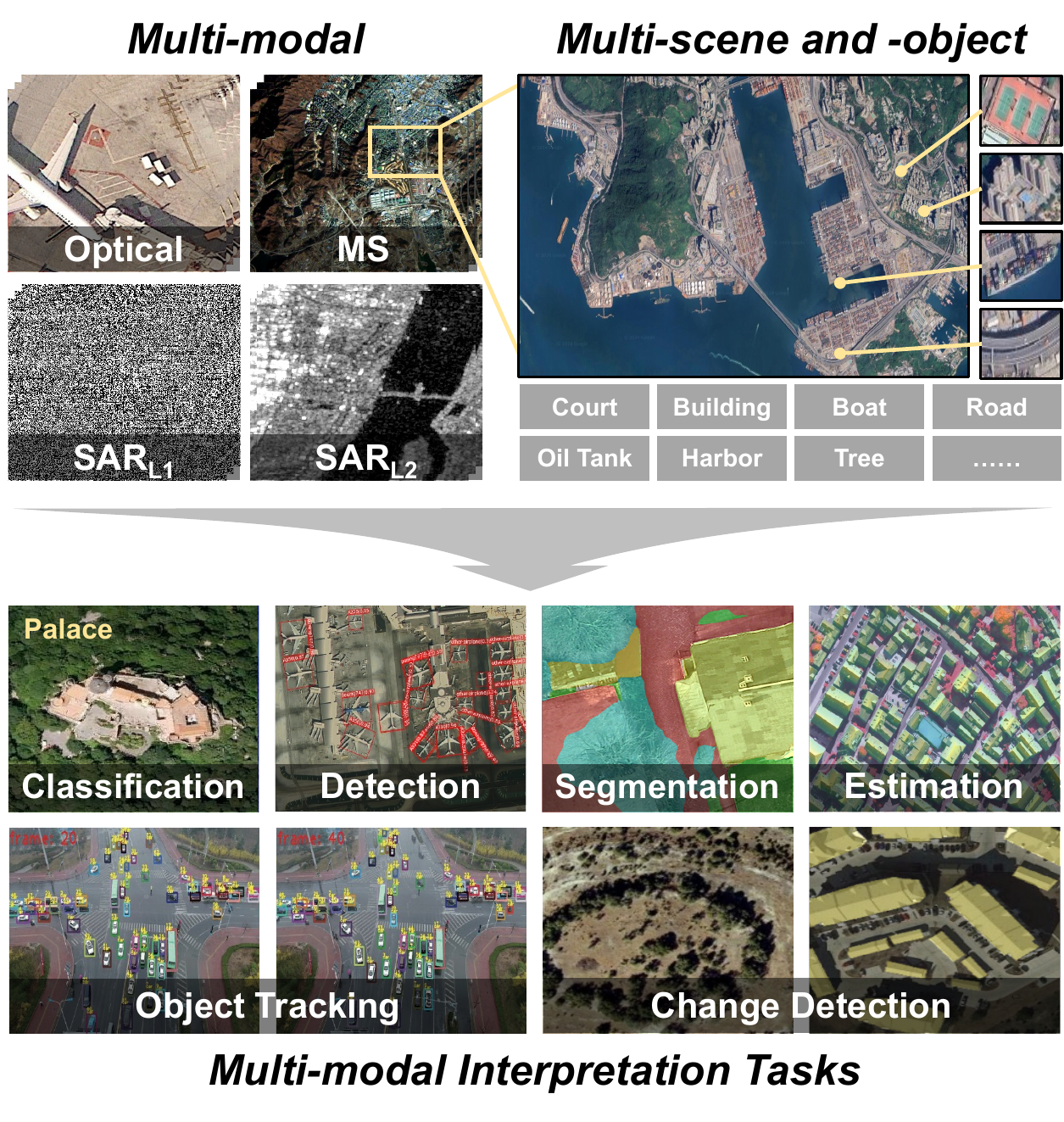}
\caption{\textbf{The motivation for developing our multi-modal RSFM, i.e., RingMoE,} is to adaptively process various image interpretation tasks from different RS modalities including optical, multi-spectral, and SAR (in complex-valued form and amplitude form).}
\label{fig:1}
\end{figure}

\IEEEPARstart{T}{he} field of remote sensing (RS) has revolutionized our ability to observe and understand Earth's surface~\cite{sun2023revealing,koppa2022deep,cavender2022integrating}, leveraging satellite systems with global coverage and advanced imaging technologies. Unlike conventional photography, RS generates diverse data types, such as optical (Opt), multi-spectral (MS), and synthetic aperture radar (SAR), each offering unique and complementary insights. Opt imagery captures Earth's reflectance in visible bands, akin to standard photographs but constrained by lighting and weather conditions. MS data extends into infrared and ultraviolet bands for material identification and environmental monitoring, while SAR utilizes microwave signals to provide all-weather, day-and-night terrain information. Despite the abundance of data generated daily by over 10,000 active satellites~\cite{spacedaily2024}, much remains underutilized due to high annotation costs and the lack of robust analysis frameworks, highlighting the need for innovative solutions to harness the full potential of multi-modal RS data.


Foundation models have recently emerged as transformative in AI, leveraging self-supervised learning~\cite{he2020momentum} to extract generalized, task-agnostic representations from large-scale unlabeled data. Compared to smaller, task-specific models tailored for individual scenarios, these ``one-for-all" general foundation models demonstrate superior performance and generalization across diverse interpretation tasks, particularly in data-scarce environments. Notable examples such as ChatGPT~\cite{chen2023minigpt} in NLP, SiMIM~\cite{xie2022simmim} and MAE~\cite{he2022masked} in CV, and emerging models~\cite{zhou2023foundation,moor2023foundation, wu2023interpretable} in specialized fields like medical imaging and meteorology have redefined benchmarks in their respective domains. 

Inspired by these breakthroughs, remote sensing foundation models (RSFMs) have gained momentum (see Fig.\ref{fig:1}). Community researchers have explored generic representations based on unlabeled RS data employing masked image modeling (MIM) \cite{xie2022simmim,peng2022unified} and contrastive learning (CL) \cite{chen2020simple,wang2021dense} paradigms. Early works like RingMo~\cite{sun2022ringmo} propose employing the MIM paradigm to capture optical representations. Scale-MAE \cite{reed2023scale} and Billion-scale MAE \cite{cha2024billion} have explored expanding model parameters to improve generalization. In addition, Hong et al. \cite{hong2024spectralgpt} propose SpectralGPT, specifically designed for processing spectral RS images. 

We summarize the success of the existing RSFMs and consider that the following two key issues can be further explored and optimized. On this basis, we built a more intelligent remote sensing interpretation model (see Fig.\ref{fig:1}).

\begin{table}[t]
\setlength{\abovecaptionskip}{0pt}
\setlength{\belowcaptionskip}{0pt}
\caption{Supported RS modalities in different RSFMs. While most existing RSFMs leverage only one or two modalities, our model seeks to achieve a more comprehensive multi-modal understanding.} \label{tab:1}
\renewcommand\arraystretch{1.2}
\centering
\resizebox{0.9\linewidth}{!}{
\begin{tabular}{r|cccc} 
\toprule
\multirow{2}{*}{Foundation model} & \multicolumn{4}{c}{Supported RS modalities}  \\ 
\cline{2-5}
            & Opt & MS & SAR-L1 & SAR-L2               \\ 
\hline
RingMo (2022) \cite{sun2022ringmo}                 & \ding{51}   &\textcolor[rgb]{0.703,0.703,0.703}{\ding{55}}    & \textcolor[rgb]{0.703,0.703,0.703}{\ding{55}}       & \textcolor[rgb]{0.703,0.703,0.703}{\ding{55}}                     \\
SatMAE (2022) \cite{cong2022satmae}                 & \ding{51}   & \ding{51}  & \textcolor[rgb]{0.703,0.703,0.703}{\ding{55}}       & \textcolor[rgb]{0.703,0.703,0.703}{\ding{55}}                     \\
Scale-MAE (2023) \cite{reed2023scale}              & \ding{51}   & \textcolor[rgb]{0.703,0.703,0.703}{\ding{55}}   & \textcolor[rgb]{0.703,0.703,0.703}{\ding{55}}       & \textcolor[rgb]{0.703,0.703,0.703}{\ding{55}}                     \\
SpectralGPT (2024) \cite{hong2024spectralgpt}            & \textcolor[rgb]{0.703,0.703,0.703}{\ding{55}}    & \ding{51}  & \textcolor[rgb]{0.703,0.703,0.703}{\ding{55}}       & \textcolor[rgb]{0.703,0.703,0.703}{\ding{55}}                     \\
SARATR-X (2024) \cite{yang2024saratr}            & \textcolor[rgb]{0.703,0.703,0.703}{\ding{55}}    & \textcolor[rgb]{0.703,0.703,0.703}{\ding{55}}  & \textcolor[rgb]{0.703,0.703,0.703}{\ding{55}}       & \ding{51}                    \\

DiffusionSat (2024)\cite{khanna2023diffusionsat}      & \ding{51}   & \textcolor[rgb]{0.703,0.703,0.703}{\ding{55}}   & \textcolor[rgb]{0.703,0.703,0.703}{\ding{55}}       &  \textcolor[rgb]{0.703,0.703,0.703}{\ding{55}}    \\ 
MM Contrastive (2022)\cite{jain2022multimodal}         & \ding{51}   & \textcolor[rgb]{0.703,0.703,0.703}{\ding{55}}   & \textcolor[rgb]{0.703,0.703,0.703}{\ding{55}}       & \ding{51}                    \\
DINO-MM (2022)\cite{wang2022self}                & \ding{51}   & \textcolor[rgb]{0.703,0.703,0.703}{\ding{55}}   & \textcolor[rgb]{0.703,0.703,0.703}{\ding{55}}       & \ding{51}                    \\
FG-MAE (2023)\cite{wang2023feature}                & \textcolor[rgb]{0.703,0.703,0.703}{\ding{55}}   & \ding{51}   & \textcolor[rgb]{0.703,0.703,0.703}{\ding{55}}       & \ding{51}                    \\
CROMA (2024)\cite{fuller2024croma}                  & \ding{51}   & \textcolor[rgb]{0.703,0.703,0.703}{\ding{55}}   & \textcolor[rgb]{0.703,0.703,0.703}{\ding{55}}       & \ding{51}                    \\
SkySense (2024)\cite{guo2024skysense}               & \ding{51}   & \ding{51}  &  \textcolor[rgb]{0.703,0.703,0.703}{\ding{55}}      & \ding{51}                    \\ 
\hline
\textbf{Our RingMoE}     & \ding{51}   & \ding{51}  & \ding{51}      & \ding{51}                    \\
\bottomrule
\end{tabular}
}
\end{table}

\begin{figure}[t]
\setlength{\abovecaptionskip}{0pt}
\setlength{\belowcaptionskip}{0pt}
\centering
\includegraphics[width=1.0\linewidth]{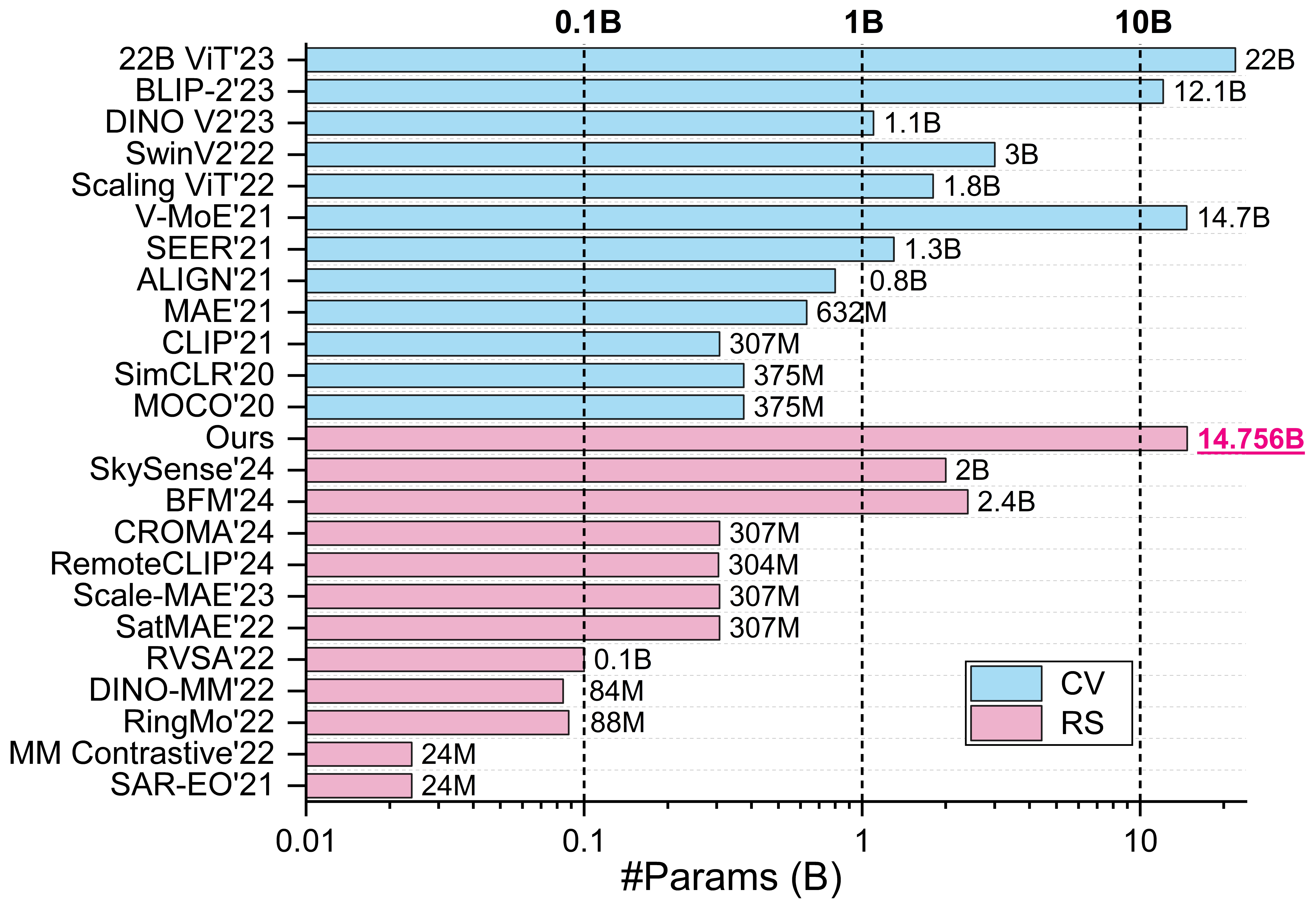}
\caption{\textbf{Parameter-scale of foundation models in CV and RS fields} \cite{chen2020simple,radford2021learning,goyal2021self,jia2021scaling,zhai2022scaling,dehghani2023scaling,liu2022swin,riquelme2021scaling,oquab2023dinov2,li2023blip,he2020momentum,guo2024skysense,fuller2024croma,liu2024remoteclip,cha2024billion,reed2023scale,cong2022satmae,wang2022advancing,wang2022self,sun2022ringmo,jain2022multimodal,cha2021contrastive}. Our RingMoE is the largest in RS and ranks among the top in CV.}
\label{fig:2}
\end{figure}

\noindent \textbf{(1) Unified Multi-modal Perception.}
Building a complete Earth observation system usually requires multiple sensors (e.g., Opt, MS, and SAR) working in tandem to capture Earth's surface properties. However, most existing RSFMs only serve single or narrow modalities (see Tab.\ref{tab:1}), e.g. RingMo~\cite{sun2022ringmo} focus on Opt and DINO-MM~\cite{cha2021contrastive} on Opt and SAR. Indeed, such modality-specific tailored models hinder the enormous potential of exploring RS multi-modal data. Numerous works have revealed that although heterogeneous sensors have different imaging modes (active imaging for SAR and passive imaging for Opt/MS) and non-uniform technological regimes (high-resolution imaging, wide-area observation, and dynamic monitoring, etc), {the information from different modalities can complement each other~\cite{dalla2015challenges}, e.g., the fusion of Opt and SAR can improve the imaging quality during nighttime and bad weather conditions~\cite{zhang2024optical}.} At this point, combining these modalities to construct a unified RS cross-modal representation can fully unleash the potential of multi-modal data and facilitate a better understanding of RS images. 

{Meanwhile, most existing SAR-based RSFMs utilize SAR amplitude data (namely SAR-L2 in this paper) as a visual proxy, treating it similarly to grayscale images for contrastive or masked image modeling~\cite{fuller2024croma,yang2024saratr,wang2022self}. While this simplification facilitates model design, it discards the phase component of the SAR signal, resulting in the loss of structural and scattering clues essential to understanding radar imaging physics~\cite{huang2022physically}. In contrast, complex-valued SAR-L1 data retains both amplitude and phase, enabling richer physical interpretations of the scene. Recent studies~\cite{huang2022physically,zeng2022sar,asiyabi2023complex} demonstrate that incorporating phase information helps capture physically grounded clues, such as surface roughness, dielectric properties, and object geometry, that are inaccessible via amplitude alone. Fully exploiting SAR-L1's unique characteristics is therefore crucial for building RSFMs that move beyond visual approximations to a more faithful physical understanding of radar data.}


In summary, the ideal RSFM should uniformly process RS heterogeneous data from various imaging mechanisms and spectral bands, effectively capturing both inter-modal and intra-modal representations for application across diverse Earth observation tasks.

\noindent \textbf{(2) Larger-scale Foundation Model.}
Experience in the deep learning field has demonstrated that increasing the capacity and computational scale of the network typically improves performance. {Larger models pre-trained on large datasets tend to reach the state-of-the-art in CV \cite{zhai2022scaling,dehghani2023scaling,liu2022swin}, and such a paradigm has achieved an even greater response in NLP \cite{brown2020language,dai2024deepseekmoe}.} In the RS field, research on RSFMs has predominantly focused on pre-training strategies \cite{sun2022ringmo,jain2022multimodal} and dataset scale \cite{wang2022self,cha2021contrastive}, while relatively little attention has been given to parameter scaling \cite{cha2024billion}. As illustrated in Fig.\ref{fig:2}, 
the parameter scale of the foundation models in RS remains relatively small compared to those in CV, with most models containing fewer than 100M (0.1B) parameters. However, {recent advancements, such as Billion-scale MAE~\cite{cha2024billion} and SkySense~\cite{guo2024skysense}, demonstrate that increasing model parameters can significantly boost performance.} Given the success of large-scale foundation models (i.e., 22B ViT~\cite{dehghani2023scaling}) in CV, developing an RSFM with a larger parameter scale is crucial for improving generalization across diverse imaging conditions, including variations in time periods, scenes, resolutions, and sensor types. Admittedly, training and deploying larger-scale foundation models is time-consuming and resource-intensive. Therefore, it is worth exploring and considering how to build larger-scale foundation models that balance computational cost with deployment efficiency while ensuring high performance.

Based on the exploration of the above two issues, in this paper, we augment our work RingMo \cite{sun2022ringmo} and propose \textbf{RingMoE}, a multi-modal RSFM with \textbf{14.7 billion} parameters, which is the largest foundation model in RS to date. In contrast to previous RSFMs, RingMoE is pre-trained on a much large-scale and comprehensive dataset, comprising 400 million images from four modalities: Opt, MS\footnote[1]{Multi-spectral data includes infrared band data in this paper.}, SAR-L1, and SAR-L2. This dataset provides extensive global coverage and captures diverse geospatial features. Notably, we introduce SAR-L1 data with amplitude and phase to enable the model to better understand and interpret the scattering properties of surface objects. The key innovations of RingMoE are summarized as follows:

\begin{figure*}[t]
\setlength{\abovecaptionskip}{0pt}
\setlength{\belowcaptionskip}{0pt}
\centering
\includegraphics[width=0.9\linewidth]{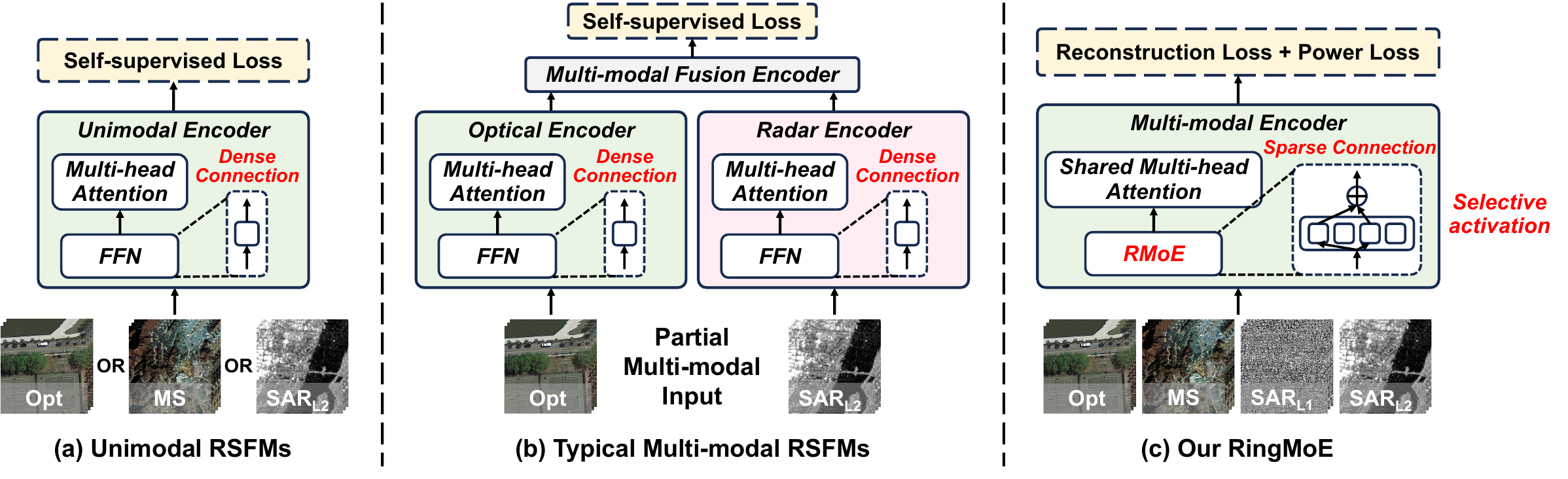}
\caption{\textbf{Structure comparison between the previous RSFMs and the proposed RingMoE}. \textbf{(a) Unimodal RSFMs} \cite{cha2024billion,reed2023scale,cong2022satmae,wang2022advancing,sun2022ringmo,wang2024scaling}: Given a certain unimodal input, the latent representations are extracted by the unimodal encoder, followed by decoding by contrast supervision or target reconstruction. \textbf{(b) Typical Multi-modal RSFMs} \cite{cha2021contrastive,jain2022multimodal,wang2022self,guo2024skysense,fuller2024croma}: take SAR-EO \cite{cha2021contrastive} as an example, these models process two modalities separately through their respective encoders, followed by inter-modal interaction via a multi-modal fusion encoder, with decoding performed through contrastive supervision or target reconstruction. \textbf{(c) Our RingMoE}: Given four modal inputs, RingMoE employs a multi-modal encoder with a sparse RMoE structure that selectively activates different experts for each modality, capturing both inter- and intra-modal correlations. Additionally, modal-specific decoders are introduced for self-supervised learning, incorporating a power loss function to embed radar-specific imaging characteristics.}
\label{fig:3}
\end{figure*}

\noindent \textbf{(1) Joint Encoder across Modalities and Experts for Multi-Modal Representation.} For effectively capturing stable and discriminative multi-modal representations, RingMoE employs a hierarchical Mixture-of-Experts (MoE) architecture, balancing scalability with computational efficiency (see Fig.\ref{fig:3}). Conventional MoE designs~\cite{dai2024deepseekmoe,chen2023mod,fedus2022switch} share a common set of experts across all modalities, which can lead to knowledge entanglement and conflicts, especially between inherently different modalities like Opt and SAR-L1. To address this, RingMoE introduces a structured RMoE framework, comprising three specialized expert types: \textbf{Modal-Specialized experts}, capturing fine-grained intra-modal representations; \textbf{Collaborative experts}, modeling inter-modal correlations; and a \textbf{Shared expert}, distilling common knowledge across modalities to reduce parameter redundancy. By strategically partitioning expert roles, RingMoE effectively captures both intra-modal expertise and cross-modal dependencies, ensuring robust and efficient multi-modal representation learning.

\noindent \textbf{(2) Modal-Specific Decoders for Physics-Informed Self-Supervised Learning.} As a multi-modal extension of MIM~\cite{xie2022simmim,peng2022unified}, RingMoE employs modal-specific decoders to reconstruct the original targets from latent representations, embedding sensor-specific physical properties into the self-supervised learning process. For Opt, MS, and SAR-L2 images, the model reconstructs original pixel values, preserving spatial and spectral integrity. For SAR-L1 images, given their complex-valued nature, we enforce a physics-informed reconstruction strategy based on the principle of power conservation before and after polarization decomposition\cite{yamaguchi2005four}, choosing to reconstruct image power instead. This physically grounded learning mechanism enhances the model's ability to capture polarimetric SAR characteristics, improving its understanding of RS modalities.

\noindent \textbf{(3) Dynamic Expert Pruning for Efficient Deployment across Various Computational Resources.} Leveraging the modular nature of the proposed expert design, RingMoE can be flexibly decomposed into multiple modality-specific (unimodal) models for downstream tasks. Instead of retaining modal-specialized experts across all modalities, each unimodal model selectively preserves only its respective modal-specialized, collaborative, and shared experts, ensuring computational efficiency while maintaining performance. Furthermore, through targeted expert pruning strategies, we provide multiple sparse and dense lightweight versions (e.g., 1B parameters), enabling adaptable deployment across diverse computational environments without compromising effectiveness in Earth observation applications.

\begin{figure}[t]
\setlength{\abovecaptionskip}{0pt}
\setlength{\belowcaptionskip}{0pt}
\centering
\includegraphics[width=1.0\linewidth]{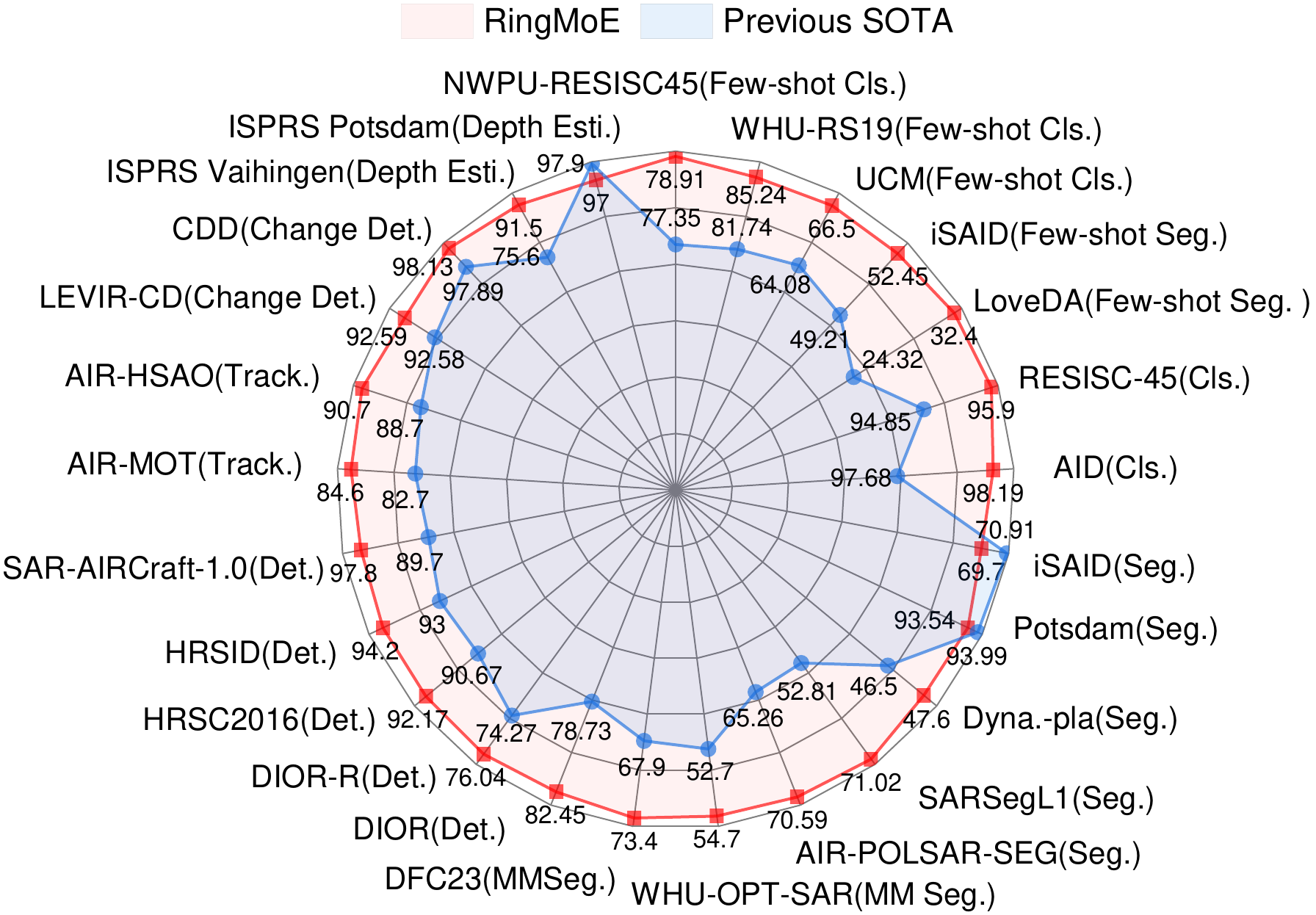}
\caption{\textbf{The proposed RingMoE achieves 23 SOTAs on 25 benchmarks in 6 RS key tasks}, outperforming existing foundation models.}
\label{fig:4}
\end{figure}

The model's performance is rigorously evaluated across 25 publicly available benchmarks encompassing six key RS tasks: classification, detection, segmentation, tracking, change detection, and depth estimation, spanning various modalities. Experimental results demonstrate that RingMoE achieves SOTA performance on 23 benchmarks, e.g., surpassing SkySense~\cite{guo2024skysense} by 3.72\% and Scale-MAE~\cite{reed2023scale} by 8.64\% on the widely used DIOR detection task (see Fig.\ref{fig:4}). These results establish its superior interpretability in multi-modal RS and its strong few-shot adaptation ability, underscoring its potential for advancing RS research and enabling real-world applications in complex multi-modal environments.

In summary, the contributions of this paper can be summarised as follows:
\begin{itemize}
\item We propose RingMoE, with 14.7 billion parameters, the largest multi-modal RSFM to date, can handle diverse interpretation tasks from Optical, Multi-spectral, and SAR (in complex-valued and amplitude form). 
\item We introduce the sparse Mixture-of-Experts architecture into RSFM for the first time, designing modal-specialized, collaborative, and shared experts to efficiently structure multi-modal learning and mitigate modality conflicts.

\item RingMoE embeds sensor-specific physical characteristics into the self-supervised pre-training, enhancing interpretability for remote sensing tasks.

\item We develop adaptive expert pruning strategies that enable RingMoE to be efficiently deployed across varying computational constraints. Extensive experiments on 25 datasets from 6 tasks across different modalities demonstrate that the proposed RingMoE performs excellently even when pruned to 1B, yielding 23 SOTA results.
\end{itemize}

\section{Related Work}
\subsection{Remote Sensing Foundation Models (RSFMs)}
Self-supervised learning \cite{he2020momentum,he2022masked,xie2022simmim} in the CV field has significantly contributed to the flourishing of RSFMs~\cite{guo2024skysense,fuller2024croma,liu2024remoteclip,cha2024billion,reed2023scale,cong2022satmae,wang2022advancing,wang2022self,sun2022ringmo,jain2022multimodal,cha2021contrastive}, which can leverage the vast amount of unlabeled RS images to efficiently capture the generic feature representation. Mainstream RSFMs mainly employ masked image modeling (MIM) and contrastive learning (CL) for pre-training. (1) MIM-based RSFM, which learns image representations by reconstructing masked patches. RingMo~\cite{sun2022ringmo} optimizes the masking strategy in MIM to focus on dense and small objects. SpectralGPT~\cite{hong2024spectralgpt} designs a 3D tensor-shaped spatial-spectral mask strategy to comprehensively capture locally spatial-spectral features and spectral sequential information. Scale-MAE~\cite{reed2023scale} reconstructs low-frequency and high-frequency information of the masked image at different scales to improve multi-scale perception. (2) CL-based RSFM, which optimizes the model by closing the distance between positive samples and widening the distance between negative samples. GASSL \cite{ayush2021geography} takes time-series views as positive sample pairs to neglect temporal variation and focus on more spatial variation. SeCo \cite{manas2021seasonal} constructs seasonal-based multi-augmentation contrast loss to perceive short- and long-term temporal changes. 


\subsection{Multi-modal Remote Sensing Foundation Models}
The rapid expansion of RS data has heightened the demand for intelligent models capable of handling multi-modal scenarios. Existing uni-modal RSFMs struggle with cross-modal generalization, while developing separate models for each modality incurs significant computational and storage costs. Consequently, constructing multi-modal RSFMs has emerged as a key research focus. Most current multi-modal RSFMs~\cite{cha2021contrastive,jain2022multimodal,wang2022self,guo2024skysense,fuller2024croma} rely on contrastive learning. DINO-MM~\cite{wang2022self} and MM Contrastive~\cite{jain2022multimodal} align features across SAR and optical modalities, while CROMA~\cite{fuller2024croma} integrates contrastive learning with mask reconstruction to enhance cross- and intra-modal feature capture. SkySense~\cite{guo2024skysense} further employs multi-granularity contrastive learning to improve robustness across optical, multi-spectral, and SAR data. However, these methods typically require strictly paired multi-modal data, limiting scalability and generalization to unpaired scenarios. To address this, OFA-Net~\cite{xiong2024one} introduces a MIM mechanism for learning without paired data. While this improves scalability, simple parameter sharing may lead to information loss or interference between modalities. Therefore, developing a more flexible multi-modal RSFM capable of learning both shared and modality-specific features remains a crucial challenge.

\subsection{Mixture-of-Experts (MoE)}
Experience with deep learning suggests that increasing model capacity and dataset scale tends to improve performance \cite{brown2020language}. However, it must be acknowledged that scaling models to mega-scale also requires extremely high computational costs. Mixture-of-Experts (MoE) architecture \cite{zhou2022mixture,dai2024deepseekmoe,chen2023mod,fedus2022switch} is a promising solution to keep the computational cost acceptable while scaling up the model. With the sparsity of the MoE, the NLP community has expanded language models to a considerable scale \cite{fedus2022switch,dai2024deepseekmoe,zhou2022mixture,zhang2022mixture}. Switch Transformer \cite{fedus2022switch} optimizes the MoE routing pattern and builds a model with up to a trillion parameters. 
MoA \cite{zhang2022mixture} proposes a new architecture that combines multiple attention and the MoE technique. DeepseekMoE \cite{dai2024deepseekmoe} constructs fine-grained expert segmentation and shared expert isolation to increase expert specialization and model parameter efficiency (see Appendix A for details on typical MoE architecture). In addition, the researchers have also applied the MoE technique to the CV field \cite{riquelme2021scaling,zhou2022mixture,fan2022m3vit,chen2023mod}, e.g, V-MoE first applies the MoE to the CV field and proposes a 15B parameter model, which shows the potential of scale vision models. $\mathbf{M}^3$ViT \cite{fan2022m3vit} and Mod-Squad \cite{chen2023mod} also design efficient multi-task learning models with the MoE technology. In this paper, we apply the MoE technique to RS and construct a 14.7B parameter RSFM. Notably, for maximizing the parameter potential, we propose modal-specialized, collaborative, and shared experts to rationally divide up the learning of RS multi-modal knowledge.

\begin{figure*}[t]
\setlength{\abovecaptionskip}{0pt}
\setlength{\belowcaptionskip}{0pt}
\centering
\includegraphics[width=0.95\linewidth]{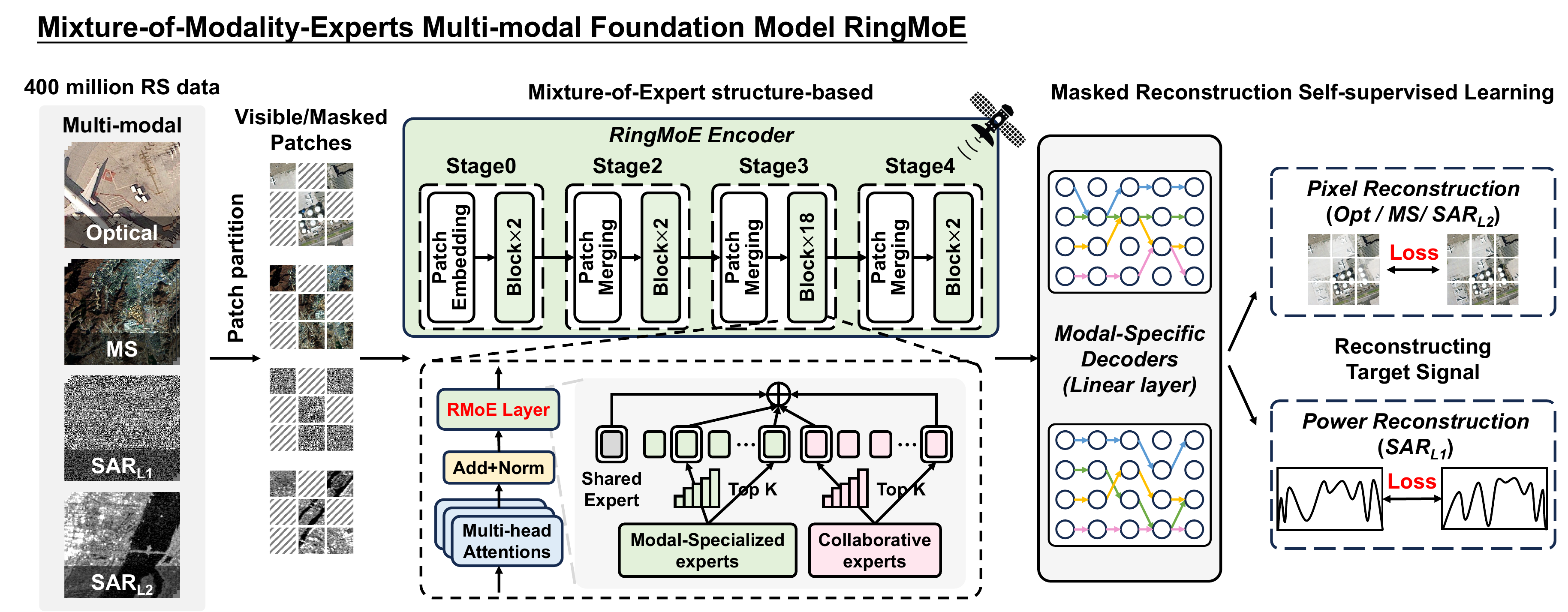}
\caption{\textbf{Overview of the proposed RingMoE framework}. Given the multi-modal inputs, the random masking operation is performed to generate visible and masked patches, i.e. tokens. {The RingMoE encoder incorporates the RMoE layer (i.e., a novel hierarchical Mixture-of-Experts structure) to replace all standard FFN layers in each stage}, effectively capturing both inter-modal commonalities and intra-modal specializations in RS. Finally, the latent representations from different modalities are decoded by modal-specific decoders to reconstruct the original targets.}
\label{fig:5}
\end{figure*}

\section{Proposed RingMoE}
This section outlines the pre-training, architecture, efficient deployment, and pre-training dataset. Sec.\ref{sec.3.1} provides an overview of the RingMoE pre-training process, followed by Sec.\ref{sec.3.2} and Sec.\ref{sec.3.3}, which detail the feature encoder and modal-specific decoder. Sec.\ref{sec.3.4} discusses the expert pruning for effective deployment on downstream tasks. Finally, Sec.\ref{sec.3.5} introduces the multi-modal dataset for pre-training.

\subsection{Overview of RingMoE Pre-Training} \label{sec.3.1}
RingMoE adopts the self-supervised learning framework for pre-training, utilizing masked image reconstruction to predict missing information, thus capturing both global structures and fine-grained details across diverse modalities. The key steps in RingMoE pre-training are as follows:
\begin{itemize}

\item \textbf{Multi-modal Input Pre-processing.} During the pre-training phase, multi-modal RS images $I_m, m \in \left \{ Opt, MS, SAR_{L1}, SAR_{L2}\right \}$ are divided into non-overlapping patches. A random masking operation $\mathbb{M}\left(\cdot \right)$ is then applied to these patches, generating visible and masked tokens $x_m$, i.e., $x_m=\mathbb{M}\odot \mathrm{Patch}\left (I_m  \right )$.


\item \textbf{Feature Representation with RingMoE Encoder.} 
These tokens are encoded into multi-modal latent representations $z_m = f_{\theta}\left(E_ex_{m} + E_{pos} \right)$ by the RingMoE encoder $f_{\theta}$. This encoder employs a hierarchical MoE architecture to balance intra-modal specialization with cross-modal generalization. Here, $E_e$ embeds input tokens into a high-dimensional space, while $E_{pos}$ provides positional information. For more details see in Sec.\ref{sec.3.2}.

\item \textbf{Target Reconstruction with Modal-Specific Decoders.} 
The latent representations $z_m$ are fed into lightweight modal-specific decoders $g_{\phi}$ to reconstruct the original data. The decoder predicts pixel values for Opt, MS, and SAR$_{L2}$, while reconstructing power values for SAR$_{L1}$, ensuring fidelity to modality-specific properties. The model is optimized by minimizing the reconstruction error between $\hat{x}_m = g_{\phi}\left(z_m\right)$ and the original inputs. For more details see Sec.\ref{sec.3.3}.


\end{itemize}

\begin{figure*}[t]
\setlength{\abovecaptionskip}{0pt}
\setlength{\belowcaptionskip}{0pt}
\centering
\includegraphics[width=0.92\linewidth]{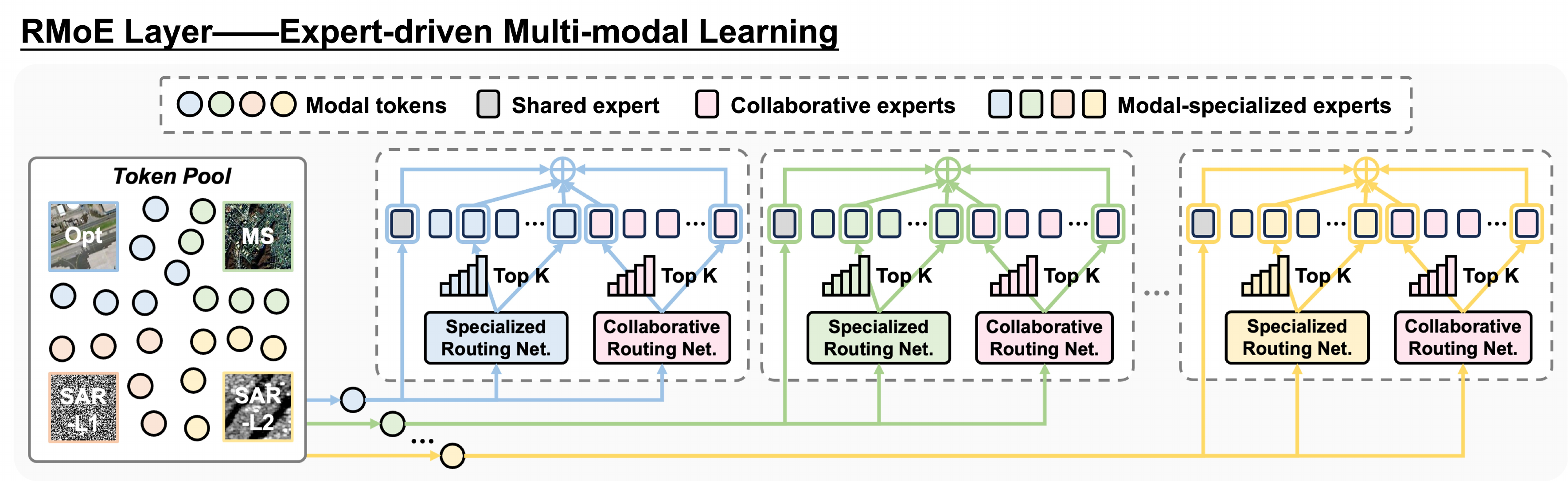}

\caption{{\textbf{Details of RMoE Layer:} RingMoE incorporates a sparse MoE architecture, replacing the standard Transformer FFN layer with modal-specialized, collaborative, and shared experts to effectively capture both intra- and inter-modal correlations.}}
\label{fig:6}
\end{figure*}

\subsection{Joint Encoder across Modalities and Experts for Multi-Modal Representation}\label{sec.3.2}
{RingMoE utilizes Swin Transformer v2~\cite{liu2022swin} as its image encoder, replacing all standard Feed-Forward Network (FFN) layers with the MoE layer to enhance model capacity and improve adaptability to multi-modal RS data (detailed in Appendix B.2).} As illustrated in Fig.\ref{fig:5}, the image tokens from different modalities $x_m$, are first transformed into feature embeddings by modal-specific linear projections $E_e$.  
Then, after embedding the position encoding $E_{pos}$, all the feature embeddings $z_m$ from different modalities are fed into the RingMoE $f_{\theta}$ to learn the latent representations $z_m$, i.e., $z_m = f_{\theta}\left(E_ex_{m} + E_{pos} \right)$. Notably, the encoder $f_{\theta}$ consists of $L$ blocks, each of which is built from the Multi-head attention (MHA) and RMoE Layer.

\subsubsection{RMoE Layer}
To further enhance model capacity and fully exploit the potential of RSFMs, {we propose replacing each FFN layer in the Swin Transformer with the MoE structure (see Appendix A for typical MoE implementations), where multiple parallel experts are dynamically activated through a routing network that assigns each token to its most relevant experts based on learned gating scores.} However, singularly sharing a group of experts across different RS modalities may lead to interference between unrelated or even conflicting modalities, causing confusion or degradation in expert knowledge. In other words, while sharing experts across certain modalities can be beneficial, it can also be detrimental when applied to modalities with distinct imaging characteristics, such as Opt and complex-valued SAR-L1 data. This challenge highlights the inherent difficulty of training a unified multi-modal RSFM.

To address this, we advocate for a structured expert design that balances cooperation across compatible modalities while preserving specialization for modality-specific knowledge. Inspired by this, we re-define the roles of experts and introduce three distinct types: Modal-specialized Experts, Collaborative Experts, and a Shared Expert (see Fig.\ref{fig:6}). This design efficiently captures both intra-modal expertise and inter-modal correlations, ensuring a rational division of multi-modal RS knowledge. 

\noindent \textbf{Modal-specialized experts.} To reduce knowledge conflicts arising from handling incompatible modalities, modal-specialized experts are assigned exclusively to specific modalities. These experts focus on intra-modal representations, achieving high specialization by isolating modality-specific knowledge and capturing nuanced features unique to each modality. For a given modality $m \in \left \{ Opt, MS, SAR_{L1}, SAR_{L2} \right \} $, the output $y_m^S$ of the modal-specialized experts is described by:
\begin{align}  \label{equation:4}
\setlength{\abovecaptionskip}{0pt}
\setlength{\belowcaptionskip}{0pt}
y_m^S=\sum_{k=1}^{N_S}G_{m;k}^{S}\left ( x_m \right )E^{S}_{m;k}\left ( x_m \right ),k=1,2,...,N_S  
\end{align}
Where $E_{m;k}^S$ denotes the $k^{th}$ modal-specialized expert from modal $m$, $G_{m;k}^S$ denotes the weight of the $k^{th}$ modal-specialized expert derived through the modal-specialized routing network $G^S_m$, $N_S$ denotes the number of the modal-specialized experts.

\noindent \textbf{Collaborative experts.} To complement modality-specific specialization, collaborative experts are introduced to capture cross-modal relationships in RS. These experts enable efficient knowledge transfer between related modalities, reducing redundancy inherent in purely modal-specialized designs and enhancing parameter efficiency. With $N_C$ experts shared across multiple modalities, the collaborative experts operate under a collaborative routing network. The token assignment from various modalities to these experts is expressed as:
\begin{align}  \label{equation:5}
\setlength{\abovecaptionskip}{0pt}
\setlength{\belowcaptionskip}{0pt}
y_m^C=\sum_{k=1}^{N_C}G_{k}^{C}\left ( x_m \right )E^{C}_{k}\left ( x_m \right ),k=1,2,...,N_C  
\end{align}
Where $E_{k}^C$ denotes the $k^{th}$ collaborative expert, $G_{m}^C$ denotes the weight of the $k^{th}$ collaborative expert derived through the collaborative routing network $G^C$, $N_C$ denotes the number of the collaborative experts.

\noindent \textbf{Shared expert.} The shared expert consolidates global knowledge across modalities, bypassing the complexity of a routing network by directly processing feature tokens from all inputs. This design reduces reliance on additional parameters in the modal-specialized and collaborative experts, enhancing efficiency while maintaining strong performance. By integrating cross-modal information, the shared expert reduces redundancy and ensures robust generalization.

Together, the triad of experts (modal-specialized, collaborative, and shared) addresses the challenges of multi-modal RS data, achieving balanced knowledge partitioning. The final output of the RMoE layer is expressed as:
\begin{equation} \label{equation:6}
\setlength{\abovecaptionskip}{0pt}
\setlength{\belowcaptionskip}{0pt}
\begin{split}
     y_m  & = y_m^S + y_m^C + E^{Shared}\left ( x_m \right )  \\
     & = \sum_{k=1}^{N_S}G_{m;k}^{S}\left ( x_m \right )E^{S}_{m;k}\left ( x_m \right )  \\
      & + \sum_{k=1}^{N_C}G_{k}^{C}\left ( x_m \right )E^{C}_{k}\left ( x_m \right ) + E^{Shared}\left ( x_m \right )
\end{split}
\end{equation}
Where $y_m^S$, $y_m^C$, and $E^{Shared}\left ( x_m \right )$ denote the output after the modal-specialized experts, the collaborative experts, and the shared expert, respectively.

\subsubsection{Optimization of the RMoE layer}
During the optimization of the MoE structure, the absence of regularization often leads the model to favor a small subset of experts while neglecting others, creating a self-reinforcing imbalance. This uneven expert selection results in inadequate training for underutilized experts, ultimately limiting both efficiency and specialization. To address this issue and ensure a balanced workload among experts, we introduce a load balance loss $\mathcal{L}_{balance}$, designed separately for cooperative experts and modal-specialized experts, formulated as follows:
\begin{align}  \label{equation:7}
\setlength{\abovecaptionskip}{0pt}
\setlength{\belowcaptionskip}{0pt}
\mathcal{L}_{balance}=\mathcal{L}_{balance}^C+\sum_{m}\mathcal{L}_{m;balance}^S
\end{align}
Where $\mathcal{L}_{balance}^C$ denotes the load-balance loss for cooperative experts, $\mathcal{L}_{m;balance}^S$ denotes the load-balance loss for modal-specialized experts from modal $m \in \left \{ Opt, MS, SAR_{L1}, SAR_{L2} \right \} $. 

\noindent \textbf{Collaborative experts.} The load balance loss for collaborative experts ensures that the workload among $N_C$ collaborative experts is distributed evenly: 
\begin{align}
\setlength{\abovecaptionskip}{0pt}
\setlength{\belowcaptionskip}{0pt}
&\mathcal{L}_{balance}^C = N_C \cdot \sum_{k=1}^{N_C} f_k \cdot P_k \tag{8} \label{equation:8} \\
&f_k = \frac{1}{N} \sum_{i=1}^{N} \mathbb{I}\left(\mathrm{Token} \ x_{m;i} \ \mathrm{selects} \ \mathrm{Expert} \ k \right) \tag{9} \label{equation:9} \\
&P_k = \frac{1}{N} \sum_{i=1}^{N} H_k \left( x_{m;i} \right) \tag{10} \label{equation:10}
\end{align}
Where $N_C$ denotes the number of cooperative experts. $f_k$ denotes the probability of tokens dispatched to expert $k$ (see Eq.\ref{equation:9}), where $N$ denotes the number of the input tokens, $\mathbb{I}(\cdot)$ is used to determine whether the $i^{th}$ input token $x_m$ is assigned to expert $k$ after passing through the routing network $G^C$ of the cooperating experts. $H\left (\cdot\right )$ denote the routing network $G^C\left (\cdot\right )$ without top-K function. 

\noindent \textbf{Modal-specialized experts.} The load balance loss for modal-specialized experts $\mathcal{L}_{m;balance}^S$ follows a similar structure. By minimizing $\mathcal{L}_{balance}$, the workload among experts is distributed more evenly, thereby promoting enhanced specialization within each expert while improving overall computational efficiency.

\subsection{Modal-Specific Decoders for Physics-Informed Self-Supervised Target Reconstruction}\label{sec.3.3}
Given the multi-modal feature representations from the RingMoE encoder, $z_m$ for each modality m $\in \left \{ Opt, MS, SAR_{L1}, SAR_{L2} \right \}$, we introduce modal-specific decoders $g_{\phi}$ tailored to each modality. These decoders reconstruct the original targets from both visible and masked image tokens, embedding sensor-specific characteristics into the self-supervised learning process. Unlike the encoder, the decoder consists of a single linear projection layer, ensuring efficiency while empirical results demonstrate that a lightweight design suffices for effective reconstruction. Mathematically, the reconstructed image tokens $\hat{x}_m $ can be formulated as $\hat{x}_m= g_{\phi }\left( f_{\theta}\left(E_ex_{m} + E_{pos} \right)\right)$. To better align with the physical characteristics of different modalities, we define the reconstruction target $t_m$ as follows:
\begin{align}  \label{equation:11}
\setlength{\abovecaptionskip}{0pt}
\setlength{\belowcaptionskip}{0pt}
t_m = \begin{cases}
I_m, \ m = \left \{ Opt,MS,SAR_{L2} \right \} 
 \\
\mathrm{Power}\left(I_m\right ), \ m =SAR_{L1}
\end{cases}
\end{align}

For Opt, MS, and SAR-L2 images, we reconstruct original pixel values to preserve spatial and spectral integrity. {For fully polarimetric SAR-L1 data, which consists of four polarization modes with complex-valued representations (each containing real and imaginary components), we adopt a physics-informed reconstruction strategy.} Instead of reconstructing raw pixel values, we leverage the principle of power conservation before and after polarization decomposition \cite{yamaguchi2005four}, reconstructing the image power as: $t_{SAR_{L1}}=\mathrm{Power}\left ( I_{SAR_{L1}} \right ) = \left |I_{SAR_{L1}}^{HH}  \right |^2 +\left |I_{SAR_{L1}}^{HV}  \right |^2+\left |I_{SAR_{L1}}^{VH}  \right |^2+\left |I_{SAR_{L1}}^{VV}  \right |^2$. Here, HH, HV, VH, and VV represent the four polarization modes. This approach ensures the reconstruction task remains physically meaningful, enhancing the model's ability to capture SAR polarimetric characteristics.

To optimize the reconstruction process, RingMoE ($f_{\theta}$) and the decoder ($g_{\phi }$) are trained end-to-end by minimizing the reconstruction loss $\mathcal{L}_{recon}$, computed as:
\begin{align}  \label{equation:12}
\setlength{\abovecaptionskip}{0pt}
\setlength{\belowcaptionskip}{0pt}
\mathcal{L}_{recon} = \sum_{m}\frac{1}{\Omega \left ( M\left (\hat{x}_m  \right )   \right ) } \sum_{i \in M\left (\hat{x}_m  \right )}\left | 
M\left (\hat{x}_{m;i}  \right ) -M\left (t_{m;i}  \right ) \right |^2
\end{align}
Where $\hat{x}_m$ and $t_m$ denote the reconstructed and original targets from modal $m$, respectively. $M\left(\cdot \right)$ denotes the set of masked pixels and $\Omega\left(\cdot\right)$ denotes the number of elements. 

The overall loss function for the pre-training phase is then formulated as:
\begin{align}  \label{equation:13}
\setlength{\abovecaptionskip}{0pt}
\setlength{\belowcaptionskip}{0pt}
\mathcal{L} = \mathcal{L}_{recon} + \alpha \mathcal{L}_{balance} 
\end{align}
Where $\alpha$ balances the contribution of two losses.

\begin{figure*}[t]
\setlength{\abovecaptionskip}{0pt}
\setlength{\belowcaptionskip}{0pt}
\centering
\includegraphics[width=0.92\linewidth]{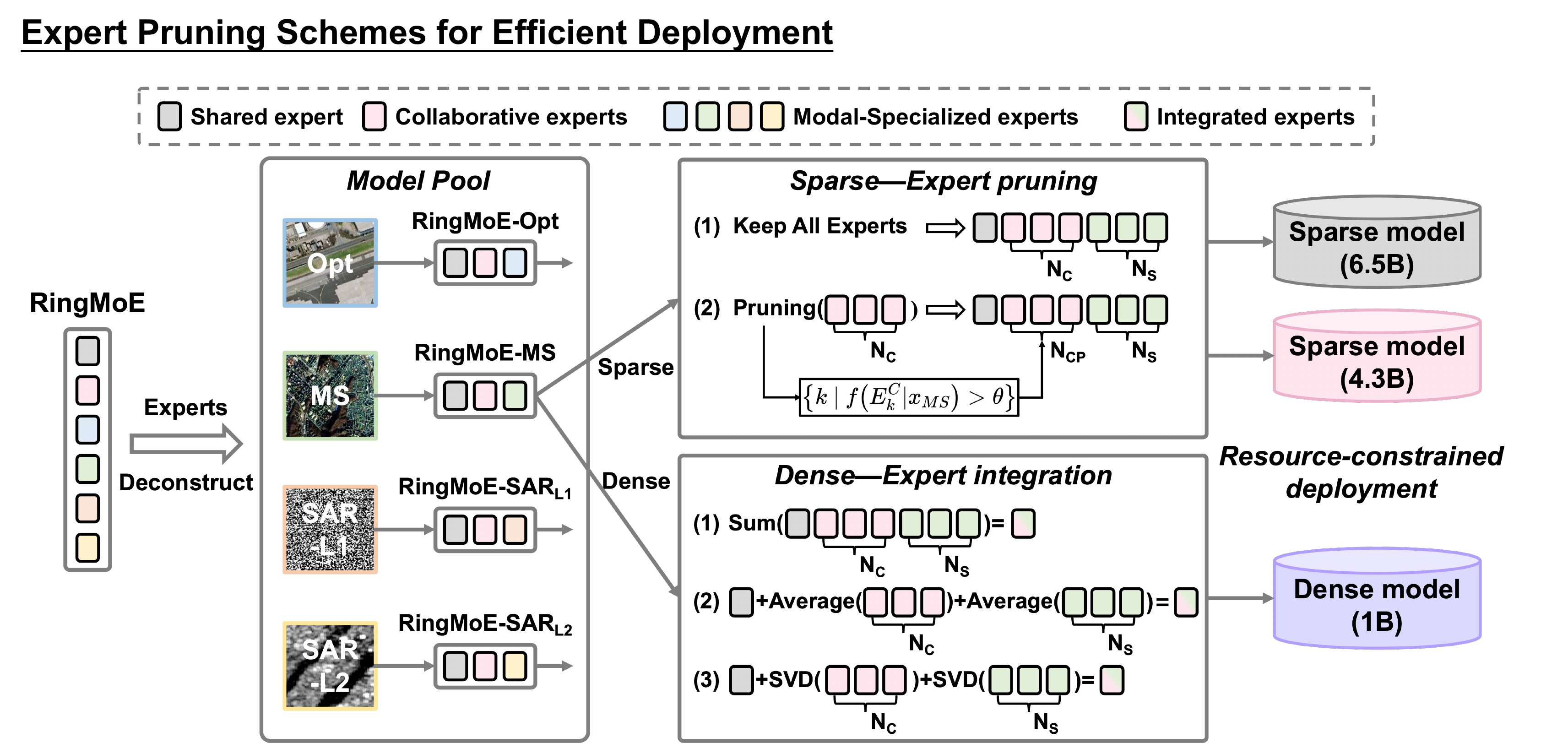}
\caption{\textbf{Details of Expert pruning.} After pre-training, RingMoE supports modular decomposition into single-modal or multi-modal sub-models of varying scales, enabling flexible adaptation to downstream tasks. RingMoE can be compressed into modality-specific (e.g., 6.5B) or lightweight (e.g., 1B) models through expert pruning, retaining adaptability for resource-constrained environments with minimal performance degradation.}
\label{fig:7}
\end{figure*}

\subsection{Dynamic Expert Pruning for Efficient Deployment across Various Computational Resources}\label{sec.3.4}
Following large-scale pre-training, RingMoE ($f_{\theta}$) serves as a universal image encoder for downstream tasks, while the modal-specific decoders ($g_{\phi}$) are discarded. Leveraging its modular expert architecture, RingMoE can be decomposed into modality-specific sub-models (e.g., RingMoE-Opt, RingMoE-MS, RingMoE-SAR$_{L1}$, RingMoE-SAR$_{L2}$). Each sub-model retains only its relevant modal-specialized, collaborative, and shared experts, rather than the full set across all modalities. This targeted expert retention reduces the parameter count from 14.7 billion to 6.5 billion, significantly lowering computational demands while maintaining performance (see Appendix Fig.3.a).

To enable efficient deployment in resource-constrained environments, we introduce tailored expert pruning strategies to further compress the model while preserving its effectiveness (see Fig.\ref{fig:7}). By selectively removing underutilized experts, the model undergoes further pruning (RingMoE-EP) while retaining essential components, such as the multi-head attention layers. For instance, RingMoE-Opt achieves a parameter reduction to 4.3 billion. Additionally, we introduce expert knowledge aggregation, transforming the sparse model into dense configurations by integrating expert knowledge through summing, averaging, or compressing. This results in RingMoE-KS, RingMoE-KA, and RingMoE-KC, each with 1 billion parameters, maintaining accuracy while significantly improving deployment flexibility. {Methodologies for pruning are as follows (see Appendix B.4 for more details):}

\subsubsection{Sparse Expert Pruning Schemes} \label{Sparse Expert Pruning Schemes}
To maximize knowledge retention in RingMoE, we initially retain all experts but also explore pruning unused or infrequently utilized ones. Specifically, for cooperative experts designed to capture inter-modal commonalities, we observe that despite implementing load-balancing loss to distribute the workload evenly across experts, certain experts are preferentially selected for tokens from specific modalities. This results in varying usage frequencies across experts when handling single-modal inputs.

{To quantify this phenomenon, we compute the activation frequency $f( E_k^C|x_m)$ of each collaborative expert $E_k^C$ on tokens from modality m, and define the retained expert set:
\begin{align}  \label{equation:14}
\setlength{\abovecaptionskip}{0pt}
\setlength{\belowcaptionskip}{0pt}
\hat{E}^C_{m} = \left \{ k \ | \ f\left ( E_k^C|x_m \right ) > \varphi ,k=1,2,...,N_C \right \}
\end{align}
Here, $\varphi$ denotes a modality-specific pruning threshold. To eliminate underutilized collaborative experts, we employ a frequency-aware expert pruning strategy (see Appendix B.4 for details). For each RMoE layer and modality $m$, we retain only the top 25\% of the most frequently activated collaborative experts, where $\varphi$ corresponds to the 75$^{th}$ percentile of the activation frequency distribution. This adaptive thresholding effectively prunes redundant experts while preserving those critical for modality representation, leading to a substantial reduction in model size with minimal performance loss. Importantly, the modal-specialized and shared experts remain intact, ensuring full capacity for capturing both modality-specific and cross-modal knowledge. Consequently, the pruning strategy yields a 4.3B-parameter RingMoE-Opt model that achieves an optimal balance between efficiency and accuracy.}

\subsubsection{Dense Expert Integration Schemes}  \label{Dense Expert Integration Schemes}
We also explore compressing the sparse RMoE model into a dense format. This involves discarding the expert selection mechanism from the routing network $G$ and consolidating all expert knowledge into a single expert, essentially converting the structure back into a standard feed-forward network (FFN) layer, reducing the total parameters to 1 billion. We propose three schemes for knowledge aggregation: summing, averaging, and compressing, each providing different strategies for integrating the expert knowledge into a unified representation.

\noindent {\textbf{Knowledge summing (KS) and averaging (KA).} For Knowledge Summing, given two linear projections $W_{k;1}^{*} \in \mathbb{R}{^{d_1\times d_2}}$ and $W_{k;2}^{*} \in \mathbb{R}{^{d_2\times d_1}}$ in every expert (modal-specialized, cooperative, and shared) in the RMoE layer, we sum the weights of all experts to aggregate their knowledge. For Knowledge Averaging, similar to summing, we average the weights of the experts.}

\noindent \textbf{Knowledge Compressing.}
{Beyond direct aggregation, we propose a low-rank fusion strategy to retain essential knowledge from all experts while achieving effective compression. As illustrated in Fig.~\ref{fig:7}, each expert’s weight matrix is approximated utilizing Singular Value Decomposition (SVD):
\begin{equation} \label{equation:17}
\setlength{\abovecaptionskip}{0pt}
\setlength{\belowcaptionskip}{0pt}
\begin{split}
W_{d_1\times d_2}&= U_{d_1\times d_1}\Sigma_{d_1\times d_2}V_{d_2\times d_2}^T \\ &\approx U_{d_1\times K}\Sigma_{K\times K}V_{K\times d_2}^T
\end{split}
\end{equation}
Where $U_{d_1\times d_1}$ and $V_{d_2\times d_2}$ are unitary matrices and $\Sigma_{d_1\times d_2}$ is a diagonal matrix with non-zero elements only on its main diagonal (i.e., singular values). We retain the top K singular values from $\Sigma_{d_1\times d_2}$ to construct the low-rank approximation, forming $U_{d_1\times K}$ and $V_{K\times d_2}$. Notably, for modal-specialized experts, $K=\frac{d_2}{N_S}$ and for collaborative experts, $K=\frac{d_2}{N_C} $. By concatenating these low-rank matrices across all expert types, a new weight matrix is generated:
\begin{align}  \label{equation:18}
\setlength{\abovecaptionskip}{0pt}
\setlength{\belowcaptionskip}{0pt}
M_1^* =\left [U_{d_1\times K},...,U_{d_1\times K}  \right ] \begin{bmatrix}
  \Sigma_{K\times K}&  & \\
  & ... & \\
  &  &\Sigma_{K\times K} 
\end{bmatrix} \begin{bmatrix}
 V_{K\times d_2}\\
 ...\\
V_{K\times d_2}
\end{bmatrix}
\end{align}
Where $M_1^* \in \mathbb{R}{^{d_1\times d_2}}$ denotes the new weight matrix capturing the essential knowledge from all experts. The final linear projection $W_1 \in \mathbb{R}{^{d_1\times d_2}}$ is then formulated as:
\begin{align}  \label{equation:19}
\setlength{\abovecaptionskip}{0pt}
\setlength{\belowcaptionskip}{0pt}
W_1 = M_1^S+M_1^C+W_1^{Shared}
\end{align}
Another projection $W_2 \in \mathbb{R}{^{d_2\times d_1}}$ has a similar process.}

To enhance adaptability across downstream tasks, the pruned and modularized RingMoE supports flexible deployment. Modality-specific sub-models can operate independently for unimodal tasks or be combined for multimodal applications. Users may select from full fine-tuning, parameter freezing, or parameter-efficient fine-tuning (PEFT) to balance accuracy and efficiency (see Appendix Fig.3.b). Such flexibility enables deployment across heterogeneous and geographically distributed computing platforms, improving resource utilization and sustainability in practical RS applications.


\begin{figure*}[t]
\setlength{\abovecaptionskip}{0pt}
\setlength{\belowcaptionskip}{0pt}
\centering
\includegraphics[width=0.85\linewidth]{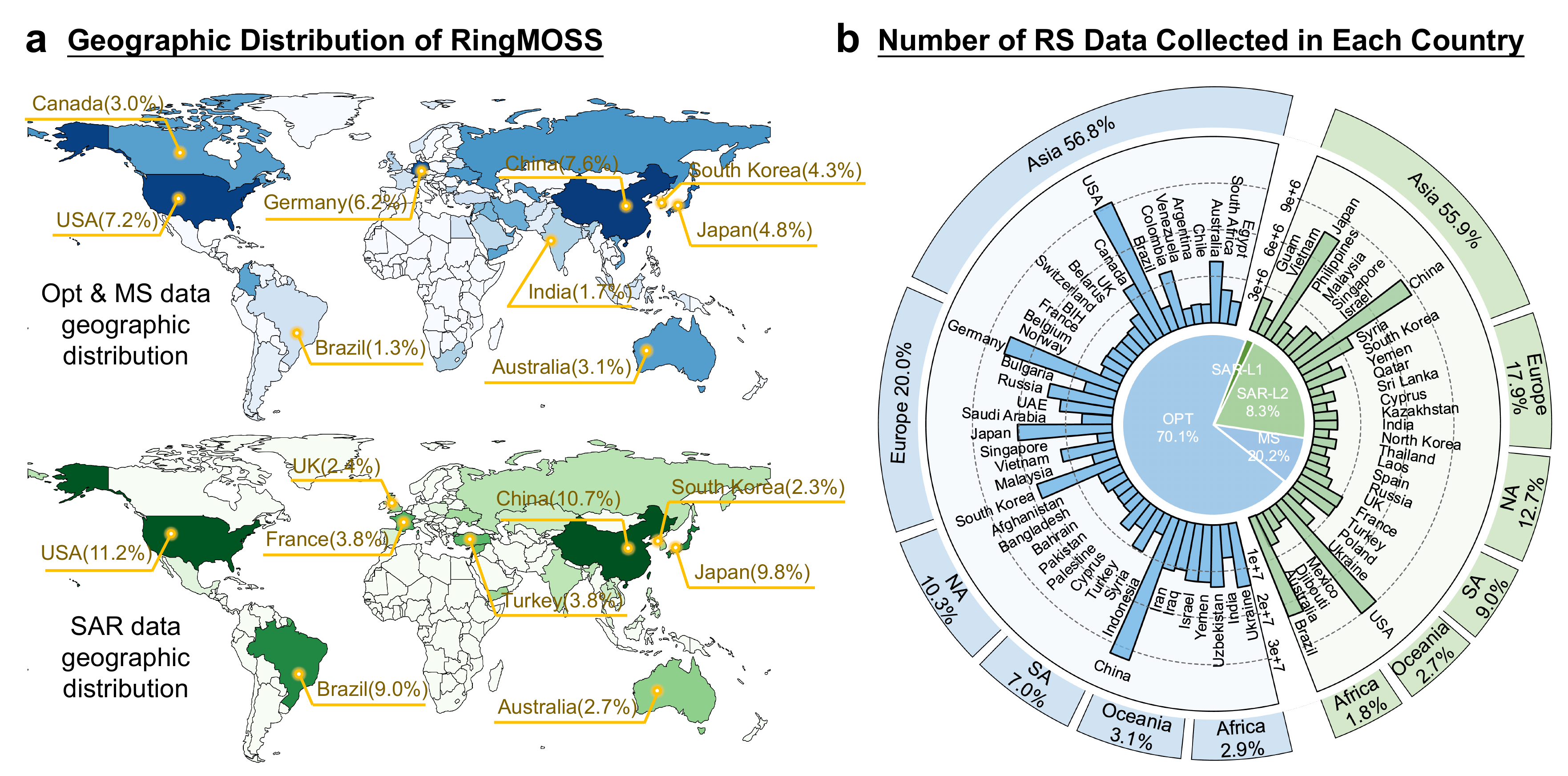}
\caption{\textbf{The global geographic distribution of RingMOSS.} The blue-shaded areas represent regions covered by optical data (Opt and MS), while the green-shaded areas indicate SAR data coverage (amplitude and complex-valued). Countries like the United States, China, and Japan contribute significantly, reflecting their advanced RS infrastructure.}
\label{fig:8}
\end{figure*}

\subsection{RingMOSS: A Comprehensive Multi-Modal Pre-Training Dataset}\label{sec.3.5}
To establish a versatile RSFM adaptable to diverse tasks, RingMoE requires a comprehensive and diverse pre-training dataset that accounts for significant variations in imaging modalities, spatial resolutions, temporal dynamics, geographic regions, and scene complexities. Thus, we curate RingMOSS\footnote[2]{{All public data, along with detailed documentation are available at:
\href{https://github.com/HanboBizl/RingMoEDatasets}{https://github.com/HanboBizl/RingMoEDatasets}.}}, a large-scale multi-modal RS dataset comprising 400 million images from nine satellite platforms, covering a broad spectrum of Earth observation scenarios. 

RingMOSS spans four key modalities: \textbf{M}S, \textbf{O}pt, and SAR in complex-valued and amplitude forms (i.e., \textbf{S}AR-L1, and \textbf{S}AR-L2) (see Appendix Fig.1-2). Opt imagery constitutes the largest portion of the dataset, with 302,862,240 images, representing 70.1\% of the total dataset. MS data includes 35,931,400 images, accounting for 20.2\%. SAR data, sourced from the GF-3 satellite, is available in two forms: SAR-L1, in complex-valued with amplitude and phase information, and SAR-L2, containing only amplitude information. By integrating amplitude and phase information, SAR provides reliable, all-weather coverage, capturing surface contours, elevation, and structural details, thus overcoming the inherent limitations of Opt and MS data~\cite{zhang2024optical,yang2024saratr}. Further details on the dataset, including collection strategies and pre-processing, can be found in Appendix B.1.

Beyond modality diversity, RingMOSS ensures extensive geographic coverage, spanning six continents and 57 countries, with significant data contributions from regions such as the United States, China, and Japan (Fig.\ref{fig:8}). This dataset includes a wide range of landscapes, urban environments, and natural terrains, capturing geospatial elements across various temporal and spatial scales. The integration of diverse modalities across global locations strengthens RingMoE’s ability to generalize across different RS tasks, making it well-suited for applications such as environmental monitoring, disaster response, and urban planning.

\section{Experiments}
This section provides a comprehensive evaluation of RingMoE, emphasizing its practical performance. Sec.\ref{Sec.4.2} evaluates RingMoE on the challenging task of recognizing unseen classes using only a single labeled sample, without additional fine-tuning, across five benchmarks. Then, Sec.\ref{Sec.4.3} compares RingMoE with SOTA foundation and task-specific models across 20 public benchmarks, covering six distinct RS interpretation tasks. Sec.\ref{Sec.4.4} presents ablation studies and in-depth analyses. Notably, the full 14.7B multi-modal model is decomposed into multiple 6.5B single-modal sub-models for evaluation, enabling both unimodal and multi-modal applications. {Additionally, we assess pruned variants, including RingMoE-EP (4.3B for Opt), RingMoE-KC (1B), RingMoE-KA (1B), and RingMoE-KS (1B), providing insights into scalability and resource-efficient deployment (See Appendix E.1 for more comprehensive evaluation).}

Further implementation details of RingMoE, including pre-training settings (Appendix B.3) and downstream tasks such as dataset descriptions, experimental settings, and evaluation metrics (Appendix B.5), are provided in the Appendix. For comprehensive performance comparisons, Appendix C presents both quantitative and qualitative analyses, {while Appendix E offers additional ablation studies on expert pruning strategies, SAR-L1 pretraining effects, and the impact of the physics-informed loss used in SAR-L1.} 

\begin{figure*}[t]
\setlength{\abovecaptionskip}{0pt}
\setlength{\belowcaptionskip}{0pt}
\centering
\includegraphics[width=0.95\linewidth]{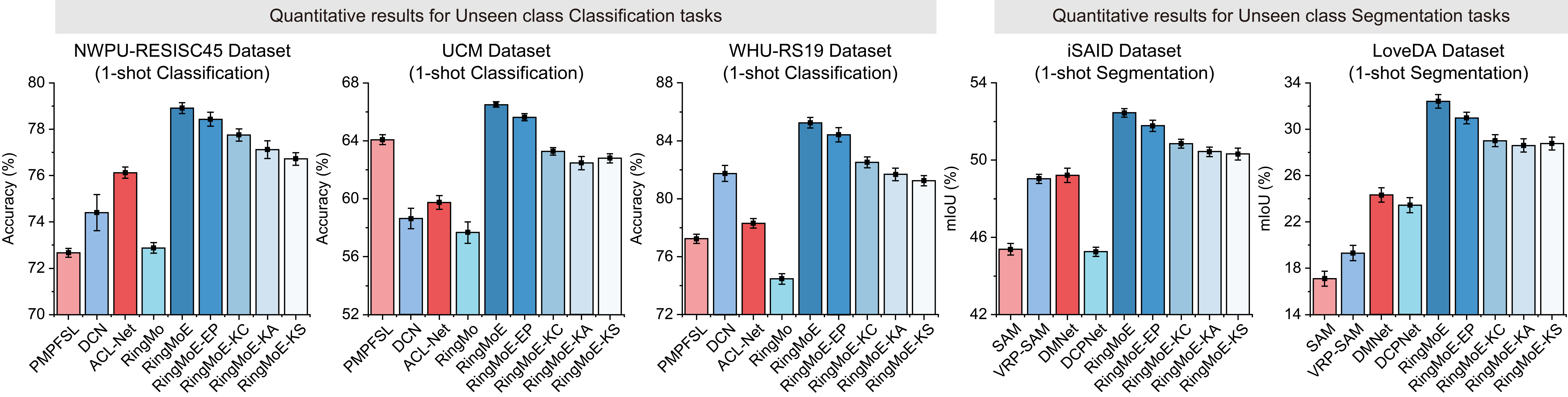}
\caption{{\textbf{Performance comparison on Unseen Class Recognition tasks}. We evaluate the proposed RingMoE foundation model against SOTA methods on diverse unseen class recognition tasks. For classification benchmarks (NWPU-RESISC45, UCM, and WHU-RS19), we report overall accuracy (OA), while for segmentation datasets (iSAID and LoveDA), mean IoU (mIoU) is utilized as the primary metric. All evaluations follow the 1-shot setting, where only a single sample is provided as guidance.}}\label{fig:9}
\end{figure*}

\subsection{Performance on Unseen Class Recognition} \label{Sec.4.2}

Due to the rapid changes in RS task requirements, comprehensive labeling of RS data has become impractical. To address this challenge, models need the ability to generalize to unseen classes with a few labeled examples, enabling faster and more flexible interpretation of novel data. To this end, we directly apply the RingMoE model to few-shot tasks (including few-shot classification~\cite{sung2018learning} and few-shot segmentation~\cite{lang2024few}) to evaluate its generalization capability. These tasks aim to quickly generalize to unseen classes with only a few labeled samples. Notably, RingMoE serves as the image encoder for feature extraction in these tasks, with its parameters fixed during both the training and testing phases (See Appendix B.5.8 for detailed settings).

Few-shot classification experiments (Fig.\ref{fig:9} and Appendix Tab.4) on the NWPU-RESIS45~\cite{cheng2017remote}, WHU-R19~\cite{sheng2012high}, and UCM~\cite{yang2010bag} datasets reveal that RingMoE with 6.5B parameters achieves the highest accuracies: 78.91\%, 85.25\%, and 66.50\%. Even after compression to 1B parameters using expert pruning techniques such as knowledge summing and averaging, e.g., RingMoE-KS and RingMoE-KA, our models retain SOTA performance. In few-shot segmentation (Fig.\ref{fig:9} and Appendix Tab.5), similar trends are observed on iSAID~\cite{waqas2019isaid} and LoveDA~\cite{wang2021loveda}, where the 6.5B parameter model achieves 52.45\% and 32.40\% mIoU, respectively. Even the pruned 4.3B model (i.e., RingMoE-EP) scores 51.78\% and 30.96\% mIoU, outperforming current methods. {Importantly, the 1B parameter version still maintains competitive SOTA performance, significantly outperforming general foundation models such as SAM~\cite{kirillov2023segment}, VRP-SAM~\cite{sun2024vrp}, underscoring the model's powerful feature extraction capabilities.}

The above results highlight RingMoE's robust generalization with minimal labeled data while maintaining leading performance even with reduced model sizes (down to 1B parameters). It offers a practical solution for real-world RS tasks under limited labeling and resource constraints.

\subsection{Performance on Single/Multi-modal Interpretation Tasks}\label{Sec.4.3}
We validate RingMoE across 20 benchmarks spanning single and multi-modal tasks, covering diverse data modalities and common tasks in remote sensing. {We also compare RingMoE with recent multimodal large language models (MLLMs) to highlight its advancements.} In scene classification, we evaluate the performance of various parameter-scale versions of RingMoE. For other downstream tasks like segmentation and detection, we focus on the 1B version of RingMoE, considering practical applications and deployment constraints. Detailed experimental settings for each task are provided in Appendix B.5.

\noindent \textbf{(1) Scene classification.}
We evaluate RingMoE and other RSFMs on two optical classification benchmarks, NWPU-RESISC45~\cite{cheng2017remote} and AID~\cite{xia2017aid} (Tab.\ref{cls}). RingMoE achieves 95.90\% and 98.19\% accuracies, exceeding the previous SOTA SkySense~\cite{guo2024skysense} by 1.05\% and 0.51\%, respectively. T-SNE visualizations (Appendix Fig.4) further confirm its discriminative feature representation, showing tighter intra-class clusters and larger inter-class separation than competing RSFMs. For efficient deployment, we investigate pruning-based compression. The 4.3B RingMoE-EP (Opt), using sparse expert pruning, achieves 95.43\% and 97.88\% accuracies, nearly matching the full model. Even the 1B dense-compressed variant (RingMoE-KC) maintains strong performance, surpassing most RSFMs and approaching the 2B SkySense. These results demonstrate that RingMoE preserves powerful feature extraction and generalization ability even under significant parameter reduction. Given its practical advantages, we further evaluate the 1B-parameter version (RingMoE-KC) across other downstream tasks.



\begin{table}[t]
\setlength{\abovecaptionskip}{0pt}
\setlength{\belowcaptionskip}{0pt}
\centering
\renewcommand\arraystretch{1.25}
\caption{Detailed performance comparison on \textbf{Scene Classification} tasks, i.e., RESISI-45 and AID. \textbf{Bold} and \underline{underline} denote the best and second-best results, respectively.}\label{cls}
\resizebox{0.95\linewidth}{!}{
\begin{threeparttable}
\begin{tabular}{r|c|cc} 
\toprule
\multirow{2}{*}{Foundation model} & \multirow{2}{*}{Backbone}  & \begin{tabular}[c]{@{}c@{}}NWPU-RESISC45\\(TR=10\%)\end{tabular} & \begin{tabular}[c]{@{}c@{}}AID\\(TR=20\%)\end{tabular}  \\ \cline{3-4}
                                  &                             & OA(\%)              & OA(\%)          \\ 
\hline
GASSL~\cite{ayush2021geography}                    & ResNet-50                  & 90.86                                                        & 93.55                                                   \\
SeCo~\cite{manas2021seasonal}                      & ResNet-50                  & 89.64                                                        & 93.47                                                   \\
SatMAE~\cite{cong2022satmae}                   & ViT-Large                  & 91.72                                                        & 95.02                                                   \\
RingMo~\cite{sun2022ringmo}                   & Swin-Base                   & 94.25                                                        & 96.9                                                    \\
RVSA~\cite{wang2022advancing}                    & ViT-Base                    & 93.93                                                        & 97.03                                                   \\
CMID~\cite{CMID}                     & Swin-Base                   & 94.05                                                        & 96.11                                                   \\
CACo~\cite{CACo}                     & ResNet-50                   & 88.28                                                        & 90.88                                                   \\
SatLas~\cite{bastani2023satlaspretrain}                   & Swin-Base                    & 92.16                                                        & 94.96                                                   \\
GFM~\cite{GFM}                      & Swin-Base                   & 92.73                                                        & 95.47                                                   \\
Scale-MAE~\cite{reed2023scale}                & ViT-Large                   & 92.63                                                        & 96.44                                                   \\
SelectiveMAE~\cite{wang2024scaling} & ViT-Large & 94.57 &97.25 \\
SkySense (2B)~\cite{guo2024skysense}                & Swin-Huge                    & 94.85                                                        & 97.68                                                   \\ 
\hline
\multirow{5}{*}{\textbf{RingMoE}} & RingMoE(6.5B)$^\dagger$              & \textbf{95.90} & \textbf{98.19}                                                \\
& RingMoE-EP(4.3B)$^\dagger$              & \underline{95.43}  & \underline{97.88}         \\
& RingMoE-KS(1B)$^\dagger$             & 94.81  & 97.30                                     \\
& RingMoE-KA(1B)$^\dagger$             & 94.87  & 97.23                                     \\
& RingMoE-KC(1B)$^\dagger$             & 95.05  & 97.42                                    \\
\bottomrule
\end{tabular}
\begin{tablenotes}
\footnotesize
\item[$\dagger$] RingMoE(6.5B) represents the modal-specific complete model, i.e., RingMoE-Opt. RingMoE-EP(4.3B) represents the RingMoE-Opt after pruning partial experts. RingMoE-KS(1B), -KA(1B), and -KC(1B) represent the models after summing, averaging, and compressing all expert knowledge, respectively.
\end{tablenotes}
\end{threeparttable}

}
\end{table}

\begin{table}[t]
\setlength{\abovecaptionskip}{0pt}
\setlength{\belowcaptionskip}{0pt}
\centering
\renewcommand\arraystretch{1.25}
\caption{Detailed performance comparison on \textbf{Optical and Multi-spectral Semantic Segmentation} tasks. \textbf{Bold} and \underline{underline} denote the best and second-best results, respectively.}
\label{sementaticsegmentation}
\resizebox{0.95\linewidth}{!}{
\begin{tabular}{r|c|ccc} 
\toprule
\multirow{2}{*}{Foundation model} & \multirow{2}{*}{Backbone} & iSAID & Potsdam &Dyna.-pla  \\ 
\cline{3-5}
&                                       & mIoU(\%)           & mF1(\%)              & mIoU(\%)                \\ 
\hline
GASSL~\cite{ayush2021geography}                            & ResNet-50 & 65.95          & 91.27            & 40.8                \\
SeCo~\cite{manas2021seasonal}                            & ResNet-50                              & 57.20          & 89.03            & -                   \\
SatMAE~\cite{cong2022satmae}                          & ViT-Large                               & 62.97          & 90.63            & 39.9                \\
RingMo~\cite{sun2022ringmo}                           & Swin-Base                               & 67.20          & 91.27            & -                   \\
RVSA~\cite{wang2022advancing} &ViT-Base &64.49 &- &44.4 \\
CMID~\cite{CMID}                             & Swin-Base                               & 66.21          & 91.86            & 43.5                \\
CACo~\cite{CACo}                             & ResNet-50                               & 64.32          & 91.35            & 42.7                \\
SAMRS~\cite{SAMRS}                            & ViT-Base                               & 66.26          & 91.43            & -                   \\
SatLas~\cite{bastani2023satlaspretrain}                          & Swin-Base                               & 68.71          & 91.28            & 40.7                \\
GFM~\cite{GFM}                              & Swin-Base                               & 66.62          & 91.85            & 45.6                \\
Scale-MAE~\cite{reed2023scale}                        & ViT-Large                               & 65.77          & 91.54            & 41.7                \\
SkySense(2B)~\cite{guo2024skysense}                         & Swin-Huge                               & \textbf{70.91}          & \textbf{93.99}            & \underline{46.5}                \\ 
\hline
\textbf{RingMoE-KC}                         & -                                     & \underline{69.70}          & \underline{93.54}            & \textbf{47.6}                \\
\bottomrule
\end{tabular}}
\end{table}

\noindent \textbf{(2) Semantic segmentation.}
\begin{itemize}
\item Optical/Multi-spectral segmentation. {We evaluate RingMoE and other RSFMs on two optical datasets, iSAID~\cite{waqas2019isaid} and Potsdam~\cite{isprs2018semanticlabeling}}, as well as the multi-spectral dataset Dyna.-pla~\cite{toker2022dynamicearthnet} (Tab.\ref{sementaticsegmentation} and Fig.\ref{fig:11}.a). RingMoE achieves SOTA performance on the Dyna.-pla dataset and ranks second on iSAID and Potsdam, just behind SkySense~\cite{guo2024skysense}. Notably, RingMoE accomplishes these results with a compressed 1B-parameter version, highlighting its efficiency and effectiveness. Confusion matrix analysis in Appendix Fig.5 reveals RingMoE's stable per-category performance. Qualitative results in Fig.\ref{fig:11}.b and Appendix Fig.7.a-b highlight its superior detail capture, outperforming comparable RSFMs like Satlas~\cite{bastani2023satlaspretrain} and SatMAE~\cite{cong2022satmae}.

\item SAR segmentation. We evaluate RingMoE and several specialized methods on two SAR datasets: the complex-valued SAR-L1 dataset SARSegL1~\cite{DENet(SARsegL1)} and the amplitude SAR-L2 dataset AIR-POLSAR-SEG~\cite{AIR-PolSAR-Seg} (Fig.\ref{fig:11}.a). On SARSegL1, RingMoE achieves significant improvements, surpassing existing methods in mIoU and OA by 18.21\% and 9.91\%, respectively, with the highest IoU across all categories (Appendix Tab.6). On AIR-POLSAR-SEG, it surpasses the second-best method by 5.33\% in mIoU, with notable gains in Industrial Area (12.8\%), Natural Area (4.19\%) and Water (6.35\%) (Appendix Tab.7). Qualitative results in Fig.\ref{fig:11}.b and Appendix Fig.7.c-d highlight its superior precision in processing complex SAR features.

\item Multi-modal segmentation. RingMoE is evaluated on two multi-modal segmentation datasets, WHU-OPT-SAR~\cite{li2022mcanet} and DFC23~\cite{DFCdata}, which combine optical and amplitude SAR-L2 modalities, using mIoU and OA metrics (Fig.\ref{fig:11}.a and Appendix Tab.8-9). It achieves SOTA performance on both datasets, with mIoU and OA of 54.7\% and 84.1\% on WHU-OPT-SAR, and 73.4\% and 95.8\% on DFC23. Fig.\ref{fig:11}.b and Appendix Fig.7.e qualitatively demonstrate RingMoE's advanced multimodal segmentation capabilities.
\end{itemize}

\begin{table}[t]
\setlength{\abovecaptionskip}{0pt}
\setlength{\belowcaptionskip}{0pt}
\caption{Detailed performance comparison on \textbf{Optical Object Detection} tasks, i.e., DIOR and DIOR-R. \textbf{Bold} and \underline{underline} denote the best and second-best results, respectively.}\label{dior_det}
\renewcommand\arraystretch{1.25}
\centering
\resizebox{0.9\linewidth}{!}{
\begin{tabular}{r|c|cc} 
\toprule
\multirow{2}{*}{Foundation model}  & \multirow{2}{*}{Backbone} & DIOR      & DIOR-R   \\ 
\cline{3-4}
&                              & mAP$_{50}$(\%) & mAP$_{50}$(\%)  \\ 
\hline
SatMAE~\cite{cong2022satmae}                   & ViT-Large                    & 70.89     & 65.66    \\
RingMo~\cite{sun2022ringmo}                   & Swin-Base                    & 75.90      & -        \\
RVSA~\cite{wang2022advancing}                     & ViT-Base                    & 73.22     & 71.05    \\
CMID~\cite{CMID}                     & Swin-Base                   & 75.11     & 66.37    \\
CACo~\cite{CACo}                     & ResNet-50                    & 66.91     & 64.10     \\
SatLas~\cite{SAMRS}                   & Swin-Base                    & 74.10      & 67.59    \\
GFM~\cite{GFM}                      & Swin-Base                    & 72.84     & 67.67    \\
Scale-MAE~\cite{reed2023scale}      &  ViT-Large                             & 73.81     & 66.47    \\
BFM(2.4B) ~\cite{cha2023billion} & ViT-Giant & - & 73.62 \\
SelectiveMAE~\cite{wang2024scaling} & ViT-Large &77.80 & 70.31 \\
SkySense(2B)~\cite{guo2024skysense}            & Swin-Huge                    & \underline{78.73}     & \underline{74.27}    \\ 

\hline
\textbf{RingMoE-KC} & -                            & \textbf{82.45}     & \textbf{76.04}       \\
\bottomrule
\end{tabular}}
\end{table}

\noindent \textbf{(3) Object detection.}
\begin{itemize}

\item Optical detection. RingMoE is evaluated on three optical detection datasets: DIOR-horizontal (DIOR)~\cite{dior}, DIOR-rotation (DIOR-R), and HRSC2016~\cite{hrsc2016} (Tab.\ref{dior_det} and Fig.\ref{fig:12}.a). It achieves SOTA performance on DIOR and DIOR-R, surpassing SkySense~\cite{guo2024skysense} by 3.72\% and 1.77\% mAP$_{50}$, respectively. On HRSC2016, RingMoE also exceeded the previous SOTA by 1.5\% mAP$_{50}$. These results are further supported by qualitative visualizations in Fig.\ref{fig:12}.b and Appendix Fig.8.a, underscoring its superior optical detection capabilities.
\item SAR Detection. RingMoE achieves SOTA performance on two amplitude SAR-L2 detection datasets, with 94.2\% mAP$_{50}$ on HRSID~\cite{wei2020hrsid} (+1.2\%) for ship detection and an 8.1\% mAP$_{50}$ improvement on SAR-AIRcraft-1.0~\cite{zhirui2023sar} for plane detection (Fig.\ref{fig:12}.a and Appendix Tab.10). Fig.\ref{fig:12}.b and Appendix Fig.8.b highlights its robustness in complex scenarios such as harbors and airports, showcasing its adaptability to SAR detection.
\end{itemize}

\begin{figure*}[t]
\setlength{\abovecaptionskip}{0pt}
\setlength{\belowcaptionskip}{0pt}
\centering
\includegraphics[width=0.92\linewidth]{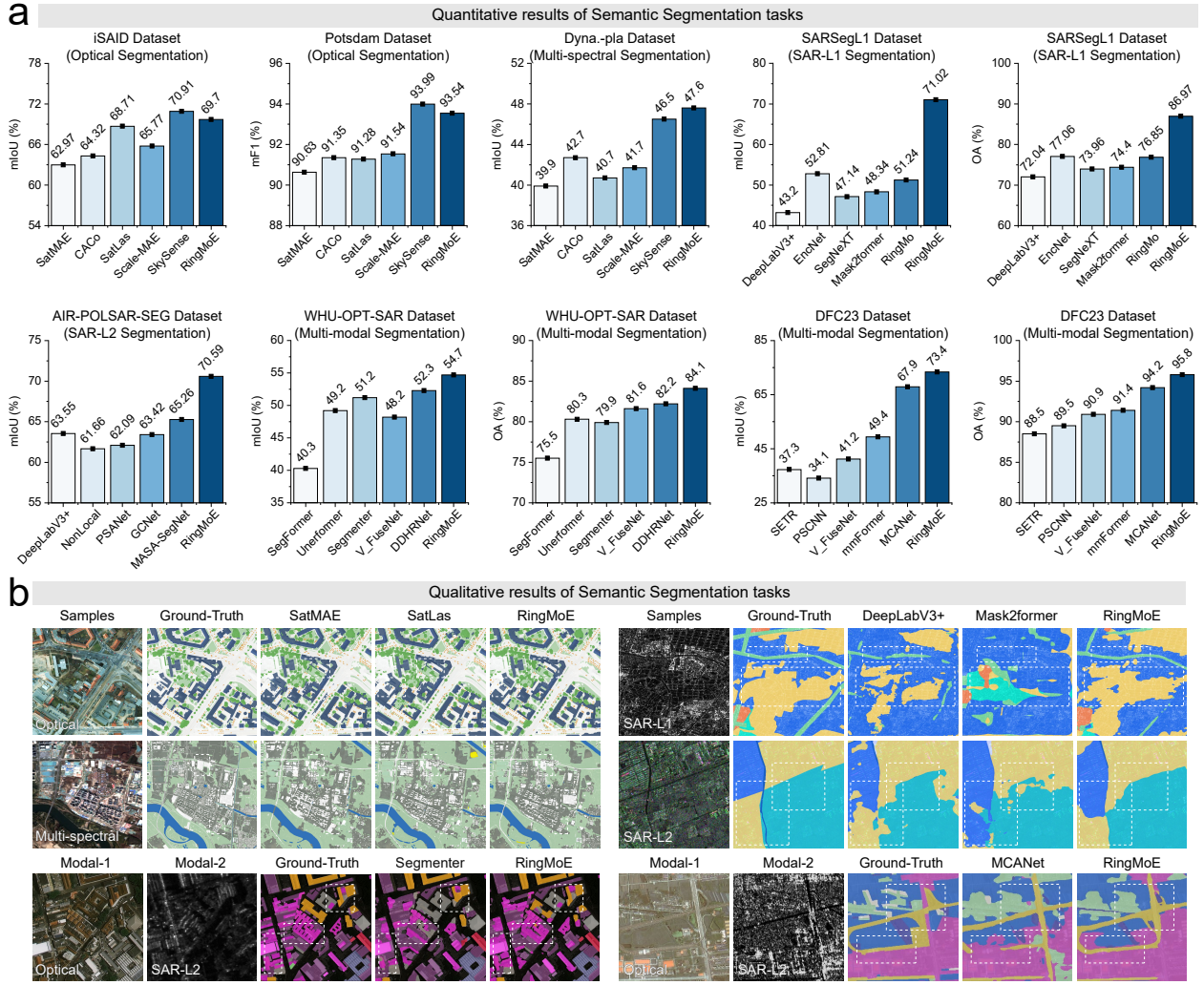}
\caption{\textbf{Performance comparison on Semantic Segmentation.} \textbf{a. Quantitative analysis:} We perform a quantitative comparison across diverse segmentation datasets, covering optical, multi-spectral, SAR-L1 (complex-valued), SAR-L2 (amplitude), and multi-modal scenarios, using mean IoU (mIoU), mean F1 score (mF1) and overall accuracy (OA) as evaluation metrics. \textbf{b. Qualitative analysis:} Examples from multiple segmentation tasks highlight RingMoE's adaptability across different modalities, showcasing its effectiveness in various remote sensing scenarios.}\label{fig:11}
\end{figure*}

\noindent \textbf{(4) Object tracking.}
RingMoE achieves SOTA performance on the AIR-MOT~\cite{he2022multi} and AIR-HSAO~\cite{ren2024motion} optical tracking datasets (Fig.\ref{fig:12}.a and Appendix Fig.6.b). On AIR-MOT, it improves MOTA by 1.3\% and IDF1 by 1.9\% over the previous SOTA. Similarly, on AIR-HSAO, RingMoE achieves a 2.2\% improvement in MOTA and a 2.0\% increase in IDF1. Sample visualizations in Fig.\ref{fig:12}.c and Appendix Fig.9.a demonstrate RingMoE’s capability to accurately capture and track objects in complex remote sensing scenarios.

\begin{figure*}[t]
\setlength{\abovecaptionskip}{0pt}
\setlength{\belowcaptionskip}{0pt}
\centering
\includegraphics[width=0.92\linewidth]{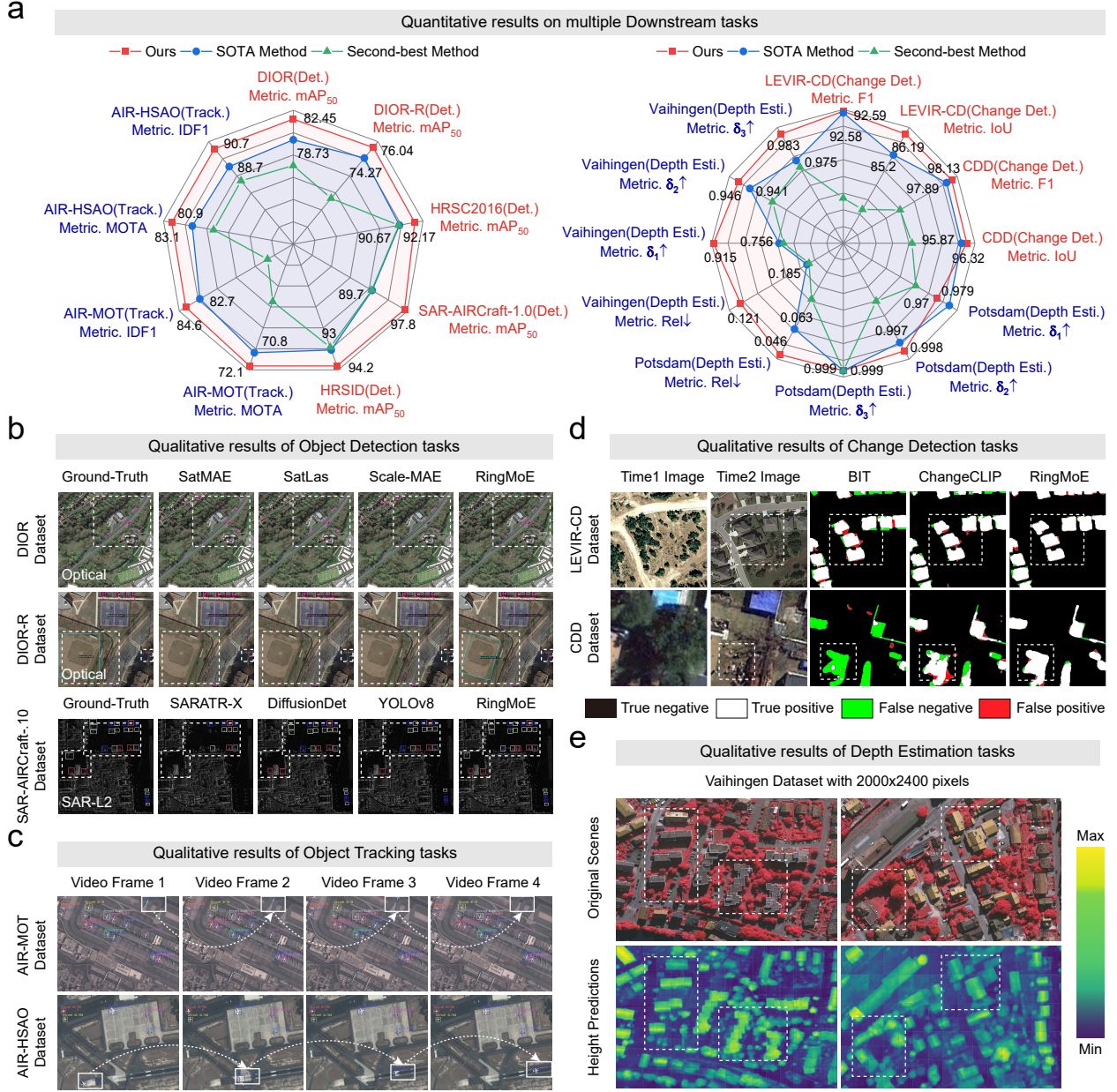}
\caption{\textbf{Performance Comparison across Object Detection, Object Tracking, Change Detection, and Depth Estimation tasks.} \textbf{a. Quantitative Analysis:} RingMoE is quantitatively compared with current SOTA methods across four tasks. {On the left column, object detection (red labels) is evaluated using mAP${_{50}}$, object tracking (blue labels) is assessed with MOTA and IDF1. On the right column, change detection (red labels) uses F1 score and IoU, while depth estimation (blue labels) is measured with Rel, $\delta{1}$, $\delta_{2}$, and $\delta_{3}$ as primary indicators.} \textbf{b,c,d,e. Qualitative Analysis:} Selected examples from detection, tracking, change detection, and depth estimation tasks highlight RingMoE's adaptability and superior performance across diverse scenarios.}
\label{fig:12}
\end{figure*}

\noindent \textbf{(5) Change detection.}
Fig.\ref{fig:12}.a and Appendix Tab.11 showcase RingMoE's superior performance on the LEVIR-CD~\cite{chen2020spatial} and CDD~\cite{lebedev2018change} datasets, outperforming all competing RSFMs and task-specific methods to achieve the highest metrics on both benchmarks. By leveraging its pre-trained foundation architecture, RingMoE effectively extracts generalized features, significantly improving semantic change detection in bi-temporal images. Visualizations (Fig.\ref{fig:12}.d and Appendix Fig.9.b) illustrate its accuracy in detecting changes with minimal false positives.

\noindent \textbf{(6) Depth estimation.}
{RingMoE achieves SOTA performance on two large-scale RS depth estimation benchmarks, ISPRS Vaihingen and Potsdam~\cite{isprs2018semanticlabeling} (Fig.\ref{fig:12}.a and Appendix Tab.12).} On the Vaihingen dataset, RingMoE reduces the Rel error by 35\% compared to the previous leading model, Heightformer, and significantly improves accuracy metrics $\delta_1$, $\delta_2$, and $\delta_3$ by 15.9\%, 0.5\%, and 1.0\%, respectively, marking a substantial advancement in depth estimation. Similarly, on the Potsdam dataset, RingMoE achieves the best performance. Visualization results on the ISPRS Vaihingen dataset (Fig.\ref{fig:12}.e and Appendix Fig.9.c) further validate RingMoE's precision, closely aligning with ground truth, particularly in areas dense with buildings and trees. These findings highlight RingMoE's capability for accurate and detailed depth estimation.

\noindent {\textbf{(7) Comparison with MLLMs.} 
Recent multimodal large language models (MLLMs) have shown strong generalization in vision-language tasks. We compare RingMoE with general MLLMs (e.g., MiniGPTv2~\cite{chen2023minigpt}, LLaVA-1.5~\cite{liu2024improvedbaselinesvisualinstruction}) and RS-specific MLLMs (e.g., GeoChat~\cite{kuckreja2024geochat}, EarthGPT~\cite{zhang2024earthgpt}) on several key RS tasks: optical classification, optical/SAR-L2 detection (Appendix Tab.13–15 in D). RingMoE consistently outperforms all MLLMs. On NWPU-RESISC45 classification, it achieves 96.28\% OA using only 20\% of training data. On DIOR detection, it surpasses the best RS-MLLMs by over 17\% mAP$_{50}$. Notably, on the challenging SARDet-100k~\cite{li2024sardet}, RingMoE reaches 89.72\% mAP$_{50}$, while general MiniGPTv2 completely fail (0.00\%). These results demonstrate the effectiveness of RingMoE in perception and semantic interpretation across diverse remote sensing scenarios.
}

\begin{figure*}[t]
\setlength{\abovecaptionskip}{0pt}
\setlength{\belowcaptionskip}{0pt}
\centering
\includegraphics[width=0.82\linewidth]{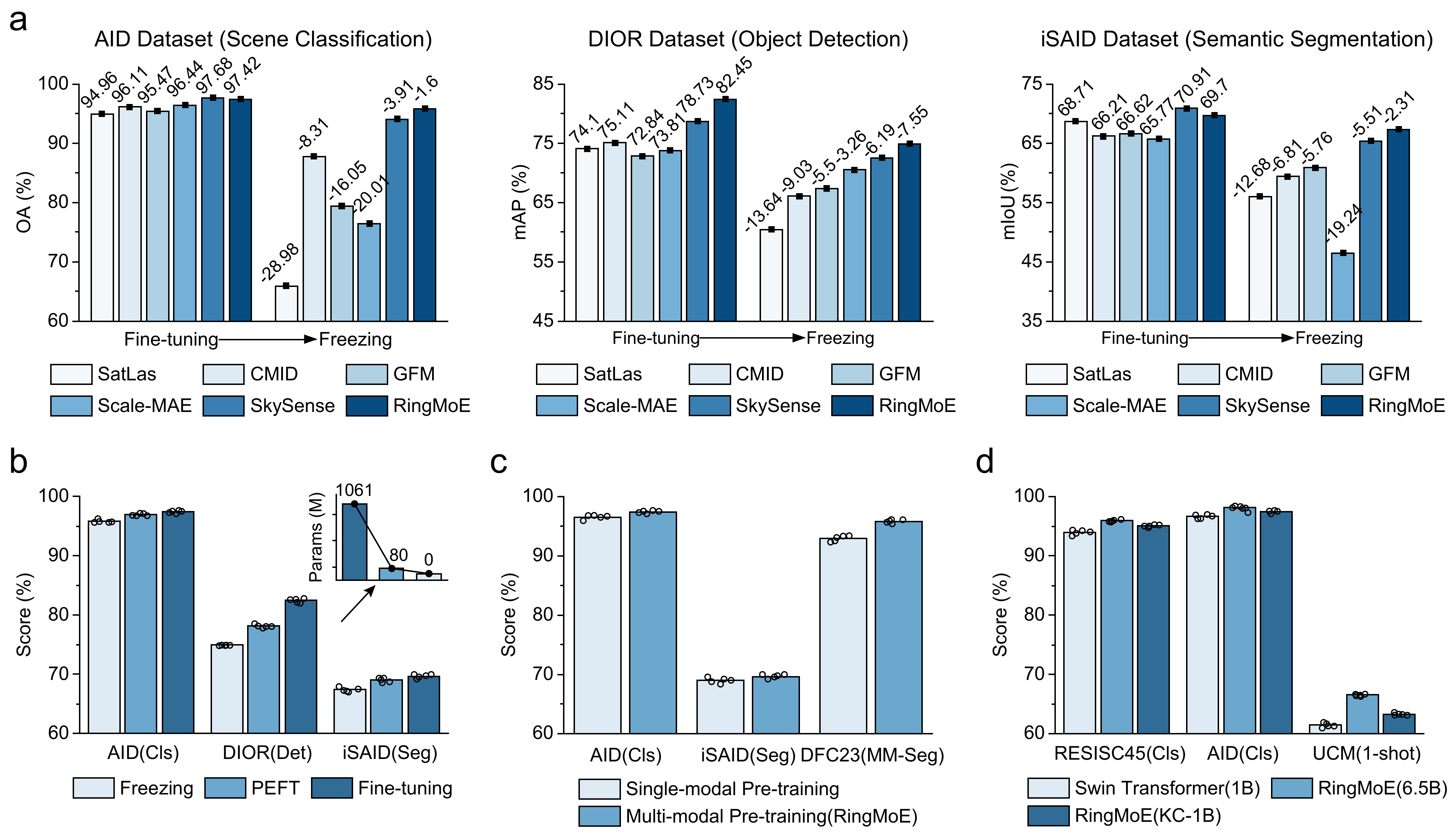}
\caption{\textbf{Ablation Studies and Analysis of RingMoE.} \textbf{a. Generalization Analysis:} We examine the generalization capabilities of pre-trained feature representations from various RSFMs by freezing backbone parameters and fine-tuning only task-specific heads across three downstream tasks: scene classification, object detection, and semantic segmentation. \textbf{b. Parameter-efficient Fine-tuning (PEFT):} To reduce resource demands while preserving accuracy, we employ the Convpass fine-tuning strategy, incorporating lightweight adaptation modules into pre-trained RingMoE. This method freezes RingMoE's original weights and only fine-tunes the added modules. Experiments are conducted on AID classification, DIOR detection, and iSAID segmentation, repeated across five random seeds. \textbf{c. Necessity of Multi-modal Pre-training:} To evaluate the advantages of RingMoE’s multi-modal pre-training, we construct uni-modal RingMoE variants and perform comparative experiments. These variants are pre-trained solely on optical/SAR images under identical settings to RingMoE. The experiments cover AID classification, iSAID segmentation, and DFC23 multi-modal segmentation tasks, repeated across five random seeds. \textbf{d. Effectiveness of Scaling Foundation Model to 10-Billion-scale Parameters:} To explore the practicality of developing a foundation model with 10-billion-scale parameters, we pre-train a 1-billion-parameter Swin Transformer v2 using SimMIM self-supervised on the same optical dataset as RingMoE. Experiments are conducted on NWPU-RESISC45 and AID scene classification and UCM unseen class recognition tasks, repeated across five random seeds.
}
\label{fig:13}
\end{figure*}

\subsection{Ablation Studies and Analysis}\label{Sec.4.4}

\subsubsection{Generalization Analysis of Remote Sensing Foundation Models} 
To evaluate the generalization capability of RingMoE's learned representations, we freeze its backbone parameters and fine-tune only the task-specific heads across three downstream tasks: scene classification, object detection, and semantic segmentation. As shown in Fig.\ref{fig:13}.a, recent RSFMs \cite{CMID,bastani2023satlaspretrain,GFM,reed2023scale,guo2024skysense} suffer significant performance degradation under this setting. For instance, SatLas~\cite{bastani2023satlaspretrain} exhibits a 28.98\% accuracy drop on the AID classification dataset, indicating limited generalization of its pre-trained features across diverse RS tasks. In contrast, RingMoE maintains high performance across all three tasks, achieving 95.82\% on AID classification, 74.90\% on DIOR detection, and 67.39\% on iSAID segmentation. These results demonstrate that RingMoE effectively captures robust and transferable feature representations during pretraining, enabling strong performance even with a frozen backbone.

\subsubsection{Discussion of RingMoE's Usage for Downstream} 
To enhance RingMoE's resource efficiency while maintaining accuracy, we evaluate its performance using parameter-efficient fine-tuning (PEFT). Specifically, the Convpass strategy~\cite{jie2024convolutional} introduces lightweight adaptation modules into the pre-trained RingMoE, freezing the core model weights while fine-tuning only these modules. Fig.\ref{fig:13}.b compares three fine-tuning strategies, i.e., freezing, PEFT, and full fine-tuning. Remarkably, RingMoE with PEFT achieves near-equivalent accuracy to full fine-tuning for tasks such as semantic segmentation, while using only 7.5\% of the parameters (i.e., 80M) required for full fine-tuning. This balance between precision and reduced resource demands underscores RingMoE’s suitability for real-world RS applications, enabling efficient and rapid interpretation on resource-constrained systems.

\begin{table}[t]
\setlength{\abovecaptionskip}{0pt}
\setlength{\belowcaptionskip}{0pt}
\caption{{Performance comparison between original RingMoE@Swin-1B and a lightweight variant (RingMoE@Swin-88M), evaluated on two RS classification datasets. Both models use identical settings; the only difference lies in the backbone size (88M vs. 1B).}} \label{tab:8}
\renewcommand\arraystretch{1.25}
\centering
\resizebox{0.9\linewidth}{!}{
\begin{tabular}{cc|cc} 
\toprule
\multirow{2}{*}{Method}  & \multirow{2}{*}{Backbone} & \begin{tabular}[c]{@{}c@{}}NWPU-RESISC45\\(TR=10\%)\end{tabular} & \begin{tabular}[c]{@{}c@{}}AID\\(TR=20\%)\end{tabular}  \\ 
\cline{3-4}
&                           & OA(\%)                                                           & OA(\%)                                                  \\ 
\hline
\multirow{2}{*}{RingMoE} 
& RMoE@Swin-88M             & 93.64                                                            & 96.57    \\
& RMoE@Swin-1B              & 95.90                                                            & 98.19                                                   \\
\bottomrule
\end{tabular}
}
\end{table}

\begin{table*}[t]
\setlength{\abovecaptionskip}{0pt}
\setlength{\belowcaptionskip}{0pt}
\caption{{Ablation study of the \textbf{Expert Design of RMoE} on four key RS tasks, including optical classification, SAR detection, multi-spectral segmentation, and multi-modal segmentation. We evaluate the impact of each expert type by selectively removing them.}} \label{tab:9}
\renewcommand\arraystretch{1.25}
\centering
\resizebox{0.8\linewidth}{!}{
\begin{tabular}{ccc|cccc} 
\toprule
\multicolumn{3}{c|}{\multirow{2}{*}{Expert Design in RMoE}} & Optical Cls.   & SAR-L2 Det.    & Multi-spectral Seg. & Multi-modal Seg.  \\
\multicolumn{3}{c|}{}                                       & NWPU-RESISC45~\cite{cheng2017remote}  & HRSID~\cite{wei2020hrsid}          & Dyna.-pla~\cite{toker2022dynamicearthnet}           & WHU-OPT-SAR~\cite{li2022mcanet}       \\ 
\hline
Modal-specialized & Collaborative & Shared                  & Acc(\%)        & mAP$_{50}$(\%)      & mIoU(\%)            & mIoU(\%)          \\ 
\hline
\ding{51}                 & \textcolor[rgb]{0.702,0.702,0.702}{\ding{55}}              & \ding{51}                       & 94.15          & 93.66          & 47.09               & 52.63             \\
\textcolor[rgb]{0.702,0.702,0.702}{\ding{55}}                  & \ding{51}             & \ding{51}                       & 93.78          & 93.24          & 46.28               & 53.39             \\
\ding{51}                 & \ding{51}             & \textcolor[rgb]{0.702,0.702,0.702}{\ding{55}}                        & \underline{94.59}  & \underline{93.92}  & \underline{47.37}       & \underline{54.33}     \\
\ding{51}                 & \ding{51}             & \ding{51}                       & \textbf{95.05} & \textbf{94.20} & \textbf{47.60}      & \textbf{54.70}    \\
\bottomrule
\end{tabular}
}
\end{table*}

\subsubsection{Effectiveness of Multi-modal Pre-training}
To assess the benefits of RingMoE's multi-modal pre-training, we construct uni-modal RingMoE variants and conduct comparative experiments on both uni-modal and multi-modal tasks. These variants are pre-trained solely on optical or SAR data and activate only their corresponding modal-specialized experts. Results illustrated in Fig.\ref{fig:13}.c, reveal that multi-modal pre-training not only boosts performance on multi-modal tasks, such as segmentation on the DFC23 dataset, but also enhances uni-modal task performance by enriching modality-specific representations. This underscores the effectiveness of RingMoE's ``division of labor and cooperation'' MoE structure, which captures intra-modal features while integrating multi-modal information. Such adaptability establishes RingMoE as a robust solution for diverse RS applications demanding comprehensive and accurate multi-modal analysis.

\subsubsection{Effectiveness of Scaling Foundation Models to 10 Billion-scale Parameters}
{To investigate the effectiveness of scaling RSFMs to the 10B-parameter level, we conduct a series of ablation studies. Specifically, we pre-train a 1B-parameter Swin Transformer v2 baseline (without the RMoE layer) using the SimMIM~\cite{xie2022simmim} self-supervised framework on the same optical datasets as RingMoE. As illustrated in Fig.\ref{fig:13}.d, the full 14.7B RingMoE model (with 6.5B parameters allocated per modality) consistently outperforms the 1B baseline across multiple downstream tasks, demonstrating the advantages of scaling through sparse expert modeling. Furthermore, even the 1B dense variant, RingMoE-KC, surpasses the original Swin baseline, highlighting the effectiveness of expert pruning strategy in retaining high-capacity knowledge.}

{To further validate the impact of parameter scaling, we replace the FFN layers in the Swin-B (88M) backbone with RMoE layers, forming a lightweight variant (RingMoE@Swin-88M), and compare it against the original RingMoE@Swin-1B. Both models share identical settings, differing only in backbone size (88M vs. 1B). As reported in Tab.\ref{tab:8}, RingMoE@Swin-1B consistently outperforms RingMoE@Swin-88M on the NWPU-RESISC45~\cite{cheng2017remote} and AID~\cite{xia2017aid} datasets, providing rigorous evidence that increasing parameters alone yields significant performance gains.In summary, these findings confirm the necessity and effectiveness of parameter scaling for enhancing the representational capacity of RSFMs.}

\subsubsection{Ablation Studies on Model Components}

{\textbf{Expert Design.} 
We conduct ablation experiments by selectively removing each expert type from RingMoE. As reported in Tab.\ref{tab:9}, removing modal-specialized experts leads to notable performance drops in single-modal tasks (-1.27\% accuracy on optical classification, -1.32\% mIoU on multi-spectral segmentation), highlighting their effectiveness in capturing modal-specific knowledge. Excluding collaborative experts leads to a 2.07\% mIoU decline on multi-modal segmentation, underscoring their role in facilitating cross-modal representation learning. Removing the shared expert consistently causes minor performance declines across all tasks, indicating its stabilizing role in maintaining globally transferable features. These results collectively demonstrate the complementary benefits of the three expert types and underscore the necessity of our hybrid expert design for robust multimodal learning.}

\noindent {\textbf{Physics-informed Loss Function.} 
To evaluate the impact of incorporating SAR physical priors, we replace the proposed physics-informed reconstruction loss with a pixel-wise reconstruction loss during SAR-L1 pre-training. As reported in Appendix Tab.18, this substitution leads to consistent performance degradation on both SAR-L1 and SAR-L2 downstream tasks (–5.67\% mIoU on SARSegL1, –3.14\% mAP on SAR-AIRCraft). Notably, although the physics-informed loss is applied only to SAR-L1 data during pretraining, performance gains are also observed on SAR-L2 tasks. This improvement stems from collaborative experts in RMoE enabling implicit cross-modal knowledge transfer. These findings indicate that embedding physics-based constraints effectively enhances both modal-specific representation learning and generalization across SAR branches.}

\section{Limitations and Future Work}
{Despite the strong performance of RingMoE, several limitations remain.
First, the pre-training corpus consists of unpaired multi-modal RS imagery, limiting the use of cross-modal self-supervised paradigms such as contrastive learning.
Second, the RingMOSS dataset suffers from regional and modality imbalance, with data predominantly sourced from a few countries.
Third, while expert pruning enables efficient deployment, pre-training the full-scale RingMoE remains computationally intensive.
Fourth, the benefits of multi-modal fusion are only verified on two benchmarks due to the scarcity of geo-aligned multi-modal RS datasets. To address these limitations, we plan to curate large-scale, geographically diverse, and geo-registered multi-modal RS datasets to support fine-grained cross-modal learning and enhance generalization.
We will also explore further model compression via knowledge distillation to enable deployment on resource-limited platforms (e.g., UAVs).
Moreover, RingMoE will be extended as a universal vision encoder for RS multi-modal large language models and integrated with prompt-based frameworks such as SAM to enhance interactivity and generality in referential perception tasks.}

\section{Conclusion}
This paper presents RingMoE, a 14.7-billion-parameter multi-modal foundation model that achieves SOTA performance on 23 out of 25 public benchmarks across six key RS tasks. Pre-trained on 400 million multi-modal images, including Opt, MS, and SAR data, RingMoE demonstrates strong generalization and robust few-shot performance, making it particularly effective in data-scarce scenarios. In addition to accuracy, RingMoE's modular design allows for flexible adaptation to diverse downstream tasks through expert pruning, maintaining competitive performance even in compressed versions. These results underscore the potential of large-scale multi-modal foundation models in RS, with promising applications in disaster response, environmental monitoring, and urban planning. 


%



\ifCLASSOPTIONcompsoc
\section*{Acknowledgments}
\else
\fi

This work was supported by the National Natural Science Foundation of China (NSFC) under Grant 62301538.


\ifCLASSOPTIONcaptionsoff
\newpage
\fi



%






\appendices

\setcounter{table}{0}
\setcounter{figure}{0}
\renewcommand{\thetable}{A\arabic{table}}
\renewcommand{\thefigure}{A\arabic{figure}}
\makeatletter
\renewcommand{\theHfigure}{appendix.A.\arabic{figure}} 
\renewcommand{\theHtable}{appendix.A.\arabic{table}}   
\makeatother

\section*{Appendix Overview}

The Supplementary materials offer a detailed extension of the main body, organized into four sections:
\begin{itemize}
\item \textbf{Specific architecture of MoE (Appendix \ref{secB})}: This section provides a detailed explanation of the Mixture-of-Experts (MoE) architecture. It focuses on the efficiency of MoE in scaling models while maintaining computational feasibility, achieved through a sparse gating mechanism that optimizes expert utilization and reduces computational load.
\item \textbf{RingMoE implementation (Appendix \ref{secC})}: This section provides a detailed discussion of the design and core components of the RingMoE model. {It covers the collection and pre-processing of the pre-training dataset, the architectural details of the model, the specifics of the pre-training process, and the implementation of expert pruning.} Furthermore, it details the implementation of RingMoE for downstream tasks, including dataset descriptions, experimental setups, and evaluation metrics.
\item \textbf{More detailed performance comparison (Appendix \ref{secD})}: This section extends the performance results presented in the main body by comparing a broader range of existing methods in various benchmarks, offering a more comprehensive analysis of the RingMoE model's performance.
\item {\textbf{Comparison with popular MLLMs (Appendix \ref{secE})}: This section compares the proposed RingMoE model with recent Multimodal Large Language Models (MLLMs) in RS scenarios. The evaluation focuses on task specialization, modality generalization, and performance.}
\item {\textbf{More comprehensive ablation studies and analysis (Appendix \ref{secF})}: This section conducts additional ablation analyses, covering the comparison of expert pruning strategies, the effect of SAR-L1 pretraining on improving SAR understanding, and the impact of the physics-informed loss applied during SAR-L1 pretraining.}
\end{itemize}

\section{Specific architecture of MoE}\label{secB}

A typical case to construct a Mixture-of-Experts (MoE) language model usually replaces the FFNs in the transformer with MoE Layers. The MoE layer consists of multiple experts, i.e., $E_1, E_2, ..., E_N$, each of which is structurally identical to a standard FFN. Then, each input token $x \in \mathbb{R}{^{d}}$ will be assigned one or more experts through a routing network $G$. The output $y$ of a MoE layer is the weighted sum of the outputs $E_k(x)$ from the assigned experts, which can be formulated as:
\begin{align}  \label{equation:1}
\setlength{\abovecaptionskip}{1pt}
\setlength{\belowcaptionskip}{1pt}
y = \sum_{k=1}^{N}G_k\left ( x \right )  E_k\left ( x \right ) ,k=1,2,...N
\end{align}
Where $N$ denotes the number of experts, $E_k$ denotes the $k^{th}$ expert, and $G_k$ denotes the weight of the $k^{th}$ expert derived through the routing network $G$. 

{The routing network $G$ is a Noisy Top-K Gating network with trainable weight matrix $W_g \in \mathbb{R}{^{d \times N}}$ and $W_{noise} \in \mathbb{R}{^{d \times N}}$, which models the probability $P\left( E_k|x\right)$ of using different experts for the current token and selects only Top-K experts for the final output. This sparse gating mechanism enables a further increase in model capacity with limited computational cost. The specific process can be formulated as: 
\begin{align}  \label{equation:2}
\setlength{\abovecaptionskip}{1pt}
\setlength{\belowcaptionskip}{1pt}
G\left ( x \right ) =\mathrm{TopK}\left ( H\left ( x \right ), K  \right )
\end{align}
\begin{equation}\label{equation:3}
\setlength{\abovecaptionskip}{1pt}
\setlength{\belowcaptionskip}{1pt}
H\left ( x \right )=\mathrm{Softmax}( x W_g+N\left ( 0,1 \right )\mathrm{Softplus}\left ( x W_{noise} \right )  )  
\end{equation}
Where $\mathrm{TopK}\left( \cdot, K \right)$ keeps only the top $K$ values and sets the rest to zero. The noise term helps load balancing to encourage experts to receive roughly equal numbers of training samples, where $\mathrm{Softplus}\left(x\right)=\log \left ( 1+\exp\left ( x \right )   \right )$ is the smooth approximation to the $\mathrm{ReLU}$ function.}

\begin{figure*}[t]
\setlength{\abovecaptionskip}{1pt}
\setlength{\belowcaptionskip}{1pt}
\centering
\includegraphics[width=0.95\linewidth]{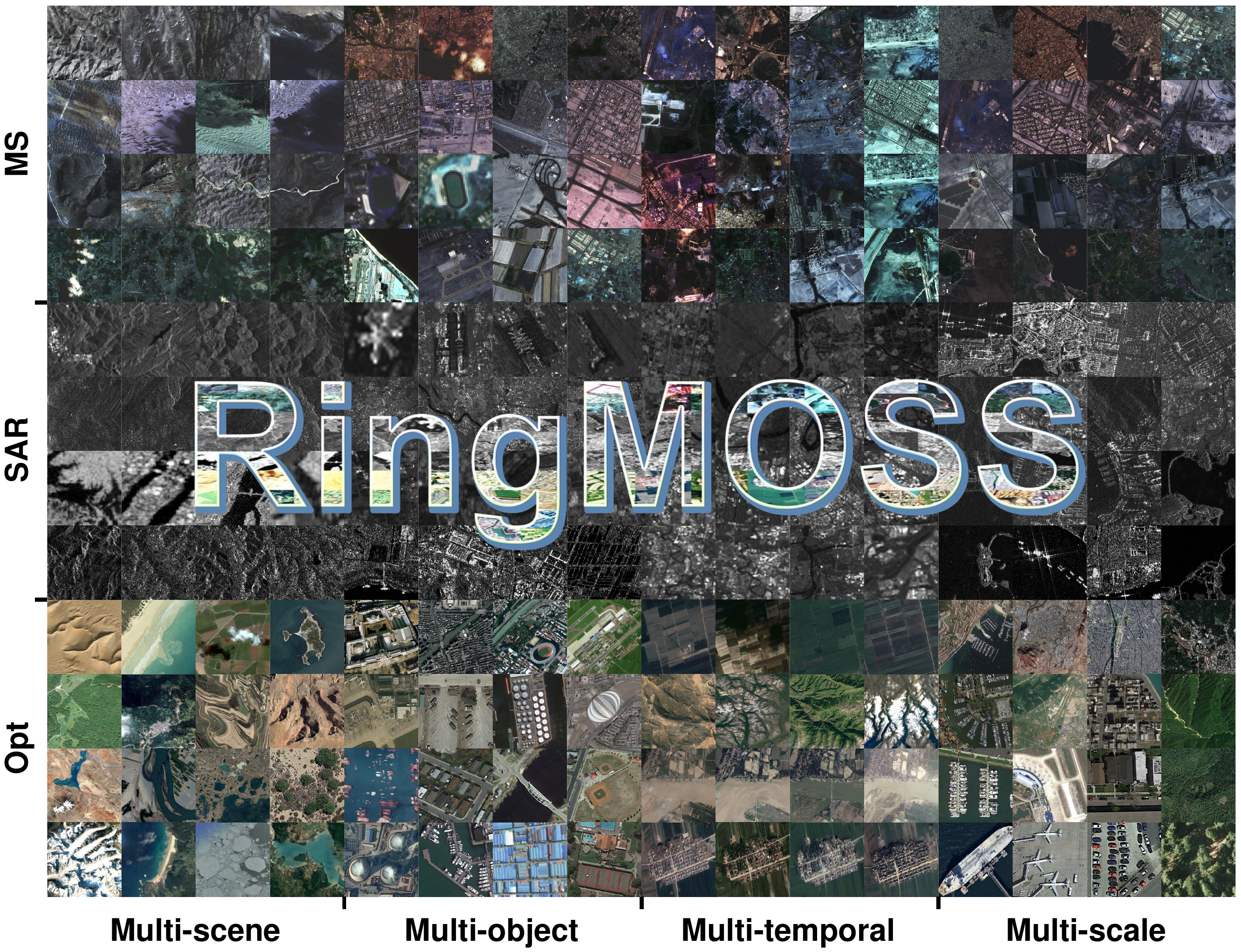}
\caption{RingMOSS includes various terrains, geographic regions, and scene complexities, as well as geospatial elements at different temporal and spatial scales, ensuring comprehensive global coverage.}
\label{fig:8_1}
\end{figure*}

\begin{figure*}[t]
\setlength{\abovecaptionskip}{1pt}
\setlength{\belowcaptionskip}{1pt}
\centering
\includegraphics[width=0.95\linewidth]{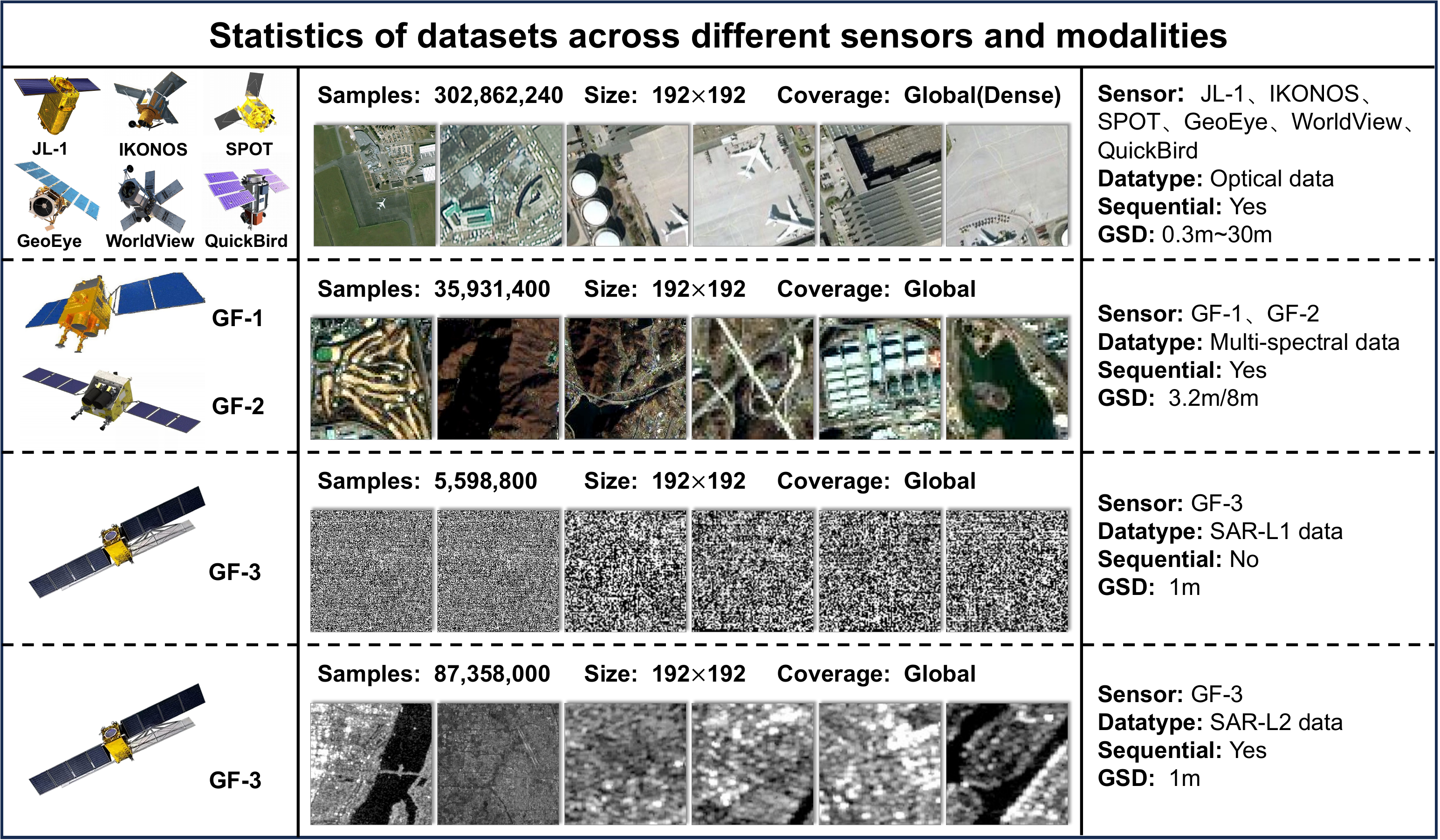}
\caption{\textbf{Overview of the RingMoE pre-training dataset, i.e., RingMOSS.} The dataset integrates Opt, MS, and SAR imagery. Opt data is sourced from high-resolution satellites (JL-1, GF-2, GeoEye, WorldView, etc.), with resolutions from 0.3m to 30m. MS images, covering blue, green, red, and near-infrared bands, come from GF-1 and GF-2 (resolutions of 8m and 3.2m). SAR data from GF-3 includes SAR-L1 (1m resolution, with four polarization modes) and SAR-L2 (amplitude form), with additional geo-coding and calibration for pre-training.}
\label{fig:9_1}
\end{figure*}

\begin{figure*}[t]
\setlength{\abovecaptionskip}{1pt}
\setlength{\belowcaptionskip}{1pt}
\centering
\includegraphics[width=0.95\linewidth]{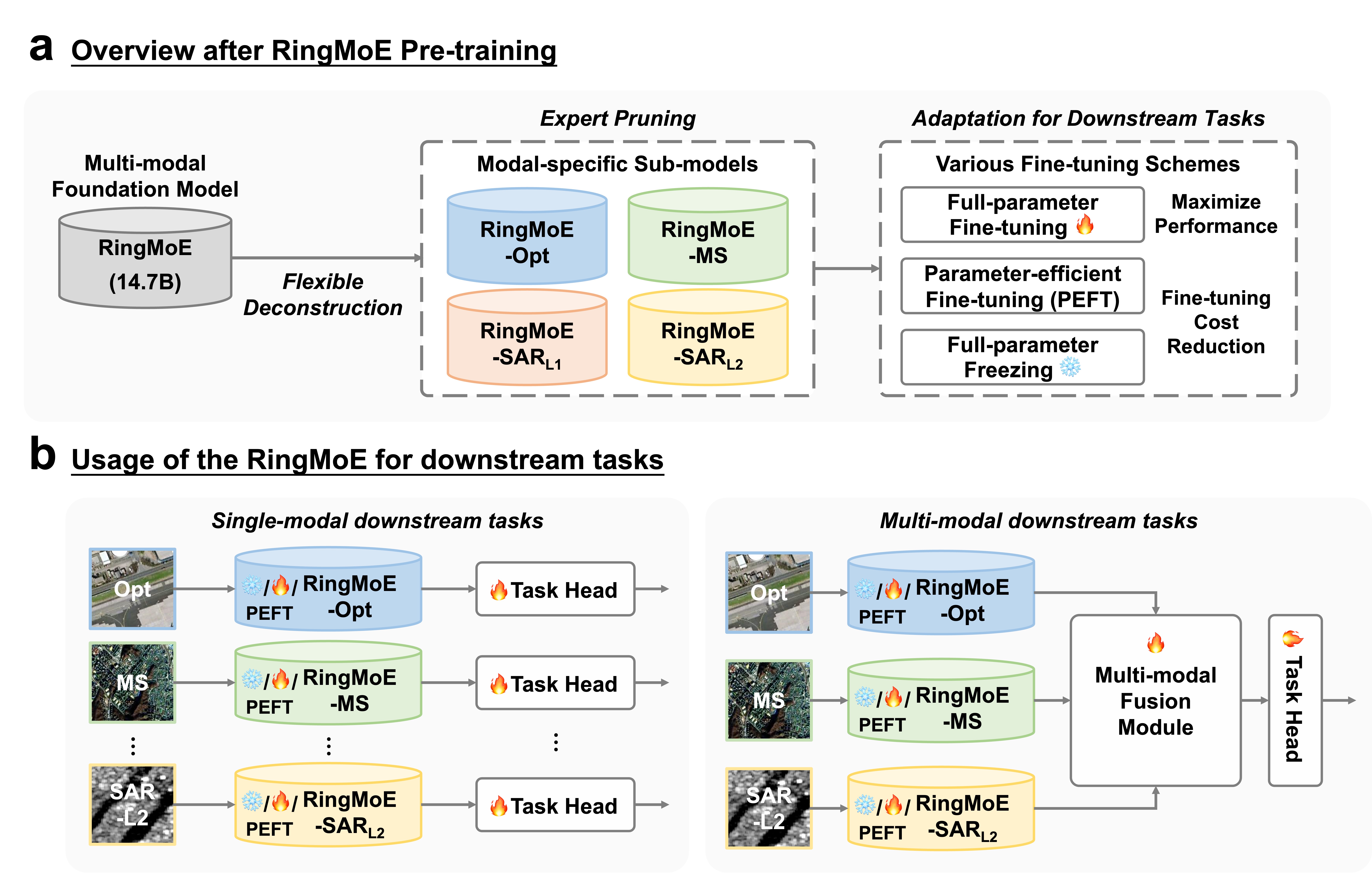}
\caption{\textbf{Details after RingMoE Pre-training}. \textbf{a.} After pre-training, RingMoE supports modular decomposition into single-modal or multi-modal models of varying scales via expert pruning, enabling flexible adaptation to downstream tasks.  \textbf{b. Usage of RingMoE in downstream:} Modal-specific models can be deployed individually or jointly, with fine-tuning options including full tuning, parameter freezing, or parameter-efficient fine-tuning (PEFT) for optimal accuracy-resource trade-offs.}
\label{fig:10}
\end{figure*}

\section{RingMoE implementation}\label{secC}

\subsection{Collection and Pre-processing of RingMoE Pre-training Dataset}\label{secC_data}
The RingMOSS pre-training dataset is designed to provide a comprehensive and multi-dimensional analysis of RS data, incorporating diverse data sources, including optical (Opt), multi-spectral (MS), and Synthetic Aperture Radar (SAR) imagery (see Fig.\ref{fig:8_1}-\ref{fig:9_1}). Opt data is sourced from high-resolution aerial and satellite platforms, such as JL-1, GF-2, GeoEye, WorldView, QuickBird, IKONOS, and SPOT, with spatial resolutions ranging from 0.3 meters to 30 meters. MS images, capturing four bands—blue, green, red, and near-infrared—are obtained from the GF-1 and GF-2 satellites. The GF-1 PMS camera provides images at a resolution of 8 meters, while the GF-2 PMS camera delivers higher-resolution images at 3.2 meters. SAR data is sourced from the GF-3 satellite, encompassing both SAR-L1 (complex-valued form) and SAR-L2 (amplitude form) images. The SAR-L1 images, with a resolution of 1 meter, include four polarization modes, HH, HV, VH, and VV, combined into eight-channel images for pre-training. These images capture both amplitude and phase information, providing high-resolution insights into Earth's surface. SAR-L2 images, derived from SAR-L1 data, undergo additional processing, including geo-coding and radiometric calibration.

The RingMOSS dataset is constructed by aggregating RS images from both public and proprietary sources, with a focus on ensuring cross-regional coverage, temporal variability, and diversity in scenes and objects. It spans urban, agricultural, forest, water, and mountainous regions, thereby providing broad representational coverage. During the pre-processing stage, all original RS images are partitioned into 192$\times$192 pixel tiles using a sliding window technique with a step size of 64 pixels. Due to edge effects, sensor artifacts, and data stitching issues, black borders often appear along the image edges. To prevent these artifacts from affecting model training, any tile containing more than 30\% black border is discarded, leaving only valid regions for further processing and training. Additionally, regions affected by cloud cover are automatically detected and removed during this stage. These steps ensure the high quality and integrity of the dataset, providing reliable input for subsequent model training and analysis.

{Due to licensing restrictions and confidentiality agreements, we are unable to release data from private sources. However, to support reproducibility and community research, we have open-sourced a curated subset of RingMOSS\footnote[2]{{All public data, along with detailed documentation are available at:
\href{https://github.com/HanboBizl/RingMoEDatasets}{https://github.com/HanboBizl/RingMoEDatasets}.}} that is fully derived from public sources. This released subset retains all four modalities and follows the exact same preprocessing and formatting protocols as the original dataset used for pretraining.}

\subsection{Details for Network Architecture}\label{secC_1}

The RingMoE foundation model developed in this study builds on the 1-billion-parameter Swin Transformer v2 architecture~\cite{liu2022swin}, which is structured into four stages containing 2, 2, 18, and 2 blocks, respectively. Each stage consists of feature channels with dimensions of 448, 896, 1792, and 3584. Within each block, Multi-head Attention (MHA), Feed-forward Networks (FFN), and Layer Normalization (LayerNorm) are employed. In our RingMoE model, we enhance the standard architecture by replacing each FFN layer with the proposed RMoE layer to improve the model's capacity for capturing intricate multi-modal interactions. The RMoE layer integrates three types of experts: modal-specialized experts, collaborative experts, and a shared expert. Each modality is assigned four modal-specialized experts (i.e., standard FFN), controlled by a routing network, while the collaborative experts comprise four experts, also governed by a routing network. The shared expert remains as a standard FFN layer. The final output of the RMoE layer is derived by summing the contributions from the modal-specialized, collaborative, and shared experts, providing a more efficient mechanism for multi-modal learning.

\subsection{Implementation for Pre-training}\label{secC_3}
For pre-training, all multi-modal images adopt the following pre-processing settings, including randomly cropping the image in the range of 0.2$\times$-1.0$\times$ of the original size, resizing to 192$\times$192, applying horizontal flipping and normalization operations. For Opt, MS, and SAR-L2 images, they are input as 8-bit data formats with 3 channels, 4 channels, and 1 channel, respectively. For SAR-L1 images, which consist of four polarization modes and eight channels, we maintain the 16-bit format to maximize the preservation of complex signal information. 

The RingMoE model was implemented in the MindSpore framework and deployed on a cluster of 512 Ascend 910 AI processors for training, inference, and testing. To enable efficient large-scale training, MindSpore's Auto-parallel system was employed, utilizing four dimensions of parallelism: data parallelism, operator-level model parallelism, optimizer parallelism, and materialization. This strategy significantly accelerated training while optimizing memory utilization and computational load distribution. The AdamW optimizer was used with a weight decay of 0.05, and the learning rate was initialized at 2e-4.


\subsection{Implementation for Expert Pruning}

\subsubsection{Sparse Expert Pruning Schemes} 
{
Although we introduce a load-balancing loss (Eq.(4) in the main text) to encourage uniform expert activation, we consistently observe an uneven distribution of routing decisions across cooperative experts, often influenced by the input modality. Specifically, different modalities tend to activate distinct subsets of cooperative experts in each RMoE layer, leading to modality-dependent expert preferences despite the balancing regularization. To quantify this phenomenon, we compute the modal-specific activation frequency of each cooperative expert $E_k^C$ for modality $m$, formally defined as:
\begin{align}  \label{equation:1_1}
\setlength{\abovecaptionskip}{1pt}
\setlength{\belowcaptionskip}{1pt}
f(E_k^C|x_m) = \frac{1}{N_m} \sum_{i=1}^{N_m} \mathbb{I}\left[G^C(x_{m;i}) = E_k^C\right]
\end{align}
Where $N_m$ denotes the total number of tokens from modality $m$, and $x_{m;i}$ is the $i^{th}$ token. The indicator function $\mathbb{I}(\cdot)$ equals 1 if expert $E_k^C$ is selected for the token $x_{m;i}$, and 0 otherwise. Here, $k \in {1, 2, …, N_C}$ indexes the cooperative experts, with $N_C$ representing the total number of cooperative experts in each RMoE layer. This frequency metric captures the degree to which each expert is utilized when processing data from different modalities.}

{We consistently observe a long-tailed distribution: a small number of experts dominate the routing decisions for each modality, while many others remain underutilized. Moreover, these preferences show consistent modality-specific patterns:
\begin{itemize}
    \item Optical and multi-spectral modalities tend to activate similar sets of experts, reflecting their shared spectral characteristics.
    \item SAR-L1 and SAR-L2 also show close alignment in expert selection, though SAR-L1 activates a broader range of experts, potentially due to its richer spatial structure.
    \item In contrast, optical and SAR modalities activate almost disjoint expert subsets, which is consistent with their fundamentally different imaging mechanisms (e.g., passive reflection vs. active microwave emission).
\end{itemize}}

{These observations motivate a modality-aware, layer-wise sparse expert pruning strategy, designed to eliminate those cooperative experts that contribute little to the representation learning of any modality. For each RMoE layer and modality m, we rank all cooperative experts by their activation frequency $f(E_k^C|x_m)$, then retain only the top 25\% most frequently activated ones. The pruning threshold $\varphi$ is determined as the 75$^{th}$ percentile of the activation frequency distribution:
\begin{align}  \label{equation:2_1}
\setlength{\abovecaptionskip}{1pt}
\setlength{\belowcaptionskip}{1pt}
\varphi = \text{Percentile}_{75} \left( \left\{ f(E_k^C|x_m) \right\}_{k=1}^{N_C} \right)
\end{align}
Experts with $f(E_k^C|x_m) < \varphi$ are deemed underutilized and are pruned for modality m. After computing $\varphi$ for all modalities, we take the union of the retained experts across modalities at each layer. This ensures that experts important to any single modality are preserved, while those consistently underused are pruned.}

{We validate the pruning design on the NPWU classification dataset using our 6.5B-parameter RingMoE model. As shown in Tab.\ref{tab:6}, we compare both fixed-threshold and percentile-based strategies under various retention ratios. The Top-25\% scheme, which retains the most frequently activated 25\% of cooperative experts for each modality and layer, achieves the best trade-off between compression and performance, reducing model size from 6.5B to 4.3B parameters with only a 0.47\% drop in classification accuracy.}

\begin{table}[t]
\setlength{\abovecaptionskip}{1pt}
\setlength{\belowcaptionskip}{1pt}
\caption{{Ablation study on the pruning threshold $\varphi$ selection. We evaluate both fixed and percentile-based pruning strategies using the 6.5B RingMoE-Opt (optical modality) model on the NPWU classification dataset.}} \label{tab:6}
\renewcommand\arraystretch{1.25}
\centering
\resizebox{0.7\linewidth}{!}{
\begin{tabular}{ccc} 
\toprule
Selection Strategy                   & \#Params      & Acc(\%)         \\ 
\hline
Fixed $\varphi=0.1$ & 4.7B          & 95.06           \\
\textbf{Top 25\%}                    & \textbf{4.3B} & \textbf{95.43}  \\
Top 50\%                             & 5.0B          & 95.57           \\
Top 75\%                             & 5.7B          & 95.71           \\
No Pruning                           & 6.5B          & 95.90           \\
\bottomrule
\end{tabular}
}
\end{table}

\subsubsection{Dense Expert Integration Schemes} 
{We also explore compressing the sparse RMoE model into a dense format. This involves discarding the expert selection mechanism from the routing network $G$ and consolidating all expert knowledge into a single expert, essentially converting the structure back into a standard feed-forward network (FFN) layer, reducing the total parameters to 1 billion. We propose three schemes for knowledge aggregation: summing, averaging, and compressing, each providing different strategies for integrating the expert knowledge into a unified representation.}

\noindent {\textbf{Knowledge summing (KS) and averaging (KA).} For Knowledge Summing, given two linear projections $W_{k;1}^{*} \in \mathbb{R}{^{d_1\times d_2}}$ and $W_{k;2}^{*} \in \mathbb{R}{^{d_2\times d_1}}$ in every expert (modal-specialized, cooperative, and shared) in the RMoE layer, we sum the weights of all experts to aggregate their knowledge:
\begin{equation} \label{equation:15}
\setlength{\abovecaptionskip}{0pt}
\setlength{\belowcaptionskip}{0pt}
\begin{split}
W_1 & = \sum_{k=1}^{N_S} W_{k;1}^{S}+\sum_{k=1}^{N_C} W_{k;1}^C+W_1^{Shared}    \\
W_2 &=  \sum_{k=1}^{N_S} W_{k;2}^{S}+\sum_{k=1}^{N_C} W_{k;2}^C+W_2^{Shared}
\end{split}
\end{equation}
Where $W_1 \in \mathbb{R}{^{d_1\times d_2}}$ and $W_2 \in \mathbb{R}{^{d_2\times d_1}}$ denote the linear projections after summing the expert knowledge. $W^S_{*}$, $W^C_{*}$, and $W^{Shared}_{*}$ denote the linear projections from modal-specialized, collaborative, and the shared experts, respectively. For Knowledge Averaging, similar to summing, we average the weights of the experts.}

\noindent {\textbf{Knowledge Compressing.}
Beyond the two aforementioned schemes of direct expert knowledge aggregation, we aim to preserve key knowledge from each expert while ensuring effective compression. Low-rank decomposition is a powerful technique for model compression, enabling the transformation of a high-dimensional matrix into a low-rank representation while maximally retaining its information. To achieve this, we compress each expert’s weight matrix into a low-rank form \cite{chen2023mod} and subsequently merge them into a high-rank matrix with the same dimensionality as the original, as illustrated in Fig.7 of the main text. }

{Specifically, we apply Singular Value Decomposition (SVD) \cite{kalman1996singularly} to compress the weight matrices of modal-specialized and collaborative experts, preserving essential features while reducing redundancy. Taking the first linear projection layer of the experts as an example, we decompose its weight matrix using SVD, ensuring that the compressed representation retains the critical knowledge necessary for effective expert merging:
\begin{equation} 
\setlength{\abovecaptionskip}{0pt}
\setlength{\belowcaptionskip}{0pt}
\begin{split}
W_{d_1\times d_2}&= U_{d_1\times d_1}\Sigma_{d_1\times d_2}V_{d_2\times d_2}^T \\ &\approx U_{d_1\times K}\Sigma_{K\times K}V_{K\times d_2}^T
\end{split}
\end{equation}
Where $U_{d_1\times d_1}$ and $V_{d_2\times d_2}$ are unitary matrices and $\Sigma_{d_1\times d_2}$ is a diagonal matrix with non-zero elements only on its main diagonal (i.e., singular values). We retain the top K singular values from $\Sigma_{d_1\times d_2}$ to construct the low-rank approximation, forming $U_{d_1\times K}$ and $V_{K\times d_2}$. Notably, for modal-specialized experts, $K=\frac{d_2}{N_S}$ and for collaborative experts, $K=\frac{d_2}{N_C} $. By concatenating these low-rank matrices across all expert types, a new weight matrix is generated:
\begin{align}
\setlength{\abovecaptionskip}{0pt}
\setlength{\belowcaptionskip}{0pt}
M_1^* =\left [U_{d_1\times K},...,U_{d_1\times K}  \right ] \begin{bmatrix}
  \Sigma_{K\times K}&  & \\
  & ... & \\
  &  &\Sigma_{K\times K} 
\end{bmatrix} \begin{bmatrix}
 V_{K\times d_2}\\
 ...\\
V_{K\times d_2}
\end{bmatrix}
\end{align}
Where $M_1^* \in \mathbb{R}{^{d_1\times d_2}}$ denotes the new weight matrix capturing the essential knowledge from all experts. The final linear projection $W_1 \in \mathbb{R}{^{d_1\times d_2}}$ is then formulated as:
\begin{align} 
\setlength{\abovecaptionskip}{0pt}
\setlength{\belowcaptionskip}{0pt}
W_1 = M_1^S+M_1^C+W_1^{Shared}
\end{align}
Another linear projection $W_2 \in \mathbb{R}{^{d_2\times d_1}}$ has a similar process.}

\subsection{Implementation for Downstream Tasks}\label{secC_5}
This section outlines the datasets, experimental setups, and evaluation metrics employed across various downstream tasks, providing a comprehensive foundation for assessing model performance.

\subsubsection{Scene Classification}\label{Scene classification}

Scene classification serves as a core task in remote sensing interpretation, enabling automatic identification of land cover types and scene categories, with applications in environmental monitoring and land management. For evaluation, we select two widely recognized remote sensing scene classification datasets.

\noindent \textbf{(1) Dataset Introduction.}
\begin{itemize}
\item \textbf{NWPU-RESISC45 (Opt)~\cite{cheng2017remote}}, is a publicly available remote sensing image scene classification (RESISC) dataset containing 31,500 images of size 256$\times$256, with ground sample distances (GSD) ranging from 0.2 meters to 30 meters. It covers 45 diverse scene classes including ``Aircraft'', ``Airports'', ``Baseball diamond'', ``Basketball court'', ``Beach'', ``Bridge'', ``Jungle'', ``Church'', ``Circular farmland'', ``Cloud'', ``Commercial area'', ``Dense housing'', ``Desert'', etc., with 700 images in each category. Following previous work~\cite{sun2022ringmo,guo2023skysense}, we allocate 10\% of the dataset for training and reserve the remaining 90\% for testing.
\item \textbf{AID (Opt)~\cite{xia2017aid}}, is a large-scale dataset of aerial imagery sourced from Google Earth, containing 10,000 images of size 600$\times$600, with GSD ranging from 0.5 meters to 8 meters. This dataset spans 30 distinct aerial scene classes, including ``Airport'', ``Bare ground'', ``Baseball stadium'', ``Beach'', ``Bridge'', ``Center'', ``Church'', ``Business'', ``Dense housing'', etc., with 220 to 400 images in each category. Following previous work~\cite{sun2022ringmo,guo2023skysense}, we allocate 20\% of the dataset for training and reserve the remaining 80\% for testing.
\end{itemize}

\noindent \textbf{(2) Implementation Detail.}
Our experiments are performed in the mmpretrain framework\footnote[1]{\href{https://github.com/open-mmlab/mmpretrain}{https://github.com/open-mmlab/mmpretrain}}. RingMoE is employed to extract features while a linear mapping layer is employed for classification. The optimizer utilizes AdamW with an initial learning rate of 5e-5 and a weight decay of 0.05. The learning rate scheduler consists of two phases: a linear scheduler for the first 10 epochs with an initial factor of 0.01, followed by a cosine scheduler starting from the 10th epoch with a minimum learning rate of 5e-6. The training process continues for a total of 200 epochs. For NWPU-RESISC45, the images are randomly cropped to 224$\times$224 and for AID, the images are randomly cropped to 512$\times$512. To enhance model robustness, we implement data augmentation techniques including RandomCrop, RandomFlip, and RandomErasing for both datasets.

\noindent \textbf{(3) Evaluation Metric.}
We use overall accuracy (OA), a widely adopted metric in classification tasks, to evaluate the performance of remote sensing scene classification models. Defined as the ratio of correctly classified instances to the total instances in the dataset, OA directly reflects the model's prediction success. Formally, OA can be expressed as: $\text{OA} = \frac{\text{TP}+\text{TN}}{{\text{TP}+\text{TN}+\text{FP}+\text{FN}}}$, where $\text{TP}$ and $\text{TN}$ denote the number of true positive and true negative predictions, respectively, while $\text{FP}$ and $\text{FN}$ represent the false positive and false negative predictions, respectively.

\subsubsection{Semantic Segmentation}
Semantic segmentation is a core task in remote sensing interpretation, aiming to assign each pixel in an image to a specific ground object category, thereby delivering detailed spatial information. To thoroughly evaluate RingMoE's interpretive capabilities across various modalities, we select five widely used semantic segmentation datasets covering optical, multi-spectral, and SAR (complex-valued and amplitude form) for evaluation.

\noindent \textbf{(1) Dataset Introduction.}
\begin{itemize}
\item \textbf{iSAID (Opt) \cite{waqas2019isaid}}, is an instance-level semantic segmentation dataset consisting of 2806 aerial images captured by various satellite sensors. The Image resolutions range from 800$\times$800 to 4000$\times$13000. It includes 655,451 instances across 15 object categories, with non-object pixels labeled as background, totaling 16 categories. The dataset is split into training, validation, and testing sets, comprising 1411, 458, and 937 samples. Following \cite{sun2022ringmo,guo2023skysense}, we evaluate the model performance on the validation set. 

\item \textbf{ISPRS Potsdam (Opt)~\cite{isprs2018semanticlabeling}}, is a scene-level semantic segmentation dataset that contains 38 high-resolution aerial images with a ground sampling distance (GSD) of 0.05 meters, each with a fixed size of 6000$\times$6000 pixels. Following the setup in RingMo~\cite{sun2022ringmo}, we conduct experiments using images that include near-infrared, red, and green spectral bands, focusing on five categories: ``Impervious surface'', ``Building'', ``Low vegetation'', ``Tree'', and ``Car''. The dataset is split into 24 images for training and 14 for testing.

\item \textbf{Dyna.-pla (MS) \cite{toker2022dynamicearthnet}}, is a multi-spectral scene-level semantic segmentation dataset captured from the PlanetFusion satellite. It includes images from 75 different locations, with each featuring 24 images taken at different times, accompanied by annotations for 7 land use and land cover semantic classes. Each image is captured in four bands (Red, Green, Blue, and Near-Infrared), with a ground sampling distance (GSD) of 3 meters and an image size of 1024$\times$1024. Following the official leaderboard configuration, the dataset is divided into 55 locations for training, 10 for validation, and 10 for testing. In our experiments, we utilize the official validation and test sets for performance evaluation.

\item \textbf{SARSegL1 (SAR-L1) \cite{DENet(SARsegL1)}}, is a fully polarized semantic segmentation dataset, captured by the GF3 satellite over Guangzhou and Hangzhou, China, at a spatial resolution of 5 meters. It includes two images with dimensions of 5456$\times$4708 and 6192$\times$4888 pixels, respectively, with each image comprising four polarization modes: HH, HV, VH, and VV. Each mode has two channels that capture amplitude and phase information. The dataset is annotated with five categories: ``Water'', ``Vegetation'', ``Bare ground'', ``Building'', and ``Road'' annotated by professionals. To facilitate model training and evaluation, the images are cropped into 500$\times$500 pixels with an overlap of 200, resulting in a total of 540 samples, with an 8:2 split between training and testing sets.

\item \textbf{Air-POLSAR-SEG (SAR-L2) \cite{AIR-PolSAR-Seg}}, is a scene-level semantic segmentation dataset, contains 500 single-channel SAR images (in amplitude form) with a spatial resolution of 8 meters. Each 512$\times$512 pixel image provides pixel-level annotations across six categories: ``Housing area'', ``Industrial area'', ``Natural area'', ``Land-use area'', ``Water'', and ``Other area''. The dataset is divided into training and test sets in a 7:3 ratio, enabling precise land-use classification tasks in remote sensing applications. Due to the small proportion of Land-use and Other areas, these two categories are excluded from the experimental results.
\end{itemize}

\noindent \textbf{(2) Implementation Detail.}
All semantic segmentation downstream tasks are implemented using the mmsegmentation framework\footnote[2]{\href{https://github.com/open-mmlab/mmsegmentation}{https://github.com/open-mmlab/mmsegmentation}}. We adopt Upernet \cite{Upernet} as the segmentation framework and utilize compressed RingMoE-KC as the backbone for feature extraction. The optimizer is AdamW with an initial learning rate of 1e-4 and a weight decay of 0.05. The learning rate schedule consists of two phases: a linear warm-up for the first 1,500 iterations with a factor of 0.1, followed by a polynomial scheduler with a minimum learning rate of 1e-6. For the iSAID dataset, images are cropped to 896$\times$896, and the model is trained for 80k iterations with a batch size of 8. For the Potsdam and Air-POLSAR-SEG datasets, images are cropped to 512$\times$512, with training also set at 80k iterations and a batch size of 8. For the Dyna.-pla dataset, images are resized to 1024$\times$1024, with training set at 16k iterations and a batch size of 8. For the SARSegL1 dataset, the two-channel data from four polarization modes are concatenated to create an 8-channel sample, with images cropped to 500$\times$500 and trained for 20k iterations with a batch size of 4. In our experiments, we employ regular data augmentation strategies, including RandomFlip, RandomScaling (from 0.5 to 2.0), and RandomCrop.

\noindent \textbf{(3) Evaluation Metric.}
We employ mean intersection over union (mIoU), overall accuracy (OA), mean accuracy (mAcc), and mean F1 score (mF1) to evaluate segmentation performance. mIoU calculates the segmentation accuracy per category by taking the ratio of the intersection-over-union (IoU) between predicted and true categories and then averaging across all categories to reflect the model's capability in segmenting individual classes. For each category, the IoU can be calculated by $\text{IoU} = \frac{\text{TP}}{{\text{TP}+\text{FP}+\text{FN}}}$. Meanwhile, OA evaluates the percentage of correctly classified pixels across the dataset, providing a straightforward view of overall model accuracy. mAcc measures the average pixel accuracy for each category, defined as the ratio of correctly predicted pixels in a category to the total ground truth pixels of that category. For each category, accuracy is calculated as: $\text{Acc} = \frac{\text{TP}}{\text{TP} + \text{FN}}$. mF1 captures the harmonic mean of precision and recall for each category and averages them across all categories. For each category, the F1 score is defined as: $\text{F1} = 2 \times \frac{\text{Precision} \times \text{Recall}}{\text{Precision} + \text{Recall}}$, where $\text{Precision} =  \frac{\text{TP}}{\text{TP} + \text{FP}} $ and $\text{Recall} =  \frac{\text{TP}}{\text{TP} + \text{FN}} $. Together, these metrics offer a well-rounded understanding of a semantic segmentation model's performance across different scenarios and categories.

\subsubsection{Multi-modal Semantic Segmentation}
Multi-modal semantic segmentation is an advanced form of remote sensing segmentation that combines multi-source remote sensing data (e.g., optical images and radar images) to achieve accurate pixel-level classification and identify ground object categories. By leveraging complementary information from diverse data sources, this approach enhances both the accuracy and robustness of ground object identification. To evaluate the multi-modal representation of RingMoE, we select two popular remote sensing multi-modal datasets for verification.

\noindent \textbf{(1) Dataset Introduction.}
\begin{itemize}
\item \textbf{WHU-OPT-SAR (Opt/SAR-L2) \cite{li2022mcanet}}, is a multi-modal scene-level semantic segmentation dataset includes 100 pairs of Opt and amplitude SAR images, with each pair representing the same geographic area. The images have a spatial resolution of 5 meters and dimensions of 5556$\times$3704. The dataset features seven categories: ``Farmland'', ``City'', ``Village'', ``Water'', ``Forest'', ``Road'', and ``Others''. Following the official dataset guidelines, the images are cropped into 256$\times$256 pixel patches without overlapping. The dataset is split into training and testing sets with an 8:2 ratio.

\item \textbf{DFC23 (Opt/SAR-L2) \cite{DFCdata}}, is a large-scale, multi-modal dataset designed for building roof type classification and includes both Opt and amplitude SAR images. Captured by the SuperView-1, Gaofen-2, and Gaofen-3 satellites with spatial resolutions of 0.5, 0.8, and 1 meter, respectively, the dataset spans 12 roof geometry categories, such as ``Flat roof'', ``Gable roof'', ``Gambrel roof'', ``Row roof'', and multiple roof types. For experiments, images are cropped to 512$\times$512 pixels, and the dataset is split into training, validation, and testing sets with a 7:2:1 ratio.
\end{itemize}

\noindent \textbf{(2) Implementation Detail.}
The multi-modal segmentation downstream tasks are implemented using the mmsegmentation framework. We employ Upernet \cite{Upernet} as the segmentation network, with RingMoE-Opt and RingMoE-SAR$_{L2}$ serving as feature extractors for optical and amplitude SAR images, respectively. After extracting features from both modalities, we concatenate them and apply 3$\times$3 and 1$\times$1 convolutional layers for straightforward feature fusion, leveraging the inherent representational strengths of RingMoE. Finally, segmentation predictions are generated using the UperNet Head. 

The optimizer is AdamW with an initial learning rate of 6e-5 and a weight decay of 0.01. The learning rate schedule consists of two phases: a linear warm-up for the first 1,500 iterations with a factor of 0.1, followed by a polynomial scheduler with a minimum learning rate of 1e-6. For the WHU-OPT-SAR dataset, the images are cropped to 256$\times$256, with training set at 80k iterations and a batch size of 16. For the DFC23 dataset, the images are cropped to 512$\times$512, with training set at 80k iterations and a batch size of 12. In our experiments, we employ regular data augmentation strategies, including RandomFlip, RandomScaling (from 0.5 to 2.0), and RandomCrop.

\noindent \textbf{(3) Evaluation Metric.}
Similar to the segmentation task, we employ mIoU, OA, and mF1 to evaluate performance.

\subsubsection{Object Detection}
Object detection is a fundamental task in remote sensing interpretation, aiming to automatically identify and locate specific objects in remote sensing images, such as buildings, vehicles, or boats. Depending on the orientation of the detection box, this task can be divided into horizontal detection and oriented detection. To comprehensively evaluate model performance, we selected five commonly used detection datasets spanning various modalities and scenarios.

\noindent \textbf{(1) Dataset Introduction.}
\begin{itemize}
\item \textbf{DIOR~\cite{dior} and DIOR-R (Opt)~\cite{dior-r}}, is a large-scale benchmark for object detection in optical remote sensing. It comprises 23,463 images of 800$\times$800 pixels, sourced from over 80 countries. The dataset includes annotations for 192,472 instances across 20 object classes, each labeled with horizontal bounding boxes. The dataset is divided into 5862 images for training, 5863 images for validation, and 11738 images for testing. Following \cite{sun2022ringmo,guo2023skysense}, we utilize the training and validation sets together for training, and the test set for evaluation. DIOR-R is an enhanced version of DIOR, containing the same images but with oriented bounding box annotations, adding complexity to the detection task. Similar to DIOR, we merge the training and validation sets for training and reserve the test set for performance evaluation.

\item \textbf{HRSC2016 (Opt)~\cite{hrsc2016}}, is a public oriented ship detection dataset comprising images from Google Earth of diverse sizes, spanning from 300$\times$300 to 1500$\times$900, and most of them are about 1000$\times$600. The dataset includes 2976 objects with oriented bounding box annotations extracted from six important ports. The images are divided into 436 for training, 181 for validation, and 444 for testing.

\item \textbf{HRSID (SAR-L2)~\cite{wei2020hrsid}}, is a SAR-L2 image dataset (in amplitude form )for ship horizontal detection. It includes 5,604 single-channel SAR images captured under various backgrounds and lighting conditions, with a ground sampling distance (GSD) of 0.5, 1, or 3 meters. The dataset also features 16,951 ship instances. The images are divided into 3642 for training and 1962 for testing.

\item \textbf{SAR-AIRcraft-1.0 (SAR-L2)~\cite{zhirui2023sar}}, is a specialized SAR image dataset for aircraft horizontal detection. The dataset includes high-resolution single-channel SAR images with horizontal bounding annotations, consisting of 4,368 images in four different sizes and 16,463 annotated aircraft targets, covering seven aircraft categories. The dataset is split into training, validation, and testing sets with a 7:1:2 ratio.
\end{itemize}


\noindent \textbf{(2) Implementation Detail.}
For horizontal object detection, including DIOR, HRSID, and SAR-AIRcraft-1.0, all experiments are performed in the mmdetection framework\footnote[3]{\href{https://github.com/open-mmlab/mmdetection}{https://github.com/open-mmlab/mmdetection}}. We employ the Faster-RCNN~\cite{Faster_r-cnn} as the detector where the compressed RingMoE-KC is employed as the backbone for feature extraction.  For rotated object detection including DIOR-R and HRSC2016, all experiments are performed in the mmrotate framework\footnote[4]{\href{https://github.com/open-mmlab/mmrotate}{https://github.com/open-mmlab/mmrotate}}. We employ the Oriented-RCNN~\cite{Oriented_rcnn} as the detector where the compressed RingMoE-KC is employed as the backbone for feature extraction. 

We finetune our model in all detection datasets 12 epochs with 800 $\times$ 800 resolution. All experiments adopt the linear warmup schedule with a warmup ratio of 1e-3.  The RandomFlip is used as a data augmentation for both horizontal and oriented detection. The optimizer utilizes SGD with an initial learning rate 0.01 and weight decay of 0.05. The drop path is also adopted at a 0.3 rate. Note that multi-scale training isn't utilized in our fine-tuning.

\noindent \textbf{(3) Evaluation Metric.}
We evaluate our object detection result based on mean average precision (mAP). mAP is a key evaluation metric in object detection that summarizes a model's performance across different classes. It combines precision and recall to assess how well a model can both identify and localize objects within an image. For each class, the Average Precision (AP) is calculated as the area under the precision-recall curve, and mAP is the mean of these AP values across all classes. This metric is often computed at various intersection-over-union (IoU) thresholds, such as 0.5 (AP$_{50}$) or a range from 0.5 to 0.95 (mAP), to evaluate the model's ability to balance precision and recall under different matching criteria. A higher mAP indicates better overall performance.

\subsubsection{Object Tracking}

Object tracking is another critical task in remote sensing, focusing on continuously locating and identifying specific targets across consecutive frames, which is essential for dynamic scene analysis. To comprehensively assess RingMoE's tracking performance, we select two widely used object tracking datasets, both consisting of optical imagery, for evaluation. This allows us to investigate the model's capability to handle temporal changes and maintain accurate tracking in complex visual environments.

\noindent \textbf{(1) Dataset Introduction.}

\begin{itemize}
\item \textbf{AIR-MOT (Opt)~\cite{he2022multi}}, is an object tracking dataset consisting of 152 videos captured by the Jilin-1 satellite between October 2017 and October 2020. The videos have a resolution of 1920$\times$1080 pixels and a frame rate ranging from 5 to 10 FPS, with each video containing over 70 timestamps. The dataset includes 5736 instances of two object categories (``Aircraft'' and ``Ship''), collected from ten diverse regions across the globe. The complex and varied backgrounds, along with the multiscale characteristics of objects, make it challenging for trackers to maintain robustness. The dataset is split into 106 training videos and 46 testing videos. We evaluate model performance on the test set.

\item \textbf{AIR-HSAO (Opt)~\cite{ren2024motion}}, is a high-speed aerial object tracking dataset built from satellite videos captured by the Jilin satellite between 2016 and 2022. The dataset consists of 197 video clips, with a resolution of 1920$\times$1080 pixels, and includes 1437 airplane trajectories. The dataset focuses on fast-moving airplanes, with video clips covering various regions, such as urban areas, airports, and harbors. Object sizes range from less than 200 to over 5500 pixels, showcasing multiscale characteristics. The diverse and complex backgrounds add to the challenge of object tracking. The dataset is split into 128 training videos and 69 testing videos. We evaluate model performance on the test set.
\end{itemize}

\noindent \textbf{(2) Implementation Detail.} All object tracking tasks utilize the mmdetection framework, with ByteTrack\footnote[5]{\href{https://github.com/ifzhang/ByteTrack}{https://github.com/ifzhang/ByteTrack}} as the tracking algorithm. The training process follows a consistent strategy across two datasets. First, a two-stage training approach is applied to the detection model, using compressed RingMoE-Opt (KC) as the backbone for feature extraction and standard YOLOX as the detection framework. In the first stage, video frames are cropped to 640$\times$640 pixels for comprehensive fine-tuning of all parameters. The AdamW optimizer is employed with an initial learning rate of 1e-5, a weight decay of 0.005, and a batch size of 2. The learning rate linearly warms up over the first 5,000 iterations by a factor of 0.1, then remains fixed for a total of 5 epochs. In the second stage, the backbone parameters are frozen, and fine-tuning is performed on the original 1920$\times$1080 resolution frames, using the SGD optimizer with a learning rate of 4e-4, a weight decay of 0.005, and a batch size of 2, for an additional 2 epochs. For tracking inference, ByteTrack operates on the trained detection model with the following settings: a non-maximum suppression (NMS) threshold of 0.65, a tracking threshold of 0.3, a track buffer of 50 frames for maintaining lost tracks, and a matching threshold of 0.8. 

\noindent \textbf{(3) Evaluation Metric.} We employ multi-object tracking accuracy (MOTA) and ID F1 score (IDF1) to evaluate tracking performance. MOTA measures the overall tracking accuracy by considering three types of errors: false positives, false negatives (missed detections), and identity switches, which can be calculated as $\text{MOTA} = 1 - \frac{\text{FN} + \text{FP} + \text{IDSW}}{\text{GT}}$, where $\text{FN}$ is the number of false negatives, $\text{FP}$ is the number of false positives, $\text{IDSW}$ represents identity switches, and $\text{GT}$ is the total number of ground truth objects. MOTA provides a holistic view of the tracker's ability to accurately detect and track multiple objects. IDF1, on the other hand, evaluates the tracker's consistency by calculating the F1 score based on identity matches between predicted and true tracks. It is the harmonic mean of precision—indicating the proportion of correctly identified tracks among the predictions—and recall, which measures how well true tracks are accurately identified. 

\subsubsection{Change Detection}
Change detection aims to identify and assess changes between two temporal images, highlighting regions where significant differences have occurred. It is widely used for monitoring environmental changes, urban development, deforestation, and disaster impact assessment. We select two popular remote sensing change detection datasets for evaluation.

\noindent \textbf{(1) Dataset Introduction.}
\begin{itemize}
\item \textbf{LEVIR-CD (Opt)~\cite{chen2020spatial}}, is a large-scale benchmark for architectural change detection in remote sensing, comprising 637 pairs of very high-resolution (0.5 m/pixel) image patches from Google Earth. These bitemporal images, taken over intervals of 5 to 14 years, capture extensive land-use transformations and feature diverse building types, including villas, high-rise apartments, small garages, and large warehouses. The dataset is organized into 445 pairs for training, 64 pairs for validation, and 128 pairs for testing, providing a robust resource for assessing change detection methods in diverse urban contexts.
\item \textbf{CDD (Opt)~\cite{lebedev2018change}}, is a bitemporal general change detection benchmark focused on capturing seasonal variations in remote sensing images. It consists of 16,000 image pairs, each 256$\times$256 pixels, with spatial resolutions ranging from 0.03 to 1 meter per pixel. The dataset is split into 10,000 pairs for training and 3,000 pairs for testing and validation, offering a comprehensive foundation for evaluating change detection techniques under varying spatial and seasonal conditions.
\end{itemize}  

\noindent \textbf{(2) Implementation Detail.}
For all change detection tasks, we employ the Bidirectional Integration Transformer (BIT)\footnote[6]{\href{https://github.com/justchenhao/BIT_CD}{https://github.com/justchenhao/BIT}}~\cite{chen2021remote} framework, integrating the compressed RingMoE-Opt (KC) model as the backbone for feature extraction. Training is optimized with the AdamW optimizer and binary cross-entropy loss. For efficient fine-tuning, the parameters of the first stage in RingMoE are frozen, while the remaining three stages undergo training. The initial learning rate is set to 5e-5 and decays gradually to zero over 200 epochs. Data augmentation techniques include flipping, cropping, re-scaling, color jittering, and Gaussian blur. Given computational constraints, all images are uniformly divided into non-overlapping slices of 256$\times$256 pixels for training, validation, and testing.

\noindent \textbf{(3) Evaluation Metric.}
We employ the F1 score to evaluate the accuracy of identifying changed areas between two images. It is the harmonic mean of precision and recall, where precision measures the proportion of correctly detected changes out of all detected changes, and recall measures the proportion of correctly detected changes out of all actual changes. It can be formulated as $\text{F1} = 2 \times \frac{\text{Precision} \times \text{Recall}}{\text{Precision} + \text{Recall}}$, where $\text{Precision} =  \frac{\text{TP}}{\text{TP} + \text{FP}} $ and $\text{Recall} =  \frac{\text{TP}}{\text{TP} + \text{FN}} $.

\subsubsection{Depth Estimation}
Depth estimation focuses on predicting the depth value for each pixel, creating a three-dimensional view of the scene from two-dimensional images. For this purpose, experiments are conducted on the ISPRS Vaihingen and ISPRS Potsdam datasets, which provide high-resolution urban scenes ideal for evaluating depth estimation performance in complex environments.

\noindent \textbf{(1) Dataset Introduction.}
\begin{itemize}
\item \textbf{ISPRS Vaihingen~\cite{isprs2018semanticlabeling}}, is a widely used dataset for semantic segmentation and depth estimation tasks. The dataset contains 33 high-resolution images with a spatial resolution of 0.09 meters. In terms of data composition, the dataset contains three-channel (near-infrared, red, and green) images and corresponding digital surface model (DSM) data. Following the original experimental settings, we divide it into 16 training images and 17 test images during training and testing, and the digital surface model corresponding to each image is used as the label data for training.
\item \textbf{ISPRS Potsdam~\cite{isprs2018semanticlabeling}}, is similar to the Vaihingen dataset, which consists of 38 high-resolution images with a spatial resolution of 0.05 meters. The dataset contains four-channel (near-infrared, infrared, green, and blue) high-resolution images and corresponding digital surface model (DSM) data. Similarly, following the original dataset setting, we divide the Potsdam dataset into 24 training images and 14 testing images during training and testing. In addition, for the convenience of training, we did not consider the near-infrared channel and only input red, green, and blue as channels into the network.
\end{itemize}

\noindent \textbf{(2) Implementation Detail.}
The experiments employ the Binsformer\footnote[7]{\href{https://github.com/zhyever/Monocular-Depth-Estimation-Toolbox}{https://github.com/zhyever/Monocular-Depth-Estimation-Toolbox}} \cite{li2024binsformer} algorithm, utilizing RingMoE-Opt (KC) as the backbone network. Images are cropped to size 512$\times$512 for input, and data augmentation is performed using random rotation and flipping. The model is trained with a batch size of 8 and an initial learning rate of 1e-4 for 38.4k iterations. The AdamW optimizer is used with parameters ($\beta_1$, $\beta_2$, $\text{wd}$) = (0.9, 0.999, 0.01), where $\text{wd}$ denotes the weight decay. A linear learning rate warm-up strategy is applied over the first 30\% of the total iterations to stabilize early training.

\noindent \textbf{(3) Evaluation Metric.}
To assess the model's depth estimation performance, we follow standard methods and employ the mean absolute relative error metric (Rel) and accuracy metrics for evaluation. Rel quantifies the average absolute relative error between predicted depth $d_i^*$ and ground truth depth $d_i$, defined as: $\text{Rel} = \frac{1}{N} \sum_{i=1}^{N} \frac{|d_i - d_i^*|}{d_i}$, where N is the total number of pixels. The accuracy metrics $\delta_1$, $\delta_2$, and $\delta_3$ represent the percentage of pixels where the ratio between predicted and ground truth depths falls within thresholds of 1.25, 1.25$^2$, and 1.25$^3$, respectively, which can be formulated as: $
\text{accuracy}  = \frac{1}{N} \sum_{i=1}^{N} \mathbf{1}\left(\max\left(\frac{d_i^*}{d_i}, \frac{d_i}{d_i^*}\right) < \delta  \right)
$, where $\mathbf{1}(\cdot)$ is an indicator function that equals 1 if the condition inside is true and 0 otherwise. Together, these metrics provide a detailed perspective on the model's precision and robustness in-depth prediction.

\begin{table*}[t]
\caption{The class split settings of three RS scene classification datasets}
\label{Table 1}
\centering
\setlength{\abovecaptionskip}{1pt}
\setlength{\belowcaptionskip}{1pt}
\renewcommand\arraystretch{1.25}
\resizebox{0.9\textwidth}{!}{
\begin{tabular}{cccc} 
\hline
Dataset       & Training set                                                                                                                                                                                                                                                                                                      & Validation set                                                                                                                                                                            & Test set                                                                                                                                                                                          \\ 
\hline
NWPU-RESISC45 & \begin{tabular}[c]{@{}c@{}}Cloud; Mountain; Stadium;Airplane;\\Chaparral; Roundabout; Church; Ship;\\Golf course;Meadow; Harbor;\\Freeway; Lake; Wetland\\ Baseball diamond; Island; Railway;\\Mobile home park;~Palace;\\~Sparse residential; Bridge; Desert\\ Sea ice; Beach; Rectangular farmland\end{tabular} & \begin{tabular}[c]{@{}c@{}}Thermal power station;\\Storage tank; Terrace;\\Railway station; \\Tennis court;\\Snowberg;\\Industrial area;\\Runway; Overpass;\\Commercial area\end{tabular} & \begin{tabular}[c]{@{}c@{}}Airport;~Basketball court;\\River; Parking lot;\\Ground track field;\\ Medium residential;\\Dense residential;\\Circular farmland\\ Intersection; Forest\end{tabular}  \\ 
\hline
WHU-RS19      & \begin{tabular}[c]{@{}c@{}}Parking; Port; Residential;\\Bridge; Industrial; Mountain;\\Airport; Football field; Desert;\end{tabular}                                                                                                                                                                              & \begin{tabular}[c]{@{}c@{}}Railway station;\\Beach; Forest;\\Farmland; Park\end{tabular}                                                                                                   & \begin{tabular}[c]{@{}c@{}}Viaduct; Pond;\\Meadow; River;\\Commercial\end{tabular}                                                                                                                 \\ 
\hline
UCM           & \begin{tabular}[c]{@{}c@{}}Parking lot; Dense residential;\\Harbor; Chaparral;~Freeway\\ Buildings; Overpass;\\Agricultural; Medium \\ residential; Baseball diamond;\end{tabular}                                                                                                                                & \begin{tabular}[c]{@{}c@{}}Airplane; Runway;\\Forest;\\~Intersection;\\Storage tanks;\end{tabular}                                                                                        & \begin{tabular}[c]{@{}c@{}}Golf course; River;\\Sparse residential;\\Tennis court; Beach;\\Mobile home park;\end{tabular}                                                                         \\
\hline
\end{tabular}}
\end{table*}

\begin{table*}[t]
\centering
\setlength{\abovecaptionskip}{1pt}
\setlength{\belowcaptionskip}{1pt}
\caption{The class split settings of iSAID and LoveDA RS segmentation datasets for few-shot tasks} \label{Table 2}
\renewcommand{\arraystretch}{1.25}
\resizebox{0.8\linewidth}{!}{
\begin{tblr}{
  cells = {c},
  cell{2}{1} = {r=3}{},
  cell{5}{1} = {r=3}{},
  hline{1-2,5,8} = {-}{},
  hline{3-4,6-7} = {2-4}{},
}
Datasets & Fold-i & Training classes                                                                                                                                  & Testing classes                                                           \\
iSAID    & Fold-0 & {Ground track field; Bridge; Large vehicle;\\Small vehicle; Helicopter;\\Swimming pool; Roundabout;\\Soccer ball field; Plane; Harbor}            & {Ship; Storage tank;\\Baseball diamond;\\Tennis court;\\Basketball court} \\
         & Fold-1 & {Ship; Storage tank; Baseball diamond;\\Tennis court; Basketball court;\\Swimming pool; Roundabout;\\Soccer ball field; Plane; Harbor}            & {Ground track field; Bridge;\\Large vehicle;\\Small vehicle;\\Helicopter} \\
         & Fold-2 & {Ship; Storage tank; Baseball diamond;\\Tennis court; Basketball court; \\Ground track field; Bridge; Large vehicle; \\Small vehicle; Helicopter} & {Swimming pool;\\Roundabout, \\Soccer ball field;\\Plane; Harbor}         \\
LoveDA   & Fold-0 & Water; Barren; Forest; Agriculture                                                                                                                & Building; Road                                                            \\
         & Fold-1 & Building; Road; Forest; Agriculture                                                                                                               & Water; Barren                                                             \\
         & Fold-2 & Building; Road; Water; Barren                                                                                                                     & Forest; Agriculture                                                       
\end{tblr}}
\end{table*}

\subsubsection{Unseen Class Recognition}
To evaluate the model's adaptability and generalization to novel tasks/scenarios, we explore the few-shot classification and segmentation tasks in remote sensing scenarios, which aim to rapidly recognize or segment novel (unseen) object classes with a few labeled samples. 

\noindent \textbf{(1) Dataset Introduction.}
\begin{itemize}

\item  \textbf{NWPU-RESISC45 (classification)~\cite{cheng2017remote}}, 
is a publicly available dataset for RS scene classification. It contains 31,500 RGB images with a resolution of 256$\times$256 pixels. The dataset includes a total of 45 scene categories, with each category comprising 700 images. Following the settings in~\cite{cheng2021spnet,li2020dla}, the dataset is divided into 25, 10, and 10 categories for training, validation, and testing, respectively, as detailed in Tab.\ref{Table 1}.

\item  \textbf{WHU-RS19 (classification)~\cite{sheng2012high}}, is released by Wuhan University in 2012. It comprises 1,005 RGB images across 19 scene categories, with a minimum of 50 images per category. Each image is 600$\times$600 pixels in size. Following the settings in~\cite{cheng2021spnet,li2020dla}, the dataset is divided into 9, 5, and 5 categories for training, validation, and testing, respectively, as detailed in Tab.\ref{Table 1}.

\item  \textbf{UCMerced (classification)~\cite{yang2010bag}}, is designed for land-use RS scene recognition and includes 21 distinct scene categories, such as ``Agriculture'', ``Airplane'', ``Baseball field'', and ``Beach''. Each category contains 100 images with a resolution of 256$\times$256 pixels. Following the settings in \cite{cheng2021spnet,li2020dla}, the dataset is divided into 10, 6, and 5 categories for training, validation, and testing, respectively, as detailed in Tab.\ref{Table 1}.

\item  \textbf{iSAID (segmentation)~\cite{waqas2019isaid}}, is a large-scale benchmark for evaluating instance segmentation and semantic segmentation algorithms, containing 655,451 object instances from 2,806 high-resolution images. It includes 15 object categories, such as ``Ship'', ``Baseball diamond'', and ``Airplane''. To assess the model's generalization capability on unseen classes, following the setup in \cite{bi2023not}, the 15 categories are divided into three groups (Fold-i), each consisting of 10 training classes and 5 testing classes, ensuring no overlap between training and testing classes within the same group. This cross-division approach leverages the dataset's diversity to thoroughly evaluate the model's generalization performance across different categories. Detailed class divisions are provided in Tab.\ref{Table 2}.

\item  \textbf{LoveDA (segmentation)~\cite{wang2021loveda}}, is an urban-rural land cover adaptation dataset specifically designed to evaluate semantic segmentation and unsupervised domain adaptation algorithms. It includes 5,987 high-resolution images and 166,768 semantic objects from three different cities, covering seven land cover categories such as ``Road'', ``Water'', ``Barren land'', and ``Forest''. Following the setup in~\cite{bi2023not}, the background category is excluded in this study, leaving six remaining categories. These categories are further divided into three groups for evaluation, as detailed in Tab.\ref{Table 2}.

\end{itemize}

\begin{table*}[t]
\setlength{\abovecaptionskip}{1pt}
\setlength{\belowcaptionskip}{1pt}
\centering
\renewcommand\arraystretch{1.25}
\caption{Detailed performance comparison of RingMoE with varying parameter configurations on \textbf{Few-shot Classification} tasks, i.e., NWPU-RESISI45, WHU-RS19, and UCM. \textbf{Bold} and \underline{underline} denote the best and second-best results, respectively.}\label{FSL}
\resizebox{0.85\linewidth}{!}{
\begin{threeparttable}
\begin{tabular}{cr|ccc} 
\toprule
\multicolumn{2}{c|}{\multirow{3}{*}{Method}}                                                              & \multicolumn{3}{c}{5-way 1-shot}                                 \\ 
\cline{3-5}
\multicolumn{2}{c|}{}                                                                                     & NWPU-RESISC45       & WHU-RS19            & UCM                  \\ 
\cline{3-5}
\multicolumn{2}{c|}{}                                                                                     & OA(\%)              & OA(\%)              & OA(\%)               \\ 
\hline
\multirow{10}{*}{\begin{tabular}[c]{@{}c@{}}Specialized\\method\end{tabular}} 
& RelationNet(CVPR'2018)~\cite{sung2018learning}  & 66.21±0.28    & 61.74±0.51 & 48.48±0.75  \\
& DLA-MatchNet(TGRS'2021)~\cite{li2020dla} & 68.80±0.70    & 68.27±1.83 & 53.76±0.62  \\
& TSC(ISPRS'2022)~\cite{zeng2022task}         & 73.26±0.15    & 70.99±0.74 & 55.11±0.68  \\
& TDNET(TGRS'2023)~\cite{wang2023tdnet}        & 65.85±0.53    & 64.24±0.51 & -           \\
& MMML(TGRS'2023)~\cite{chen2023mmml}         & 77.35±0.73    & 77.76±0.69 & 61.42±0.34  \\
& DCN(TGRS'2023)~\cite{ji2023dual}          & 74.40±0.78    & 81.74±0.55 & 58.64±0.71  \\
&  PMPFSL(TGRS'2024)~\cite{wang2024personalized}       & 72.67±0.19    & 77.24±0.32 & 64.08±0.35  \\
& HiReNet(TGRS'2024)~\cite{tian2024hirenet}      & 70.43±0.90    & -          & 58.60±0.80  \\
& ACL-Net(TGRS'2024)~\cite{xu2024attention}      & 76.13±0.24    & 78.30±0.32 & 59.74±0.46  \\ 
\hline
\multirow{6}{*}{\begin{tabular}[c]{@{}c@{}}Foundation\\model\end{tabular}}   & RingMo(TGRS'2023)~\cite{sun2022ringmo}       & 72.88±0.23    & 74.47±0.37 & 57.67±0.74  \\
& RingMoE(6.5B)$^\dagger$           & \textbf{78.91±0.24}    & \textbf{85.24±0.36} & \textbf{66.50±0.19}  \\
& RingMoE-EP(4.3B)$^\dagger$        & \underline{78.43±0.30}    & \underline{84.42±0.49} & \underline{65.63±0.25}  \\
& RingMoE-KS(1B)$^\dagger$          & 76.72±0.27    & 81.24±0.35 & 62.79±0.33  \\
& RingMoE-KA(1B)$^\dagger$          & 77.12±0.38    & 81.67±0.43 & 62.46±0.46  \\
& RingMoE-KC(1B)$^\dagger$          & 77.75±0.26    & 82.51±0.37 & 63.27±0.26  \\
\bottomrule
\end{tabular}
\begin{tablenotes}
\footnotesize
\item[$\dagger$] RingMoE(6.5B) represents the modal-specific complete model, i.e., RingMoE-Opt. RingMoE-EP(4.3B) represents the RingMoE-Opt after pruning partial experts. RingMoE-KS(1B), -KA(1B), and -KC(1B) represent the models after summing, averaging, and compressing all expert knowledge, respectively.
\end{tablenotes}
\end{threeparttable}
}
\end{table*}

\begin{table}[t]
\setlength{\abovecaptionskip}{1pt}
\setlength{\belowcaptionskip}{1pt}
\centering
\renewcommand\arraystretch{1.25}
\caption{{Detailed performance comparison of RingMoE with varying parameter configurations on \textbf{Few-shot Segmentation} tasks, i.e., iSAID and LoveDA. Note that SegGPT~\cite{wang2023seggpt} was pre-trained with access to all categories from both datasets, while the other methods maintain evaluations on unseen classes. \textbf{Bold} and \underline{underline} denote the best and second-best results, respectively.}}\label{FSS}
\resizebox{1.0\linewidth}{!}{
\begin{tabular}{cr|cc} 
\toprule
\multicolumn{2}{c|}{\multirow{3}{*}{Method}}                                                    & \multicolumn{2}{c}{1-shot}       \\ \cline{3-4}
\multicolumn{2}{c|}{}                                                                           & iSAID          & LoveDA          \\ \cline{3-4}
\multicolumn{2}{c|}{}                                                                           & mIoU(\%)       & mIoU(\%)        \\ 
\hline
\multirow{4}{*}{\begin{tabular}[c]{@{}c@{}}Specialized\\method\end{tabular}} & PFENet(TPAMI'2022)~\cite{tian2020prior}          & 47.30          & 22.10           \\
& NERTNet(CVPR'2022)~\cite{liu2022learning}         & 46.76          & 21.74           \\
& DMNet(TGRS'2023)~\cite{bi2023not}           & 49.21          & 24.32           \\
& DCPNet(IJCV'2024)~\cite{lang2024few}          & 45.26          & 23.44           \\ 
\hline
\multirow{8}{*}{\begin{tabular}[c]{@{}c@{}}Foundation\\model\end{tabular}}   & SAM(ICCV'2023)~\cite{kirillov2023segment}& 45.38          & 17.11           \\
& \textcolor[rgb]{0.502,0.502,0.502}{SegGPT(ICCV'2023)~\cite{wang2023seggpt}}  & \textcolor[rgb]{0.502,0.502,0.502}{52.16}          & \textcolor[rgb]{0.502,0.502,0.502}{34.27}           \\
& VRP-SAM(CVPR'2024)~\cite{sun2024vrp} & 49.03          & 19.32           \\ 
\cline{2-4}
& RingMoE(6.5B)       & \textbf{52.45} & \textbf{32.40}  \\
& RingMoE-EP(4.3B)    & \underline{51.78}  & \underline{30.96}   \\
& RingMoE-KS(1B)      & 50.31          & 29.02           \\
& RingMoE-KA(1B)      & 50.43          & 28.59           \\
& RingMoE-KC(1B)      & 50.85          & 28.77           \\
\bottomrule
\end{tabular}
}
\end{table}

\noindent \textbf{(2) Implementation Detail.}
For the few-shot classification task, we conduct evaluations under the 5-way 1-shot and 5-way 5-shot settings. Specifically, the task requires classifying the query images into one of five novel classes, where each novel class is guided by either one labeled (1-shot) or five labeled examples (5-shot). Following the RelationNet~\cite{sung2018learning}, we leverage RingMoE-KC (frozen) for feature extraction, obtaining features for both the query images and the few labeled examples from the novel classes. Then we utilize the similarity metric to calculate the score between the query image and the labeled samples for classification. The input images from both datasets are resized to 256$\times$256 pixels and underwent data augmentation, including random cropping and horizontal flipping. The SGD optimizer is used with a learning rate of 0.001 and a weight decay of 5e-4. 

For the few-shot segmentation task, we evaluate under the 1-shot setting, where a labeled sample containing a novel class (namely support image) is provided to guide the segmentation of the novel class objects in the query image. We utilize RingMoE-KC to extract features from both the support and query images. Then, we apply the foreground mask of the support image to perform the masked average pooling operation, obtaining both the foreground and background prototypes. Finally, we calculate the similarity between the query features and the support prototypes to achieve segmentation. For iSAID, the image size is resized to 256$\times$256, while for LoveDA, the image size is resized to 384$\times$384. We employ SGD for optimization with a weight decay of 0.0001 and a momentum of 0.9. The batch size is set to 8, and the model is trained for 60 epochs. The initial learning rate for iSAID is 0.005, while for LoveDA, it is 0.003.

\noindent \textbf{(3) Evaluation Metric.}
For few-shot classification, following the previous methods~\cite{li2020dla,xu2024attention}, we adopt overall accuracy as the evaluation metric, defined as the ratio of correctly predicted images to the total number of images (introduced earlier in Appendix \ref{Scene classification}). And for few-shot segmentation, we utilize the class mIoU as the primary evaluation metric. The IoU for each class is calculated as $\text{IoU} = \text{TP}/(\text{TP}+\text{FP}+\text{FN})$, where $\text{TP}$, $\text{FP}$, and $\text{FN}$ denote true positives, false positives, and false negatives, respectively. The overall mIoU is defined as $\text{mIoU} = (1/n) \sum_{i=1}^n \text{IoU}_i$, with $n$ representing the number of classes in each experimental fold (e.g., $n=5$ for iSAID, $n=2$ for LoveDA). Final mIoU values are averaged across all folds.

\section{More detailed performance comparison}\label{secD}
In the main body, we primarily focus on a comparative analysis of RingMoE against the top two to three methods for each dataset, visualized through bar and radar charts. To provide a more comprehensive perspective, this section compares the performance of additional competing methods. The following figures and tables summarize the results of the exsiting methods across different tasks, presenting a broader view of RingMoE’s performance relative to the full set of methods. This extended comparison further highlights RingMoE's effectiveness in scene classification, semantic segmentation, object detection, and other remote sensing tasks, emphasizing its competitive advantages in tackling complex challenges.

\subsection{Performance on Unseen Class Recognition}

\begin{figure*}[t]
\setlength{\abovecaptionskip}{0pt}
\setlength{\belowcaptionskip}{0pt}
\centering
\includegraphics[width=1.0\linewidth]{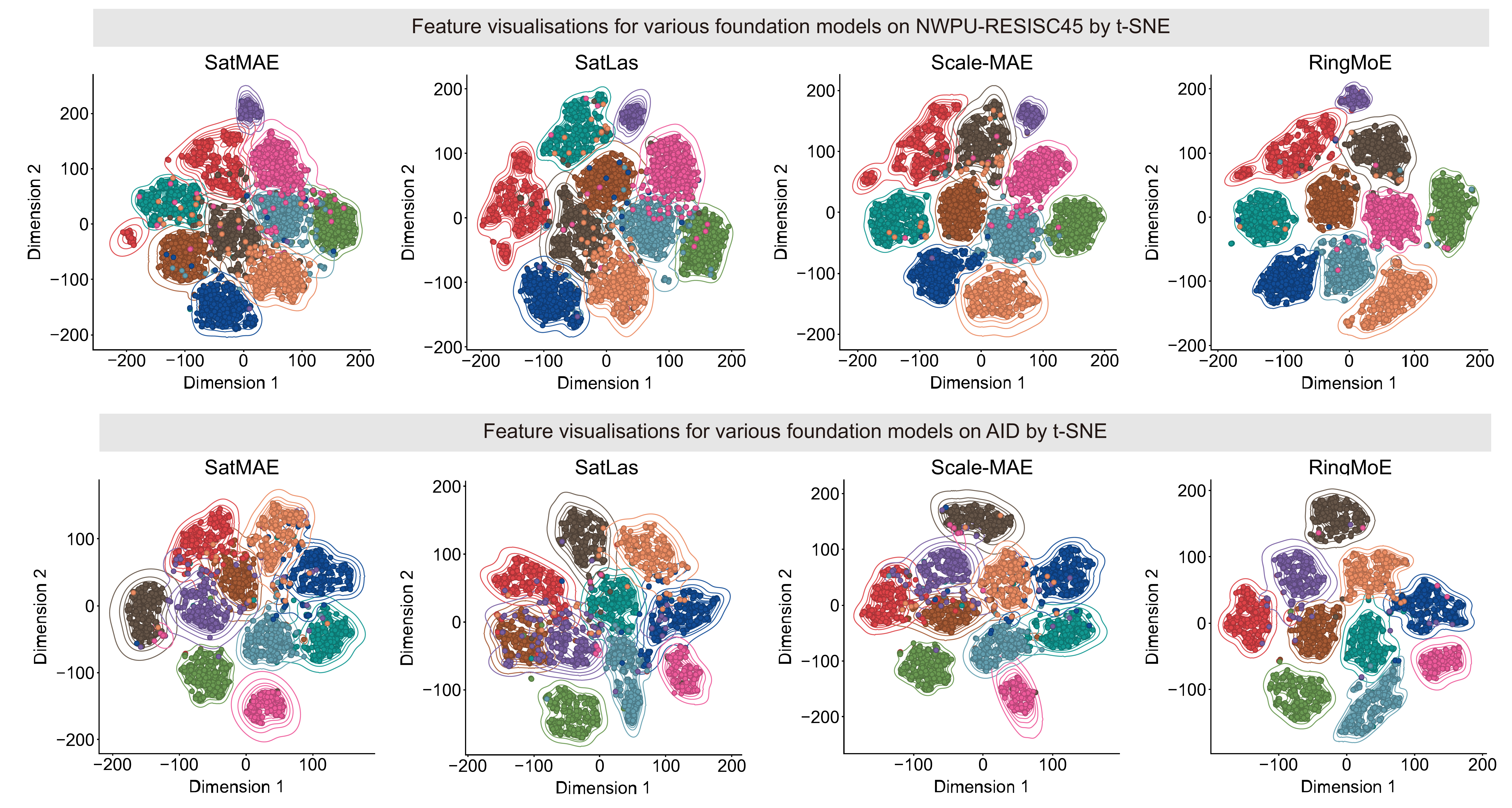}
\caption{{\textbf{Feature visualization comparison across RSFMs on Scene Classification tasks}. We extract features from various RSFMs on the test sets of both classification datasets and employ t-SNE to visualize the feature distributions of 10 randomly selected categories, highlighting the feature clustering across different models.}}\label{tsne}
\end{figure*}

\begin{table*}[t]
\setlength{\abovecaptionskip}{1pt}
\setlength{\belowcaptionskip}{1pt}
\centering
\caption{Detailed performance comparison on the \textbf{SAR-L1 Semantic Segmentation} task, i.e., SARSegL1. RingMoE achieves the best performance across all key metrics, including mean Intersection over Union (mIoU), mean accuracy (mAcc), and overall accuracy (OA). Remarkably, it also delivers superior IoU precision for each individual class, underscoring its capability in handling SAR-specific segmentation challenges effectively. \textbf{Bold} and \underline{underline} denote the best and second-best results, respectively.}
\label{tab_downstream_ss}
\renewcommand{\arraystretch}{1.25}
\resizebox{0.95\linewidth}{!}{
\begin{tabular}{r|ccccc|ccc} 
\toprule
\multirow{2}{*}{Method} & \multicolumn{5}{c|}{IoU per category(\%)}                                              & \multicolumn{1}{l}{\multirow{2}{*}{mIoU(\%)}} & \multicolumn{1}{l}{\multirow{2}{*}{mAcc(\%)}} & \multicolumn{1}{l}{\multirow{2}{*}{OA(\%)}}  \\
\cline{2-6}
                        & Water          & Vegetation     & Bare land      & Road           & Building       & \multicolumn{1}{l}{}                          & \multicolumn{1}{l}{}                          & \multicolumn{1}{l}{}                         \\  
\hline
DeepLabV3+~\cite{DeeplabV3+}             & 62.07          & 61.95          & 28.87          & 2.66           & 60.44          & 43.20                                         & 55.97                                         & 72.04                                        \\
ANN~\cite{ANN2019}                    & 59.30          & 57.90          & 26.80          & 6.56           & 56.59          & 41.44                                         & 53.66                                         & 68.50                                        \\
CCNet~\cite{CCNet2019}                  & 67.21          & 66.21          & 51.60          & 11.99          & 65.87          & 52.57                                         & \underline{63.86}                                 & 75.94                                        \\
KNet~\cite{KNet}                   & 65.92          & 66.56          & 47.14          & 10.63          & 65.08          & 51.06                                         & 61.52                                         & 75.30                                        \\
EncNet~\cite{EncNet}                  & 67.75          & \underline{69.06}  & \underline{47.93}  & 11.84          & \underline{67.47}  & \underline{52.81}                                 & 63.24                                         & \underline{77.06}                                \\
ISANet~\cite{ISANet}                 & 56.33          & 56.31          & 15.58          & 6.61           & 54.96          & 37.96                                         & 50.14                                         & 67.04                                        \\
SegNeXT~\cite{SegNext}                & 64.97          & 63.48          & 39.40          & 4.05           & 63.79          & 47.14                                         & 57.58                                         & 73.96                                        \\
DDRNet~\cite{DDRNet}                  & 56.54          & 56.96          & 25.21          & 7.67           & 55.52          & 40.38                                         & 53.23                                         & 67.30                                        \\
Mask2former~\cite{Mask2former}            & 67.78          & 63.89          & 34.33          & \underline{12.05}  & 63.61          & 48.34                                         & 58.62                                         & 74.40                                        \\ 
\hline
RingMo+PSPNet~\cite{PSPNet}          & 46.56          & 53.57          & 8.14           & 6.28           & 52.66          & 33.44                                         & 46.39                                         & 64.07                                        \\
RingMo+UperNet~\cite{Upernet}         & \underline{73.26}  & 68.31          & 43.58          & 5.35           & 65.69          & 51.24                                         & 63.53                                         & 76.85                                        \\
RingMo+Mask2former~\cite{Mask2former}     & 54.69          & 54.64          & 4.25           & 3.88           & 53.55          & 34.20                                         & 46.13                                         & 66.14                                        \\ 
\hline
\textbf{RingMoE-KC}     & \textbf{85.77} & \textbf{79.94} & \textbf{63.88} & \textbf{46.39} & \textbf{79.10} & \textbf{71.02}                                & \textbf{80.15}                                & \textbf{86.97}                               \\
\bottomrule
\end{tabular}}
\end{table*}

\begin{figure*}[h]
\centering
\setlength{\abovecaptionskip}{1pt}
\setlength{\belowcaptionskip}{1pt}
\includegraphics[width=0.95\linewidth]{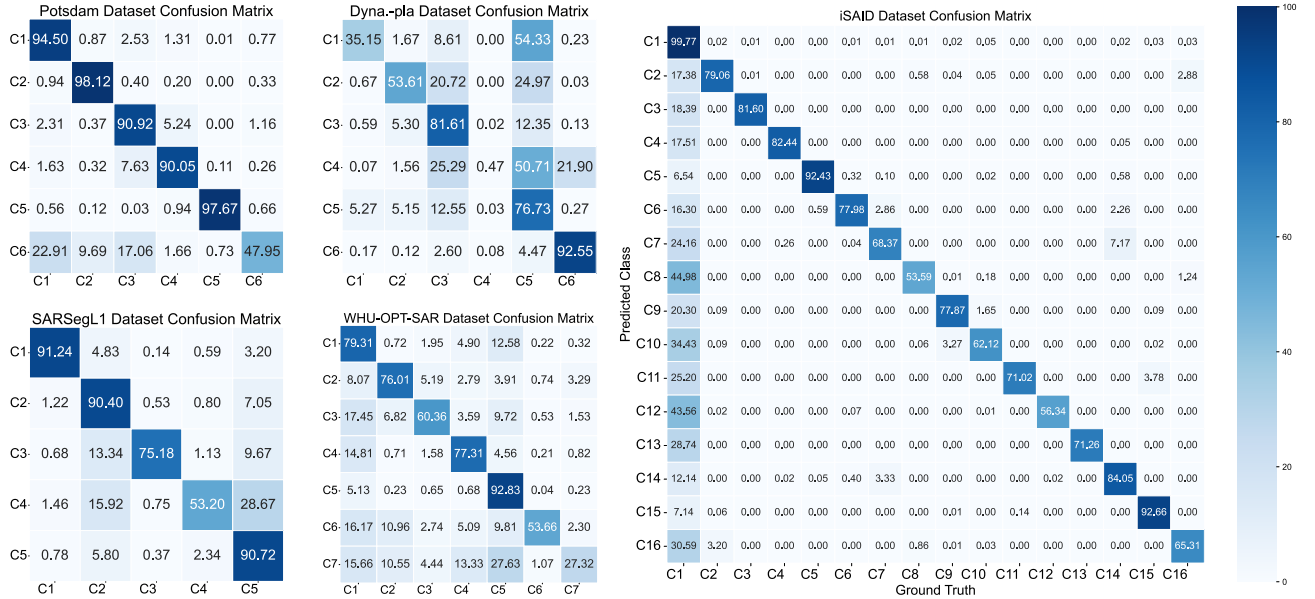}
\caption{\textbf{Confusion matrix analysis in semantic segmentation.} Confusion matrices are analyzed on selected datasets for Opt, MS, SAR-L1, and multi-modal segmentation tasks to examine per-class performance and error patterns.}\label{extend_2}
\end{figure*}

For \textbf{few-shot classification}, Tab.\ref{FSL} presents a detailed performance comparison of RingMoE with varying parameter configurations on three prominent RS classification datasets: NWPU-RESIS45~\cite{cheng2017remote}, WHU-R19~\cite{sheng2012high}, and UCM~\cite{yang2010bag}. The results highlight that RingMoE consistently achieves state-of-the-art accuracies on all three datasets, outperforming existing FSL methods. For instance, RingMoE with 6.5B parameters achieves 78.91\%, 85.25\%, and 66.50\% accuracies under the 5way-1shot setting, surpassing ACL-Net~\cite{xu2024attention} by 2.78\%, 6.94\%, and 6.76\%, respectively. Even after compression to 1B parameters through model pruning techniques like knowledge summing (RingMoE-KS), knowledge averaging (RingMoE-KA), and knowledge compressing (RingMoE-KC), it continues to deliver SOTA performance. These results underscore the robustness of RingMoE’s feature representations and its capacity to generalize effectively to unseen scenarios without relying on complex decoders.

Similarly, Tab.\ref{FSS} reports the \textbf{few-shot segmentation} performance on two segmentation datasets, iSAID~\cite{waqas2019isaid} and LoveDA~\cite{wang2021loveda}, showing comparable trends. RingMoE with 6.5B parameters achieves mIoU scores of 52.45\% and 32.40\%, outperforming DCPNet~\cite{lang2024few} by 7.19\% and 8.96\%, respectively. The pruned 4.3B parameter version of RingMoE achieves 51.78\% and 30.96\% mIoU, maintaining superiority over existing FSS methods such as DMNet\cite{bi2023not} (49.21\% and 24.32\%). Even the 1B parameter version achieves competitive SOTA results, further demonstrating the efficiency and versatility of the proposed RingMoE architecture. 

{In contrast, general models like SAM~\cite{kirillov2023segment} and SegGPT~\cite{wang2023seggpt}, although pretrained on massive RGB segmentation corpora, exhibit limited generalization to RS scenarios. SAM struggles with modality shifts, achieving only 45.38\% and 17.11\% mIoU, while SegGPT, despite being pre-exposed to all categories from both benchmarks, only marginally surpasses RingMoE on LoveDA (34.27\%) but underperforms it on iSAID (52.16\% vs. 52.45\%). These results underscore the advantages of RingMoE’s modality-aware architecture and targeted pretraining for RS applications.}

\begin{table*}[h]
\setlength{\abovecaptionskip}{1pt}
\setlength{\belowcaptionskip}{1pt}
\centering
\renewcommand\arraystretch{1.25}
\caption{Detailed performance comparison on the \textbf{SAR-L2 Semantic Segmentation} task, i.e.,AIR-POLSAR-SEG. RingMoE achieves the highest IoU accuracy across all categories, showcasing its robustness in addressing the SAR geographical objects. \textbf{Bold} and \underline{underline} denote the best and second-best results, respectively.}
\label{AIR-PoLSAR-SEG}
\resizebox{0.75\linewidth}{!}{
\begin{tabular}{r|cccc|c} 
\toprule
\multirow{2}{*}{Method} & \multicolumn{4}{c|}{IoU per category(\%)}                              & \multirow{2}{*}{mIoU(\%)}  \\\cline{2-5}
                        & Industrial area & Natural area   & Water          & Housing area        &                            \\ 
\hline
PSPNet~\cite{PSPNet}                  & 33.99           & 72.31          & 76.51          & 68.07          & 62.72                      \\
DeepLabV3+~\cite{DeeplabV3+}             & 40.62           & 70.67          & 72.93          & \underline{69.96}  & 63.55                      \\
NonLocal~\cite{NonLocal}               & 35.51           & 72.12          & 70.60          & 68.39          & 61.66                      \\
PSANet~\cite{PSANet}                 & 40.70           & 69.46          & 69.46          & 68.75          & 62.09                      \\
GCNet~\cite{Cao2019GCNetNN}                  & 38.19           & 72.64          & 74.48          & 68.37          & 63.42                      \\
FUSAR-Map~\cite{FUSARmap}              & 38.52           & 74.09          & 68.17          & 62.88          & 60.92                      \\
Mask2former~\cite{Mask2former}            & \underline{45.74}           & 73.90          & \underline{77.84}          & 69.20          &  \underline{66.67}  \\
MASA-SegNet~\cite{MASA-SegNet}            & 45.00   & \underline{74.79}  & 74.36          & 66.87          & 65.26              \\ 
\hline
\textbf{RingMoE-KC}     & \textbf{50.77}  & \textbf{77.93} & \textbf{81.37} & \textbf{72.31} & \textbf{70.59}             \\
\bottomrule
\end{tabular}
}
\end{table*}

\begin{table*}[h]
\setlength{\abovecaptionskip}{1pt}
\setlength{\belowcaptionskip}{1pt}
\centering
\caption{Detailed performance comparison on \textbf{Multi-modal Segmentation} task, i.e., WHU-OPT-SAR. RingMoE consistently achieves superior results in terms of overall accuracy (OA), mean Intersection over Union (mIoU), and mean F1-score (mF1), highlighting its effectiveness in handling multi-modal segmentation challenges. \textbf{Bold} and \underline{underline} denote the best and second-best results, respectively.}\label{WHU-OPT-SAR}
\renewcommand\arraystretch{1.25}
\resizebox{0.9\linewidth}{!}{
\begin{tabular}{r|ccccccc|ccc} 
\toprule
\multirow{2}{*}{Method} & \multicolumn{7}{c|}{Accuracy per category(\%)}                                                                & \multirow{2}{*}{OA(\%)} & \multirow{2}{*}{mIoU(\%)} & \multirow{2}{*}{mF1(\%)}  \\ \cline{2-8}
                        & Farmland      & City          & Village       & Water         & Forest        & Road          & Others        &                         &                           &                           \\ 
\hline
SegFormer~\cite{xie2021segformer}              & 79.1          & 72.9          & 38.0          & 64.7          & 88.1          & 0.3           & 0.4           & 75.5                    & 40.3                      & 51.7                      \\
SETR~~\cite{zheng2021rethinking}                   & 79.0          & 70.4          & 47.5          & 67.8          & 88.8          & 17.3          & 12.1          & 77.0                    & 45.2                      & 58.8                      \\
Unetformer~~\cite{wang2022unetformer}             & 80.9          & 69.4          & 59.7          & 76.5          & 89.8          & 47.5          & 21.2          & 80.3                    & 49.2                      & 63.7                      \\
Segmenter~\cite{strudel2021segmenter}              & 82.3          & 75.2          & 51.7          & 74.4          & 89.4          & 16.0          & 12.0          & 79.9                    & 51.2                      & 63.4                      \\
PSCNN~~\cite{hughes2018identifying}                  & 81.5          & 63.9          & 56.2          & 73.5          & 90.1          & 38.0          & 20.1          & 79.7                    & 50.7                      & 64.8                      \\
CFNet~\cite{kang2022cfnet}                  & 81.5          & \textbf{80.9} & 50.5          & 68.8          & \underline{90.5}  & 19.9          & \underline{37.5}  & 79.9                    & \underline{52.7}              & 66.2                      \\
MBFNet~\cite{li2020multimodal}                 & \underline{82.6}  & \underline{78.7}  & 59.9          & 76.9          & 90.4          & 44.0          & 13.0          & 81.5                    & 51.1                      & 64.9                      \\
CMFNet~\cite{ma2022crossmodal}                 & 80.0          & 66.7          & \underline{64.3}  & \underline{79.7}  & \textbf{90.8} & \underline{54.3}  & 24.7          & 81.1                    & 52.1                      & 66.5                      \\
V\_FuseNet~~\cite{audebert2018beyond}             & 83.9          & 73.1          & 60.7          & 78.5          & 89.2          & 53.0          & 27.2          & 81.6                    & 48.2                      & 61.0                      \\
DDHRNet~\cite{ren2022dual}                & \textbf{84.2} & 69.6          & 59.7          & \textbf{80.1} & 89.9          & 49.1          & 23.9          & \underline{82.2}            & 52.3                      & \underline{66.6}              \\ 
\hline
\textbf{RingMoE-KC}     & 81.6          & 75.6          & \textbf{67.0} & 78.2          & 88.9          & \textbf{55.7} & \textbf{54.9} & \textbf{84.1}           & \textbf{54.7}             & \textbf{68.7}             \\
\bottomrule
\end{tabular}}
\end{table*}

\begin{table*}[h]
\setlength{\abovecaptionskip}{1pt}
\setlength{\belowcaptionskip}{1pt}
\renewcommand\arraystretch{1.35}
\centering
\caption{Detailed performance comparison on \textbf{Multi-modal Segmentation} task, i.e., DFC23. RingMoE consistently achieves superior results in terms of overall accuracy (OA) and mean Intersection over Union (mIoU). \textbf{Bold} and \underline{underline} denote the best and second-best results, respectively.} \label{DFC23}
\resizebox{0.95\linewidth}{!}{
\begin{threeparttable}
\begin{tabular}{r|ccccccccccccc|cc}
\toprule
\multirow{2}{*}{Method} & \multicolumn{13}{c|}{IoU per category(\%)}                                                                                                                                                                        & \multirow{2}{*}{OA(\%)} & \multirow{2}{*}{mIoU(\%)}  \\ 
\cline{2-14}
& BG            & FR            & GR            & GR2           & RR            & MER           & HRv1          & HRv2          & MR            & PR            & AR            & Do            & OR            &                     &                        \\ 
\hline
SETR~\cite{zheng2021rethinking}                     & 88.0          & 59.7          & 41.4          & 58.0          & 66.9          & 3.5           & 19.4          & 32.2          & 6.2           & 2.7           & 37.0          & 20.9          & 48.9          & 88.5                & 37.3                   \\
FuseNett~\cite{hazirbas2017fusenet}                 & 82.4          & 36.0          & 19.1          & 18.6          & 42.8          & 3.4           & 2.6           & 7.8           & 2.3           & 5.6           & 16.2          & 6.6           & 11.9          & 82.0                & 19.6                   \\
CFNet~\cite{kang2022cfnet}                   & 84.2          & 44.7          & 22.8          & 31.2          & 41.1          & 0.0           & 5.6           & 6.8           & 0.0           & 0.0           & 7.8           & 2.8           & 19.8          & 83.9                & 20.5                   \\
TransFuser~\cite{prakash2021multi}              & 89.6          & 57.0          & 36.8          & 14.8          & 50.0          & 13.3          & 24.2          & 15.6          & 10.2          & 20.4          & 7.7           & 15.1          & 29.0          & 87.9                & 29.5                   \\
TokenFusion~\cite{wang2022multimodal}             & 90.2          & 55.5          & 40.9          & 43.2          & 56.9          & 12.4          & 24.6          & 17.0          & 9.8           & 22.6          & 13.2          & 15.3          & 18.2          & 88.9                & 32.3                   \\
PSCNN~\cite{hughes2018identifying}& 90.2          & 62.4          & 45.8          & 54.9          & 58.9          & 10.8          & 12.8          & 22.5          & 11.3          & 20.8          & 8.6           & 16.8          & 26.5          & 89.5                & 34.1                   \\
V\_FuseNet~\cite{audebert2018beyond}              & 92.0          & 64.2          & 49.5          & 55.0          & 65.1          & 12.6          & 15.8          & 36.8          & 12.0          & 5.0           & 46.0          & 37.4          & 44.8          & 90.9                & 41.2                   \\
mmFormer~\cite{zhang2022mmformer}                & 91.7          & 67.7          & 58.8          & \underline{72.2}          & \underline{76.3}          & 20.3          & 42.0          & \underline{62.1}          & 16.8          & 21.5          & 57.0          & 5.2           & 50.3          & 91.4                & 49.4                   \\
MCANet~\cite{li2022mcanet}                  & \underline{94.3}          & \underline{76.6}          & \underline{68.8}          & 70.4          & 71.4          & \underline{45.1}          & \underline{46.5}          & 61.5          & \textbf{69.6} & \underline{57.8}          & \textbf{73.1} & \underline{78.9}          & \underline{67.8}          & \underline{94.2}                & \underline{67.9}                   \\ 
\hline
\textbf{RingMoE-KC}                  & \textbf{96.4} & \textbf{81.0} & \textbf{73.0} & \textbf{72.2} & \textbf{83.4} & \textbf{77.2} & \textbf{49.7} & \textbf{70.0} & \underline{63.3}          & \textbf{62.3} & \underline{72.7}          & \textbf{83.3} & \textbf{70.1} & \textbf{95.8}       & \textbf{73.4}    \\ \bottomrule      
\end{tabular}
\begin{tablenotes}
\footnotesize
\item[] Note that the categories of the DFC23 are: Flat\_roof (FR), Gable\_roof (GR), Gambrel\_roof (GR2), Row\_roof (RR), Multiple\_eave\_roof (MER), Hipped\_roof\_v1 (HRv1), Hipped\_roof\_v2 (HRv2), Mansard\_roof (MR), Pyramid\_roof (PR), Arched\_roof (AR), Dome\_(Do), Others (OR).
\end{tablenotes}
\end{threeparttable}
}
\end{table*}

\subsection{Performance on Scene Classification}
Tab.2 in the main body presents the performance comparison of the proposed RingMoE and other foundation models on two widely used RS scene classification datasets. The results demonstrate that RingMoE achieves outstanding performance even with limited data. Specifically, the 6.5B-parameter RingMoE achieves 95.90\% and 98.19\% accuracies on NWPU-RESISC45 and AID, respectively, outperforming SkySense~\cite{guo2023skysense} by 1.05\% and 0.51\%, and Scale-MAE~\cite{reed2023scale} by 3.27\% and 1.75\%. In addition, we extract the features from various foundation models on the test set of the two datasets and randomly select 10 categories for visual representation using t-SNE (see Fig.\ref{tsne}). Results indicate that RingMoE exhibits a more discriminative representation capability than other foundation models, characterized by tighter intra-class clustering and larger inter-class variance. 

To facilitate low-cost deployment, various pruning strategies were employed to compress the model while retaining its effectiveness. For instance, the 4.3B-parameter RingMoE-EP, utilizing sparse expert pruning, achieves 95.43\% and 97.88\% accuracies on RESISC45 and AID, respectively, closely approximating the performance of the unpruned model. Notably, the 1B-parameter RingMoE-KC, compressed through dense expert integration, maintains high classification accuracy, surpassing other foundation models and performing comparably to the 2B-parameter SkySense. For example, RingMoE-KC achieves 95.05\% accuracy on NWPU-RESISC45, exceeding SkySense~\cite{guo2023skysense} by 0.2\% and SatLas~\cite{SatLas} by 2.89\%.

These results highlight the robust feature extraction capabilities of RingMoE, enabling it to adapt effectively to diverse remote sensing scenarios, even when compressed to as few as 1B parameters. This scalability and efficiency make RingMoE highly suitable for practical applications. Consequently, the performance of the 1B-parameter RingMoE-KC is further evaluated across additional downstream tasks.

\subsection{Performance on Semantic Segmentation}

\begin{itemize}
\item \textbf{Optical/Multi-spectral segmentation}. Tab.3 in the main body showcases the performance of RingMoE compared to existing RSFMs across two optical datasets (iSAID~\cite{waqas2019isaid} and Potsdam) and one multi-spectral dataset (Dyna.-pla~\cite{toker2022dynamicearthnet}). The 1B-parameter RingMoE achieves scores of 69.70\%, 93.54\%, and 47.6\% on these datasets, respectively, demonstrating SOTA performance on Dyna.-pla and securing the second position on iSAID and Potsdam, trailing only SkySense~\cite{guo2023skysense}. Furthermore, RingMoE has a significant performance advantage over other RSFMs such as Scale-MAE~\cite{reed2023scale}, GFM~\cite{GFM}, and SatLas~\cite{SatLas}, outperforming SatLas by 6.9\% mIoU on Dyna.pla. Confusion matrix analysis in Fig.\ref{extend_2} confirms RingMoE's stable segmentation performance and ability to distinguish categories effectively.

\item \textbf{SAR segmentation}. Tab.\ref{tab_downstream_ss}-\ref{AIR-PoLSAR-SEG} present the performance comparison of RingMoE against other specialized methods on two SAR datasets: the complex-valued SAR dataset SARSegL1~\cite{DENet(SARsegL1)} and the amplitude SAR dataset Air-POLSAR-SEG~\cite{AIR-PolSAR-Seg}. For SARSegL1, RingMoE demonstrates significant advancements, outperforming the existing state-of-the-art method, EncNet~\cite{EncNet}, by 18.21\%, 16.91\%, and 9.91\% in mIoU, mAcc, and OA, respectively. Remarkably, RingMoE achieves the highest mIoU scores across all fine-grained categories, reflecting its superior capability in complex SAR data interpretation. On Air-POLSAR-SEG, RingMoE achieves a 5.33\% improvement in mIoU over the second-best method, MASA-SegNet~\cite{MASA-SegNet}, with particularly notable gains in key categories: Industrial Area (12.8\%), Natural Area (4.19\%), Water (6.35\%), and Housing (3.35\%). These results highlight RingMoE’s robust adaptability and effectiveness in processing diverse SAR data types.

\item \textbf{Multi-modal segmentation}. Tab.\ref{WHU-OPT-SAR}-\ref{DFC23} highlight the comparative performance of RingMoE and other specialized methods on two multi-modal segmentation datasets: WHU-OPT-SAR~\cite{li2022mcanet} and DFC23~\cite{DFCdata}, which integrate optical and amplitude SAR modalities. For WHU-OPT-SAR, RingMoE achieves 84.1\% OA, 54.7\% mIoU, and 68.7\% mF1, surpassing the second-best models by 1.9\% in OA (DDHRNet~\cite{ren2022dual}, 82.2\%), 2.0\% in mIoU (CFNet~\cite{kang2022cfnet}, 52.7\%), and 2.1\% in mF1 (DDHRNet, 66.6\%). For DFC23, RingMoE attains 95.8\% OA and 73.4\% mIoU, outperforming the second-best method MACNet~\cite{li2022mcanet} by 1.6\% in OA and 5.5\% in mIoU. Impressively, among the 13 fine-grained categories in the DFC23 dataset, RingMoE secures 12 first-place rankings and 1 second-place ranking for mIoU scores. These results underscore RingMoE’s superior capability to handle complex multi-modal segmentation tasks, achieving consistent SOTA performance across diverse datasets and fine-grained categories.
\end{itemize}

\begin{table*}[t]
\setlength{\abovecaptionskip}{1pt}
\setlength{\belowcaptionskip}{1pt}
\renewcommand\arraystretch{1.25}
\caption{Performance comparison on \textbf{SAR Object Detection} task, i.e., SAR-AIRCraft-1.0. RingMoE achieves the highest AP$_{50}$ accuracy across all categories, effectively detecting SAR geographic objects. \textbf{Bold} and \underline{underline} denote the best and second-best results, respectively.}\label{sarplane}
\centering
\resizebox{0.9\linewidth}{!}{
\begin{tabular}{r|ccccccc|c} 
\toprule
\multirow{2}{*}{Method}  & \multicolumn{7}{c|}{AP$_{50}$ per category(\%)}                                                               & \multirow{2}{*}{mAP$_{50}$(\%)}  \\ 
\cline{2-8}
                         & A330          & A320/321      & A220          & ARJ21         & Boeing737     & Boeing787     & Other         &                           \\ 
\hline
Cascade R-CNN~\cite{cai2018cascade}           & 87.4          & 97.5          & 74.0          & 78.0          & 54.5          & 68.3          & 69.1          & 75.7                      \\
YOLOX-Nano~\cite{ge2021yolox}              & 74.7          & 96.9          & 79.7          & 78.7          & 66.6          & 78.2~         & 73.8          & 81.3                      \\
SA-Net~\cite{zhirui2023sar}                  & 88.6          & 94.3          & 90.3          & 78.6          & 59.7          & 70.8~         & 71.3          & 77.7                      \\
SkG-Net~\cite{fu2021scattering}                 & 66.4          & 78.2          & 66.4          & 65.0          & 65.1          & 69.6~         & 71.4          & 70.7                      \\
YOLOv5s~\cite{yolov5}                 & 92.1          & 98.9          & 87.4          & 86.4          & 76.3          & \underline{96.2} & 85.1          & 89.0                      \\
YOLOv8s~\cite{yolov5}                 & 95.2          & 97.7          & \underline{95.8}  & 86.6          & \underline{78.9}  & 90.9~         & 84.4          & 89.6                      \\
MLSDNet~\cite{chang2023mlsdnet}                 & 91.5          & 96.9          & 85.1          & 83.2          & 71.7          & 72.1          & 78.4          & 82.7                      \\
DiffusionDet~\cite{chen2023diffusiondet}            & 95.4          & 98.1          & 80.8          & 84.2          & 70.9          & 91.4          & \underline{86.4}  & 86.6                      \\
DiffDet4SAR~\cite{zhou2024diffdet4sar}              & \underline{97.1}  & \underline{99.4}  & 82.3          & \underline{87.2}  & 72.8          & 93.3          & 85.9          & 88.4                      \\
SARATR-X~\cite{yang2024saratr}                & -             & -             & -             & -             & -             & -             & -             & 86.1                      \\
SFS-CNet~\cite{li2024unleashing}                & 95.9          & 99.3          & 87.9          & 86.7          & 77.9          & 92.9~         & 85.6          & \underline{89.7}              \\ 
\hline
\textbf{RingMoE-KC} & \textbf{98.4} & \textbf{99.6} & \textbf{99.4} & \textbf{97.8} & \textbf{94.7} & \textbf{99.4} & \textbf{95.3} & \textbf{97.8}             \\
\bottomrule
\end{tabular}}
\end{table*}

\begin{figure*}[!h]
\setlength{\abovecaptionskip}{1pt}
\setlength{\belowcaptionskip}{1pt}
\centering
\includegraphics[width=0.75\linewidth]{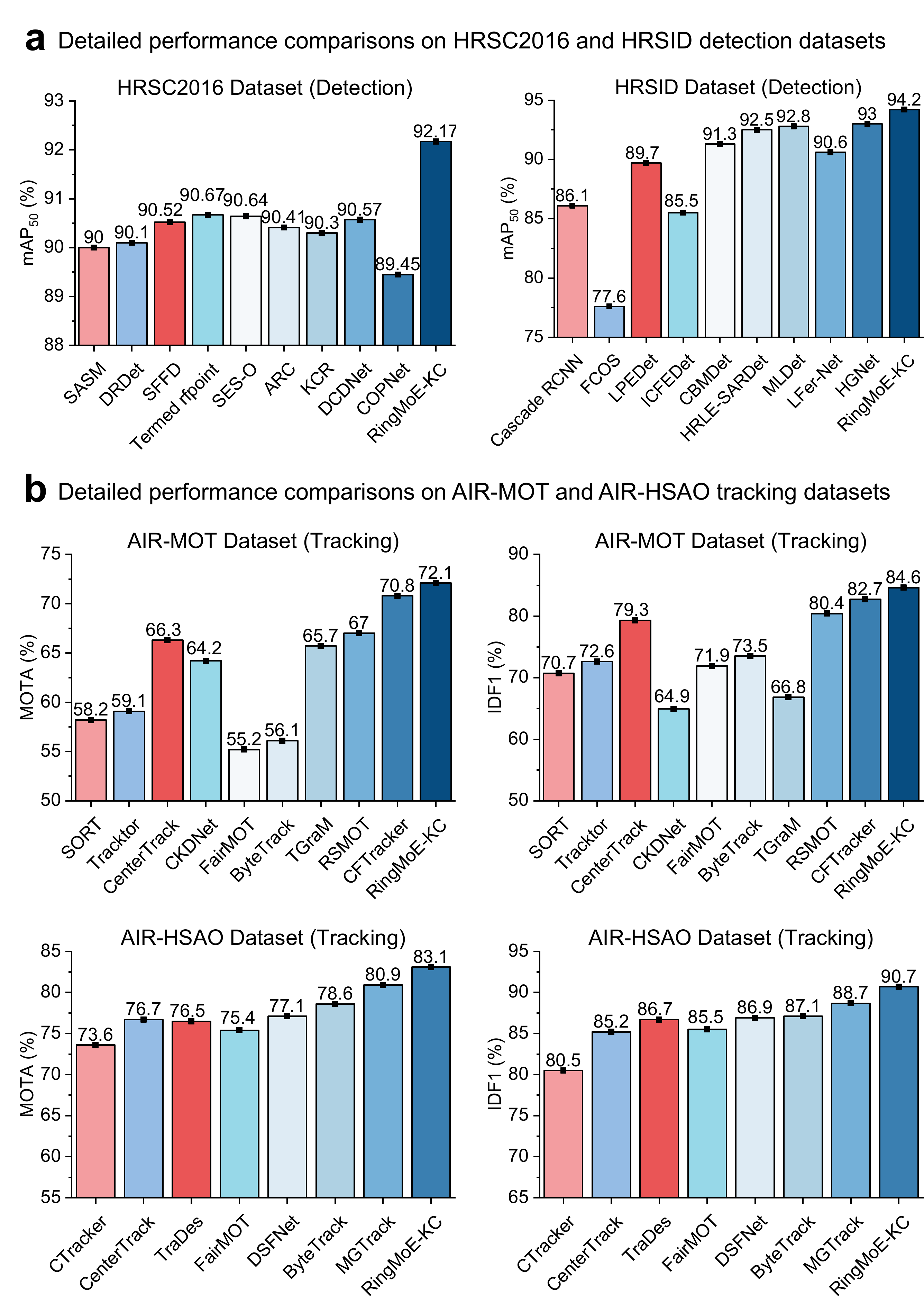}
\caption{\textbf{Detailed performance comparison on several object detection and object tracking tasks.} \textbf{a.} Detailed performance comparison on HRSC2016 and HRSID detection datasets. \textbf{b.} Detailed performance comparison on AIR-MOT and AIR-HSAO tracking datasets.}
\label{fig:apendix2}
\end{figure*}

\begin{table*}[t]
\centering
\setlength{\abovecaptionskip}{1pt}
\setlength{\belowcaptionskip}{1pt}
\caption{Detailed performance comparison on the \textbf{Change Detection} tasks, i.e., LEVIR-CD and CDD datasets. \textbf{Bold} and \underline{underline} denote the best and second-best results, respectively.}\label{cd}
\renewcommand{\arraystretch}{1.25}
\resizebox{0.65\linewidth}{!}{
\begin{tabular}{cr|c|cc} 
\toprule
\multicolumn{2}{c|}{\multirow{2}{*}{Method}}                                                         & \multirow{2}{*}{Backbone} & LEVIR-CD       & CDD             \\
\cline{4-5}
\multicolumn{2}{c|}{}                                                                                &                           & F1(\%)         & F1(\%)          \\ 
\hline
\multirow{7}{*}{\begin{tabular}[c]{@{}c@{}}Specialized\\method\end{tabular}}   
& DTSCN~\cite{liu2020building}              & ResNet Variant            & 87.67          & 92.09           \\
& BIT~\cite{chen2021remote}               & ResNet-18                 & 90.14          & 94.40           \\
& ChangeFormer~\cite{bandara2022transformer}       & Transformer Variant       & 90.40          & 94.63           \\
& TransUnetCD~\cite{li2022transunetcd}        & ResNet-50+ViT-Base       & 90.47          & 93.40           \\
& GCD-DDPM~\cite{wen2024gcd}           & U-Net                     & 90.96          & 94.93           \\
& SEIFNet~\cite{10419228}            & ResNet-18                 & 90.02          & 95.69           \\
& MDIPNet~\cite{10599227}            & Swin-Tiny                 & 91.49          & 97.35           \\ 
\hline
\multirow{7}{*}{\begin{tabular}[c]{@{}c@{}}\\Foundation\\Model\\\end{tabular}} 
& RingMo~\cite{sun2022ringmo}             & Swin-Base                 & 91.86          & -               \\
& RVSA~\cite{wang2022advancing}               & ViT-Base                  & 90.86          & -               \\
& SatMAE~\cite{cong2022satmae}             & ViT-Large                 & 87.65          & -               \\
& Scale-MAE~\cite{reed2023scale}           & ViT-Large                 & 92.07          & -               \\
& ChangeCLIP~\cite{DONG202453}         & CLIP(ResNet-50)           & 92.01          & \underline{97.89}   \\
& SkySense(2B)~\cite{guo2023skysense}& Swin-Huge                 & \underline{92.58}  & -               \\ \cline{2-5}
& \textbf{RingMoE-KC} & -                         & \textbf{92.59} & \textbf{98.13}  \\ 
\bottomrule
\end{tabular}}
\end{table*}

\begin{table*}[t]
\footnotesize
\setlength{\abovecaptionskip}{1pt}
\setlength{\belowcaptionskip}{1pt}
\caption{Detailed performance comparison on the \textbf{Depth Estimation} tasks, i.e., Vaihingen and Potsdam. \textbf{Bold} and \underline{underline} denote the best and second-best results, respectively.} \label{Generalization2DTo3D}
\renewcommand\arraystretch{1.35}
\centering
\resizebox{0.85\linewidth}{!}
{\begin{tabular}{r|c|cccc|cccc} 
\toprule
\multirow{2}{*}{Method} & \multirow{2}{*}{Publication} & \multicolumn{4}{c}{ISPRS Vaihingen} & \multicolumn{4}{c}{ISPRS Potsdam}  \\ 
\cline{3-10}
~ & ~ & Rel$\downarrow$ & $\delta_1\uparrow$ & $\delta_2\uparrow$ & $\delta_3\uparrow$ 
& Rel$\downarrow$ & $\delta_1\uparrow$ & $\delta_2\uparrow$ & $\delta_3\uparrow$  \\ 
\hline
D3Net\cite{carvalho2018regression}      & ICIP'2018  & 0.271 & 0.526 & 0.791 & 0.922 & 0.090 & 0.952 & 0.995 & 0.998   \\
Amirkolaee et al.\cite{amirkolaee2019height} & ISPRS'2019      & 1.163 & 0.330 & 0.572 & 0.741 & 0.571 & 0.342 & 0.601 & 0.782  \\       
HEOL\cite{li2020height}      & GRSL'2020  & 0.222 & 0.621 & 0.833 & 0.940 & 0.083 & 0.954 & 0.992 & 0.998  \\
Adabins\cite{bhat2021adabins}    & CVPR'2021  & 0.286 & 0.513 & 0.808 & 0.937 & -  & -  & -  & - \\ 
LeReS\cite{yin2021learning}      & CVPR'2021  & 0.260 & 0.554 & 0.800 & 0.932 & -  & -  & -  & -   \\
WMD\cite{ramamonjisoa2021single}        & CVPR'2021  & 0.272 & 0.543 & 0.798 & 0.916 & -  & -  & -  & -   \\
ASSEH\cite{liu2022associatively}      & TGRS'2022 & 0.237 & 0.595 & 0.860 & 0.971 & - & - & - & -  \\
DepthsFormer\cite{agarwal2022depthformer}      & ICIP'2022       & 0.212	& 0.716 & 0.927 & 0.967 & 0.123 & 0.871 & 0.981 & 0.997 \\ 
BinsFormer\cite{li2024binsformer}        & CVPR'2022       & 0.203	& 0.745 & 0.931 & \underline{0.975} & 0.117 & 0.876 & 0.989 & 0.999  \\
SFFDE\cite{mao2023elevation}            & TGRS'2023 & 0.222 & 0.595 & 0.897 & 0.970 & - & - & - & -   \\
Feng et al.\cite{feng2023height}       & ISPRS'2023      & 0.187 & 0.705 & 0.898 & 0.973 & \underline{0.063} & \textbf{0.979} & \underline{0.997} & 0.999   \\
HeightFormer\cite{chen2024heightformer}      & RS'2024         & \underline{0.185}	& \underline{0.756} & \underline{0.941} & 0.973 & 0.104 & 0.893 & 0.987 & 0.997  \\
\hline
\textbf{RingMoE (KC-1B)}& - & \textbf{0.121} & \textbf{0.915}  & \textbf{0.946} & \textbf{0.983} & \textbf{0.046} & \underline{0.970} & \textbf{0.998} & \textbf{0.999}   \\                    
\bottomrule
\end{tabular}}
\end{table*}

\subsection{Performance on Object Detection}

\begin{itemize}

\item \textbf{Optical detection}. Tab.4 in the main body compares the performance of RingMoE with existing remote sensing foundation models (RSFMs) on two optical object detection datasets: DIOR and DIOR-R (rotated bounding boxes). RingMoE achieves remarkable results, setting new benchmarks with 82.45\% mAP$_{50}$ on DIOR and 76.04\% mAP$_{50}$ on DIOR-R. These results surpass SkySense~\cite{guo2023skysense} by 3.72\% and 1.77\%, respectively, and SelectiveMAE~\cite{wang2024scaling} by 4.65\% and 5.73\%. Additionally, Fig.\ref{fig:apendix2}.a showcases the performance of RingMoE on the optical ship detection dataset HRSC2016. RingMoE establishes a new state-of-the-art with 92.17\% mAP$_{50}$, outperforming the second-best method (rfpoint~\cite{wang2023learnable}) by 1.5\% and the third-best method (SES-O~\cite{yang2023sampling}) by 1.53\%. These results underscore RingMoE’s robust feature representation, demonstrating its ability to excel across diverse object detection scenarios in remote sensing tasks.

\item \textbf{SAR detection}. Tab.\ref{sarplane} and Fig.\ref{fig:apendix2}.a present the performance of RingMoE and specialized methods on two amplitude SAR object detection datasets: SAR-AIRCraft-1.0~\cite{zhirui2023sar} and HRSID~\cite{wei2020hrsid}. For SAR-AIRCraft-1.0 (aircraft detection), RingMoE achieves the highest detection accuracy across all fine-grained aircraft categories, setting a new benchmark with 97.8\% mAP$_{50}$. This performance surpasses the second-best method (SFS-CFNet~\cite{li2024unleashing}) by 8.1\% and the third-best method (YOLOv8s~\cite{yolov5}) by 8.2\%. For HRSID (ship detection), RingMoE attains an impressive 94.2\% mAP$_{50}$, outperforming HGNet~\cite{chen2024hgm} by 1.2\% and MLDet~\cite{zhao2023multitask} by 1.4\%. These results highlight RingMoE’s robust detection capabilities in amplitude SAR scenarios, demonstrating superior generalization and adaptability across challenging tasks.
\end{itemize}

\subsection{Performance on Object Tracking}

Fig.\ref{fig:apendix2}.b illustrates the performance of RingMoE and specialized methods on two object tracking datasets, AIR-MOT~\cite{he2022multi} and AIR-HSAO~\cite{ren2024motion}. For AIR-MOT, RingMoE demonstrates the best performance across all metrics, surpassing the previous SOTA method CFTracker~\cite{kong2023cftracker}. Specifically, RingMoE improves MOTA by 1.3\% and IDF1 by 1.9\%. On AIR-HSAO, RingMoE also outperforms MGTrack~\cite{ren2024motion}, achieving a 2.2\% improvement in MOTA and a 2.0\% increase in IDF1. These results demonstrate the effectiveness of RingMoE in both tracking tasks, showcasing its performance in dynamic and complex scenarios. The model achieves enhanced detection and tracking accuracy, indicating its suitability for real-time applications that involve continuous adaptation to changing conditions. Its strong feature extraction and generalization capabilities are critical for handling dynamic scenes, making it an effective solution for real-time processing in advanced remote sensing tasks.

\subsection{Performance on Change Detection}

Tab.\ref{cd} presents the quantitative results of our proposed RingMoE alongside other advanced techniques on the LEVIR-CD and CDD datasets. Overall, compared with mainstream change detection methods (such as ChangeFormer \cite{bandara2022transformer} and MDIPNet \cite{10599227}), the RSFMs do bring significant performance gains. For instance, RingMo~\cite{sun2022ringmo} enhances BIT’s performance on the LEVIR-CD dataset, while ChangeCLIP~\cite{DONG202453} achieves advanced results across both datasets. SkySense~\cite{guo2023skysense} ranks second on LEVIR-CD, benefiting from the generalized feature representations learned from large-scale data, which aids in distinguishing semantic differences in bi-temporal images. Despite these advances, our model achieves the highest performance metrics on both the LEVIR-CD and CDD datasets, setting a new benchmark in change detection.

\subsection{Performance on Depth Estimation}
We achieve SOTA performance on two widely used benchmarks for large-scale remote sensing depth estimation tasks (see Tab.\ref{Generalization2DTo3D}). Specifically, on the Vaihingen dataset, our method outperforms the current leading method, Heightformer~\cite{chen2024heightformer}, by reducing the error metric Rel by 35\%. Additionally, we improve the accuracy metrics $\delta_1$, $\delta_2$, and $\delta_3$ by 15.9\%, 0.5\%, and 1.0\%, respectively, marking a significant leap in depth estimation performance. On the Potsdam dataset, our method sets a new benchmark, with performance comparable to that of Feng et al.~\cite{feng2023height}, further solidifying its effectiveness in depth estimation. These results demonstrate the superiority of our approach in achieving highly accurate depth estimation across multiple datasets.

\section{Comparison with popular MLLMs} \label{secE}
{
We appreciate the remarkable progress of recent MLLMs, which have demonstrated impressive generalization across vision-language tasks such as VQA, captioning, and image reasoning. These models are typically built upon large language models (LLMs) (e.g., GPT-4~\cite{achiam2023gpt}, LLaMA~\cite{touvron2023llama}, Qwen~\cite{bai2023qwen}) and trained using massive natural image-text corpora, making them highly effective for instruction-following and general-purpose multimodal reasoning.
}
{
\begin{itemize}
\item \textbf{Task Specialization:} RS tasks such as semantic segmentation, change detection, and object tracking involve dense, structured predictions over wide-area imagery with numerous spatially distributed targets. These require precise geometric reasoning and spatial alignment—capabilities not optimized in current MLLMs, which are primarily designed for language-centric, generative tasks (e.g., captioning, VQA). Even domain-adapted RS-MLLMs struggle with such perception-oriented challenges.
\item \textbf{Modality Diversity:} RS data encompass diverse modalities including optical, SAR (amplitude and complex-valued), and multispectral imagery, each with unique physical and visual properties. However, general MLLMs are trained mainly on RGB imagery, limiting their adaptability to non-optical modalities. Even recent RS-MLLMs typically emphasize optical data, leading to suboptimal performance on SAR or MS tasks~\cite{hu2025ringmo}.
\end{itemize}
}
{
In contrast, RingMoE is a vision-centric foundation model featuring a Mixture-of-Experts~\cite{dai2024deepseekmoe} architecture and large-scale multi-modal pretraining. Its design explicitly incorporates modality-aware processing and pixel-wise self-supervision, enabling label-efficient, and task-specific RS understanding without relying on textual input.}

\begin{table}[t]
\setlength{\abovecaptionskip}{1pt}
\setlength{\belowcaptionskip}{1pt}
\caption{{Quantitative comparison between \textbf{popular MLLMs and the proposed RingMoE on the NWPU-RESISC45~\cite{cheng2017remote} classification} dataset. All models are evaluated using Overall Accuracy (OA) under a consistent protocol of 20\% training and 80\% testing, except for RingMoE$^{\dagger}$ which adopts a stricter 10\% training split. Results of several MLLMs are quoted from ~\cite{yao2025falcon}.}} \label{tab:1_1}
\renewcommand\arraystretch{1.25}
\centering
\resizebox{0.7\linewidth}{!}{
\begin{tabular}{lcc} 
\toprule
Method            & \#Params & OA(\%)          \\ 
\hline
\multicolumn{3}{l}{\textit{General MLLMs}}     \\
MiniCPM-V~\cite{yao2024minicpm}         & 3B       & 15.00           \\
MiniGPTv2~\cite{chen2023minigpt}         & 7B       & 31.00           \\
LLaVA-1.5~\cite{liu2024improvedbaselinesvisualinstruction}         & 7B       & 46.00           \\
Qwen-VL-Chat~\cite{bai2023qwen}      & 7B       & 32.00           \\ 
\hline
\multicolumn{3}{l}{\textit{RS MLLMs }}         \\
GeoChat~\cite{kuckreja2024geochat}           & 7B       & 58.00           \\
EarthGPT~\cite{zhang2024earthgpt}          & 7B       & 93.84           \\
Falcon~\cite{yao2025falcon}            & 0.7B     & 94.00           \\
RingMo-Agent~\cite{hu2025ringmo}      & 3B       & 94.72           \\ 
\hline
\multicolumn{3}{l}{\textit{Vision Foundation Models}}     \\
\textbf{RingMoE-KC$^{\dagger}$} & 1B       & \underline{95.05}           \\
\textbf{RingMoE-KC}  & 1B       & \textbf{96.28}  \\
\bottomrule
\end{tabular}
}
\end{table}

\begin{table}[t]
\setlength{\abovecaptionskip}{1pt}
\setlength{\belowcaptionskip}{1pt}
\caption{{Quantitative comparison of \textbf{popular MLLMs and the proposed RingMoE on two RS optical detection} datasets: DIOR~\cite{dior} and HRSC2016~\cite{hrsc2016}. All results are reported using mean Average Precision (mAP$_{50}$).}} \label{tab:2}
\renewcommand\arraystretch{1.25}
\centering
\resizebox{0.8\linewidth}{!}{
\begin{tabular}{lccc} 
\toprule
Method       & \#Params & DIOR           & HRSC2016        \\ 
\hline
\multicolumn{4}{l}{\textit{General MLLMs}}                 \\
MiniGPTv2~\cite{chen2023minigpt}    & 7B       & 9.43           & 51.16           \\
Qwen-VL-Chat~\cite{bai2023qwen} & 7B       & 15.81          & 58.55           \\ 
\hline
\multicolumn{4}{l}{\textit{RS MLLMs}}                      \\
Falcon~\cite{yao2025falcon}       & 0.7B     & 56.65          & \textbf{93.75}  \\
RingMoGPT~\cite{wang2024ringmogpt}    & 14B      & \underline{64.70}          & 83.26           \\ 
\hline
\multicolumn{4}{l}{\textit{Vision Foundation Model}}       \\
\textbf{RingMoE-KC}      & 1B       & \textbf{82.45} & \underline{92.17}           \\
\bottomrule
\end{tabular}
}
\end{table}

\begin{table}[t]
\setlength{\abovecaptionskip}{1pt}
\setlength{\belowcaptionskip}{1pt}
\caption{{Quantitative comparison of \textbf{popular MLLMs and the proposed RingMoE on the SARDet-100k~\cite{li2024sardet}} dataset using mean Average Precision (mAP$_{50}$). RingMoE achieves a substantial improvement over existing MLLMs, demonstrating its superior capability for synthetic aperture radar (SAR) object detection.}}\label{tab:3}
\renewcommand\arraystretch{1.25}
\centering
\resizebox{0.6\linewidth}{!}{
\begin{tabular}{lc} 
\toprule
Method           & SARDet-100k     \\ 
\hline
MiniGPTv2~\cite{chen2023minigpt}        & 0.00            \\
RingMo-Agent~\cite{hu2025ringmo}     & \underline{53.84}           \\
\textbf{RingMoE-KC} & \textbf{89.72}  \\
\bottomrule
\end{tabular}
}
\end{table}

{
To support our analysis, we conducted head-to-head evaluations of RingMoE against both general MLLMs and emerging RS-MLLMs on three challenging RS tasks, as illustrated in Tab.\ref{tab:1_1}-\ref{tab:3}:
\begin{itemize}
    \item Optical Scene Classification (NWPU-RESISC45~\cite{cheng2017remote}): RingMoE significantly outperforms all general MLLMs (e.g., MiniGPTv2~\cite{chen2023minigpt}, LLaVA~\cite{liu2024improvedbaselinesvisualinstruction}, Qwen-VL-Chat~\cite{bai2023qwen}) in Overall Accuracy (OA), despite using a stricter 10\% training split. It also exceeds recent RS-MLLMs such as Falcon~\cite{yao2025falcon} and EarthGPT~\cite{zhang2024earthgpt}.
    \item Optical Object Detection (DIOR~\cite{dior}, HRSC2016~\cite{hrsc2016}): RingMoE achieves substantial improvements in mAP$_{50}$, especially on DIOR, which involves cluttered, multi-object scenes. This demonstrates its robustness in structured prediction and small object localization compared to language-centric MLLMs.
    \item SAR Object Detection (SARDet-100k~\cite{li2024sardet}): On this challenging benchmark involving SAR-L2 imagery, RingMoE achieves 89.72\% mAP$_{50}$, while general MLLMs completely fail (0.00), and even strong RS-MLLMs (e.g., RingMo-Agent~\cite{hu2025ringmo}) fall behind. This validates RingMoE’s strong adaptability to complex RS modalities.
\end{itemize}
}
{
These results collectively demonstrate that RingMoE not only bridges the modality gap in RS visual understanding, but also establishes new performance baselines across dense, label-efficient vision tasks under limited supervision, surpassing both general-purpose and domain-adapted MLLMs.
}

\section{More comprehensive ablation studies and analysis}\label{secF}

\subsection{Comparison of Expert Pruning Strategies}
{To provide a comprehensive analysis of our expert pruning strategies, we conduct ablation studies across six RS tasks using 17 public datasets, as summarized in Tab.\ref{tab:5}. All variants are derived from the same full-capacity RingMoE model (14.7B parameters) and share identical initialization prior to pruning, ensuring fair comparison.}

{Overall, our results indicate that:}
{
\begin{itemize}
    \item RingMoE-EP, preserves the sparse expert-parallel structure while pruning only underutilized experts. It consistently delivers the strongest performance across most tasks, validating the utility of sparse high-capacity experts under minimal pruning.
    \item RingMoE-KS/KA/KC compresses expert knowledge into dense representations via knowledge summing, averaging, and compressing, respectively. Despite reducing the parameter count from 14.7B to ~1B, these models maintain competitive accuracy, highlighting the robustness of the distilled representations.
    \item Notably, RingMoE-KC frequently provides the best trade-off among the dense variants, particularly in few-shot classification and semantic segmentation tasks, demonstrating its effectiveness in preserving critical knowledge during expert clustering and integration.
\end{itemize}
}

{These results confirm that our pruning strategies generalize well across modalities and task types, while enabling flexible deployment under varying resource constraints. They provide a clear trade-off spectrum between model compactness and performance, substantiating the design choices in our pruning pipeline.}

\begin{table*}[t]
\setlength{\abovecaptionskip}{1pt}
\setlength{\belowcaptionskip}{1pt}
\caption{{Performance comparison of four \textbf{RingMoE pruning variants} across six representative RS tasks on 17 public datasets. Unless otherwise specified, all datasets consist of optical imagery.
All models are derived from the same pre-trained full-capacity RingMoE. RingMoE-EP prunes underutilized experts while preserving the sparse expert structure. 
In contrast, RingMoE-KS, RingMoE-KA, and RingMoE-KC integrate expert knowledge into dense representations via knowledge summing, averaging, and compressing, respectively (see Sec.3.4 of the manuscript for details). All variants are evaluated using standard task-specific metrics (e.g., OA, mIoU, mAP$_{50}$, F1, MOTA, Rel), showing that the proposed pruning strategies preserve strong downstream performance while offering flexible trade-offs between efficiency and accuracy.}} \label{tab:5}
\renewcommand\arraystretch{1.25}
\centering
\resizebox{1.0\linewidth}{!}{
\begin{tabular}{c|c|c|cccc} 
\toprule
Downstream
Tasks                         & Datasets                  & Metrics   & RingMoE-EP & RingMoE-KS & RingMoE-KA & RingMoE-KC  \\ 
\hline
\multirow{3}{*}{Few-shot
Classification} & NWPU-RESISC45~\cite{cheng2017remote}             & OA(\%)    & 78.43      & 76.72      & 77.12      & 77.75       \\
& WHU-RS19~\cite{sheng2012high}                  & OA(\%)    & 84.42      & 81.24      & 81.67      & 82.51       \\
& UCM~\cite{yang2010bag}                       & OA(\%)    & 65.63      & 62.79      & 62.46      & 63.27       \\ 
\hline
\multirow{2}{*}{Few-shot
Segmentation}   & iSAID~\cite{waqas2019isaid}                     & mIoU(\%)  & 51.78      & 50.31      & 50.43      & 50.85       \\
& LoveDA~\cite{wang2021loveda}                    & mIoU(\%)  & 30.96      & 29.02      & 28.59      & 28.77       \\ 
\hline
\multirow{2}{*}{Scene Classification}      & NWPU-RESISC45~\cite{cheng2017remote}             & OA(\%)    & 95.43      & 94.81      & 94.87      & 95.05       \\
& AID~\cite{xia2017aid}                        & OA(\%)    & 97.88      & 97.30      & 97.23      & 97.42       \\ 
\hline
\multirow{4}{*}{Semantic
Segmentation}   & Potsdam (Opt)~\cite{isprs2018semanticlabeling}             & mF1(\%)   & 93.95      & 93.09      & 93.28      & 93.54       \\
& Dyna.-pla (MS)~\cite{toker2022dynamicearthnet}            & mIoU(\%)  & 48.24      & 47.15      & 46.77      & 47.60       \\
& SARSegL1 (SAR-L1)~\cite{DENet(SARsegL1)}         & mIoU(\%)  & 72.64      & 70.29      & 69.55      & 71.02       \\
& AIR-POLSAR-SEG (SAR-L2)~\cite{AIR-PolSAR-Seg}   & mIoU(\%)  & 71.18      & 69.75      & 69.52      & 70.59       \\ 
\hline
Multi-modal Segmentation                   & WHU-OPT-SAR (Opt/SAR-L2)~\cite{li2022mcanet}               & mIoU(\%)  & 55.92      & 54.36      & 54.01      & 54.70       \\ 
\hline
\multirow{2}{*}{Object Detection}          & DIOR (Opt)~\cite{dior}                & mAP$_{50}$(\%) & 83.33      & 81.37      & 81.60      & 82.45       \\
& SAR-AIRCraft-1.0 (SAR-L2)~\cite{zhirui2023sar} & mAP$_{50}$(\%) & 98.13      & 98.18      & 97.51      & 97.80       \\ 
\hline
Change Detection                           & LEVIR-CD~\cite{ren2024motion}                  & F1(\%)    & 92.68      & 92.34      & 92.53      & 92.59       \\ 
\hline
Object Tracking                            & AIR-HSAO~\cite{ren2024motion}                  & MOTA(\%)  & 83.84      & 82.66      & 82.59      & 83.10       \\ 
\hline
Depth Estimation                           & Vaihingen~\cite{isprs2018semanticlabeling}                 & Rel$\downarrow$       & 0.105      & 0.145      & 0.138      & 0.121       \\
\bottomrule
\end{tabular}
}
\end{table*}

\subsection{Discussion on SAR-L1 Pre-training for Enhanced SAR Understanding}
{To empirically validate the importance of modeling SAR-L1 data, we conduct a focused ablation comparing two variants of our RingMoE model: one pre-trained with SAR-L1 data and the other without. Both variants share the same backbone and collaborative expert structure, ensuring a fair comparison. In our evaluation, SAR-L1 downstream tasks are evaluated utilizing the RingMoE-SAR$_{L1}$, and SAR-L2 tasks are evaluated utilizing the RingMoE-SAR$_{L2}$. For models lacking SAR-L1 in pretraining, the RingMoE-SAR$_{L1}$ is randomly initialized from scratch. Results are reported on four representative downstream tasks, SARSegL1~\cite{DENet(SARsegL1)} for SAR-L1 segmentation, AIR-POLSAR-SEG~\cite{AIR-PolSAR-Seg} for SAR-L2 segmentation, and HRSID~\cite{wei2020hrsid} \& SAR-AIRCraft-1.0~\cite{zhirui2023sar} for SAR-L2 object detection.}

{As illustrated in Tab.\ref{tab:11}, introducing SAR-L1 during pretraining yields notable performance improvements across both SAR-L1 and SAR-L2 tasks. Notably, SAR-L1 segmentation mIoU improves by 9.28\%, while SAR-L2 tasks also benefit, e.g., +3.04\% on AIR-POLSAR-SEG and +4.64\% on SAR-AIRCraft-1.0, despite the absence of phase in downstream SAR-L2 inputs. We attribute this cross-modal generalization to the collaborative expert design in RingMoE, which encourages knowledge sharing between SAR-L1 and SAR-L2 modalities. This confirms our hypothesis: explicitly incorporating SAR-L1 during pretraining leads to richer, physically grounded representations that benefit a wide range of SAR tasks, even when phase information is not present at inference.}

\begin{table}[t]
\setlength{\abovecaptionskip}{1pt}
\setlength{\belowcaptionskip}{1pt}
\caption{{Ablation study on the effect of \textbf{incorporating SAR-L1 data during pretraining}. We compare two variants of the RingMoE model: one pre-trained with SAR-L1 data and the other without. Both variants share the same backbone and collaborative expert structure to ensure a fair comparison.}} \label{tab:11}
\renewcommand\arraystretch{1.25}
\centering
\resizebox{1.0\linewidth}{!}{
\begin{tabular}{c|cccc} 
\toprule
\multirow{3}{*}{SAR-L1 Data}               & SAR-L1 Seg. & SAR-L2 Seg. & \multicolumn{2}{c}{SAR-L2 Det.}  \\
                                  & SARSegL1            & AIR-POLSAR-SEG      & HRSID     & SAR-AIRcraft-1.0          \\ 
\cline{2-5}
                                  & mIoU(\%)            & mIoU(\%)            & mAP$_{50}$(\%) & mAP$_{50}$(\%)                 \\ 
\hline
\textcolor[rgb]{0.702,0.702,0.702}{\ding{55}}  & 61.74               & 67.55               & 92.68     & 93.16                     \\
\ding{51}    & 71.02               & 70.59               & 94.20     & 97.80                     \\
\bottomrule
\end{tabular}
}
\end{table}

\begin{table}[t]
\setlength{\abovecaptionskip}{1pt}
\setlength{\belowcaptionskip}{1pt}
\caption{{Ablation study on the \textbf{physics-informed loss design} for the SAR-L1 modality. We replace the proposed physics-informed reconstruction loss with a conventional pixel-wise reconstruction loss during SAR-L1 pretraining.}} \label{tab:10}
\renewcommand\arraystretch{1.25}
\centering
\resizebox{1.0\linewidth}{!}{
\begin{tabular}{cc|cccc} 
\toprule
\multicolumn{2}{c|}{\multirow{2}{*}{Loss Function}}                                                                                                 & SAR-L1 Seg. & SAR-L2 Seg.    & \multicolumn{2}{c}{SAR-L2 Det.}  \\
\multicolumn{2}{c|}{}                                                                                                                           & SARSegL1    & AIR-POLSAR-SEG & HRSID     & SAR-AIRCraft-1.0     \\ 
\hline
\begin{tabular}[c]{@{}c@{}}Physis-informed \\Reconstruction\end{tabular} & \begin{tabular}[c]{@{}c@{}}Pixel\\Reconstruction\end{tabular} & mIoU(\%)    & mIoU(\%)       & mAP$_{50}$(\%) & mAP$_{50}$(\%)            \\ 
\hline
\textcolor[rgb]{0.702,0.702,0.702}{\ding{55}} & \ding{51}                                                             & 65.35       & 68.23          & 93.16     & 94.66                \\
\ding{51}                                                                               &  \textcolor[rgb]{0.702,0.702,0.702}{\ding{55}}                                                             & 71.02       & 70.59          & 94.20     & 97.80                \\
\bottomrule
\end{tabular}
}
\end{table}

\subsection{Discussion on Physics-Informed Loss Function for SAR-L1 Pre-training}

{To enhance the physical interpretability and representational quality of SAR-L1 features, we introduce a physics-informed reconstruction loss based on power conservation during polarimetric decomposition, instead of standard pixel-wise loss. This loss enforces the model to reconstruct image power rather than raw complex pixels, embedding physical constraints into pretraining.}

{To explore its impact, we replace the physics-informed loss with a conventional pixel-level reconstruction loss while keeping all settings unchanged. Tab.\ref{tab:10} shows that this substitution leads to performance drops on multiple SAR downstream tasks, e.g., 5.7\%↓ mIoU on SARSegL1~\cite{DENet(SARsegL1)} and 3.1\% mAP↓ on SAR-AIRCraft~\cite{zhirui2023sar} detection. Notably, although the loss is applied only to SAR-L1 data during pretraining, its benefits transfer to SAR-L2 tasks as well. This cross-modal improvement arises from collaborative experts in RingMoE, which propagate physically-aligned representations from SAR-L1 to SAR-L2 branches during joint training. These results validate that incorporating physical priors at the loss level significantly enhances both modality-specific modeling and generalization across SAR variants.}

\begin{figure*}[!h]
\centering
\setlength{\abovecaptionskip}{1pt}
\setlength{\belowcaptionskip}{1pt}
\includegraphics[width=0.95\linewidth]{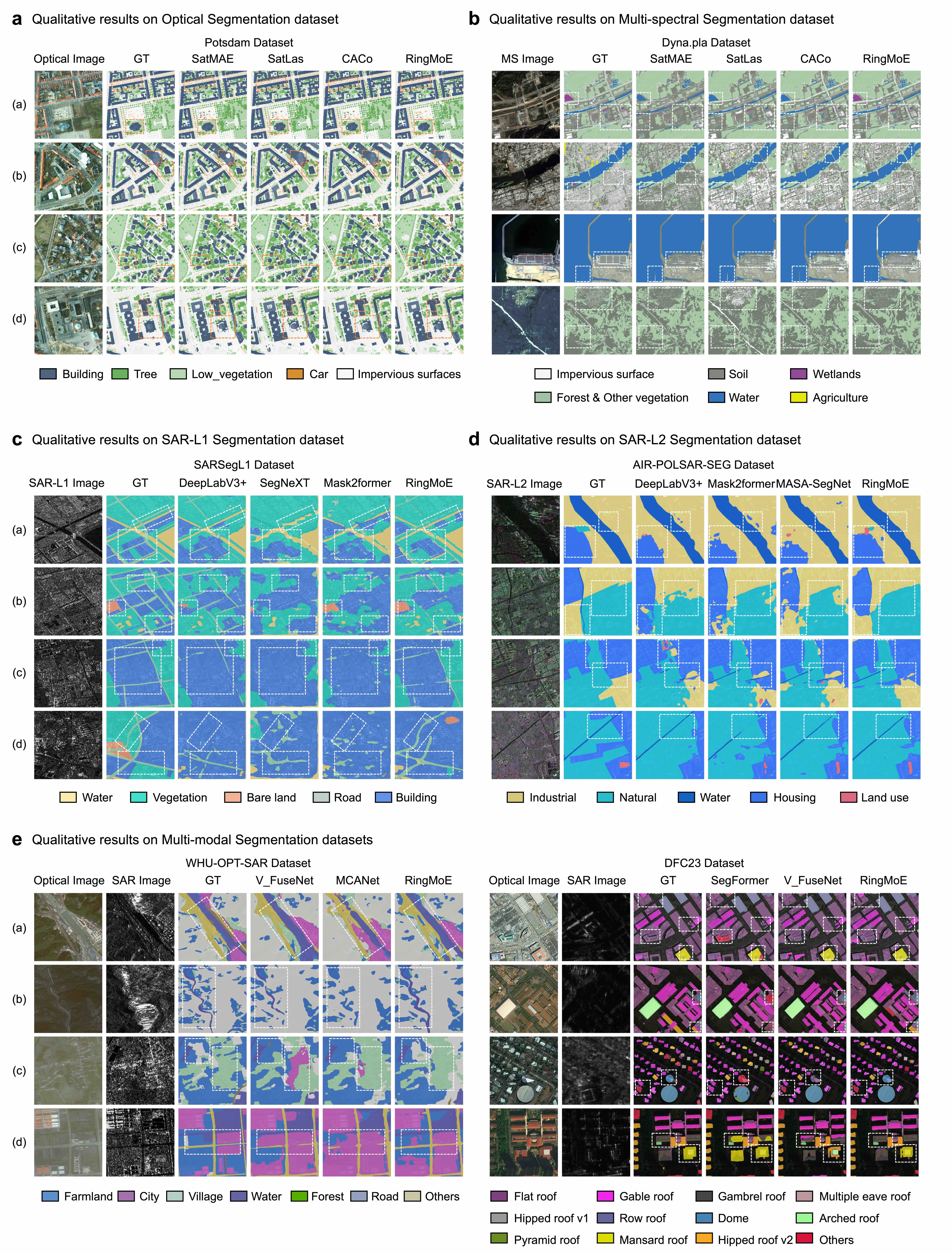}
\caption{\textbf{Qualitative results of RingMoE are provided across multiple semantic segmentation datasets, covering various data modalities.} These include the optical dataset Potsdam, the multi-spectral dataset Dyna.pla, the SAR-L1 dataset SARSegL1, the SAR-L2 dataset AIR-POLSAR-SEG, the multi-modal dataset WHU-OPT-SAR, and DFC23. For each dataset, four representative cases are presented, with significant regions of interest highlighted using dashed-line boxes.}\label{extend_3}
\end{figure*}

\begin{figure*}[!h]
\centering
\setlength{\abovecaptionskip}{1pt}
\setlength{\belowcaptionskip}{1pt}
\includegraphics[width=1.0\linewidth]{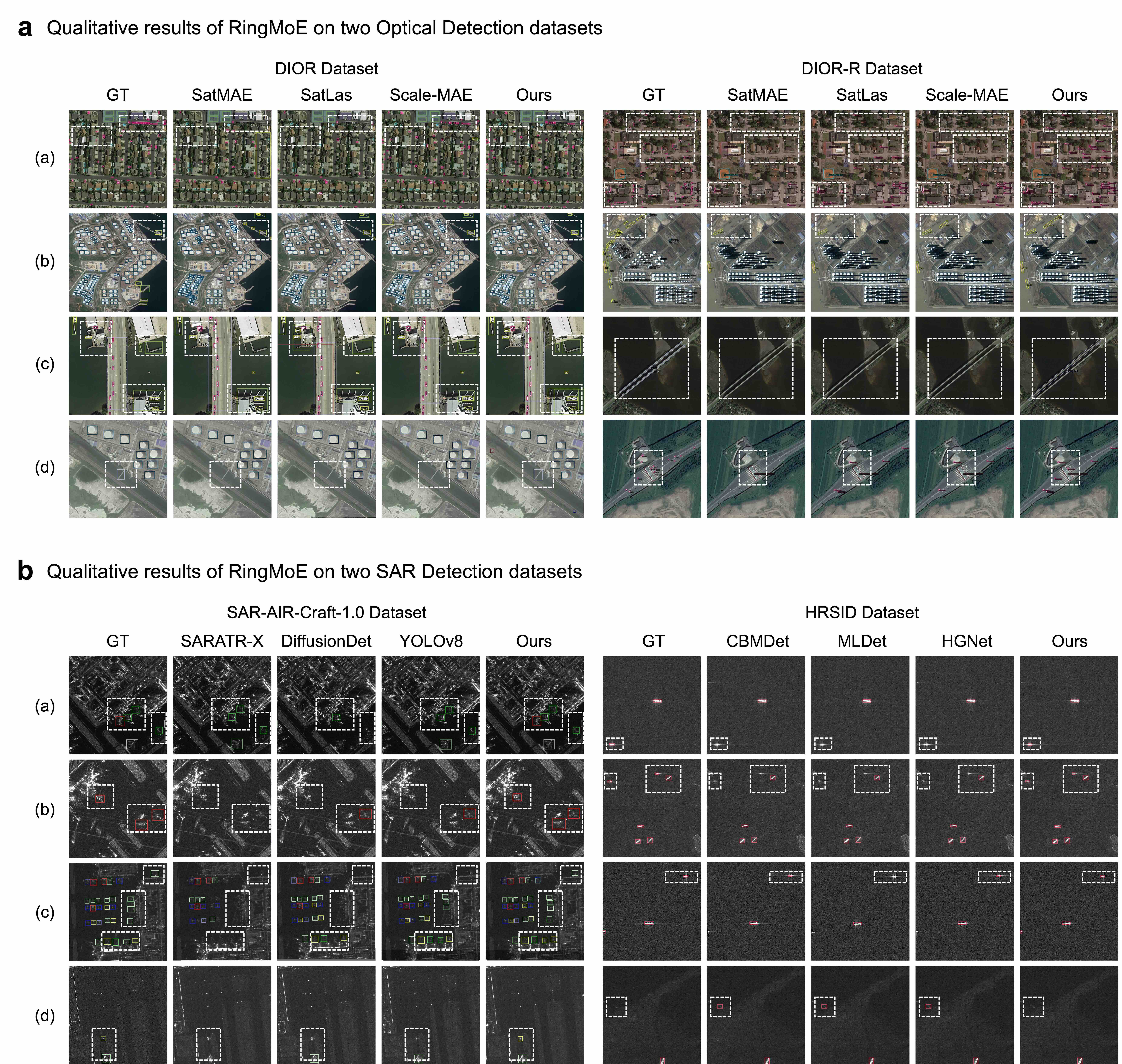}
\caption{\textbf{Qualitative results of RingMoE are demonstrated across multiple object detection datasets, encompassing various data modalities.} These include the optical datasets DIOR and DIOR-R (rotated bounding boxes) and the SAR-L2 datasets SAR-AIR-Craft-1.0 and HRSID. For each dataset, four representative cases are showcased, with key prominent regions highlighted using dashed-line boxes.}\label{extend_4}
\end{figure*}

\begin{figure*}[!h]
\centering
\setlength{\abovecaptionskip}{1pt}
\setlength{\belowcaptionskip}{1pt}
\includegraphics[width=0.8\linewidth]{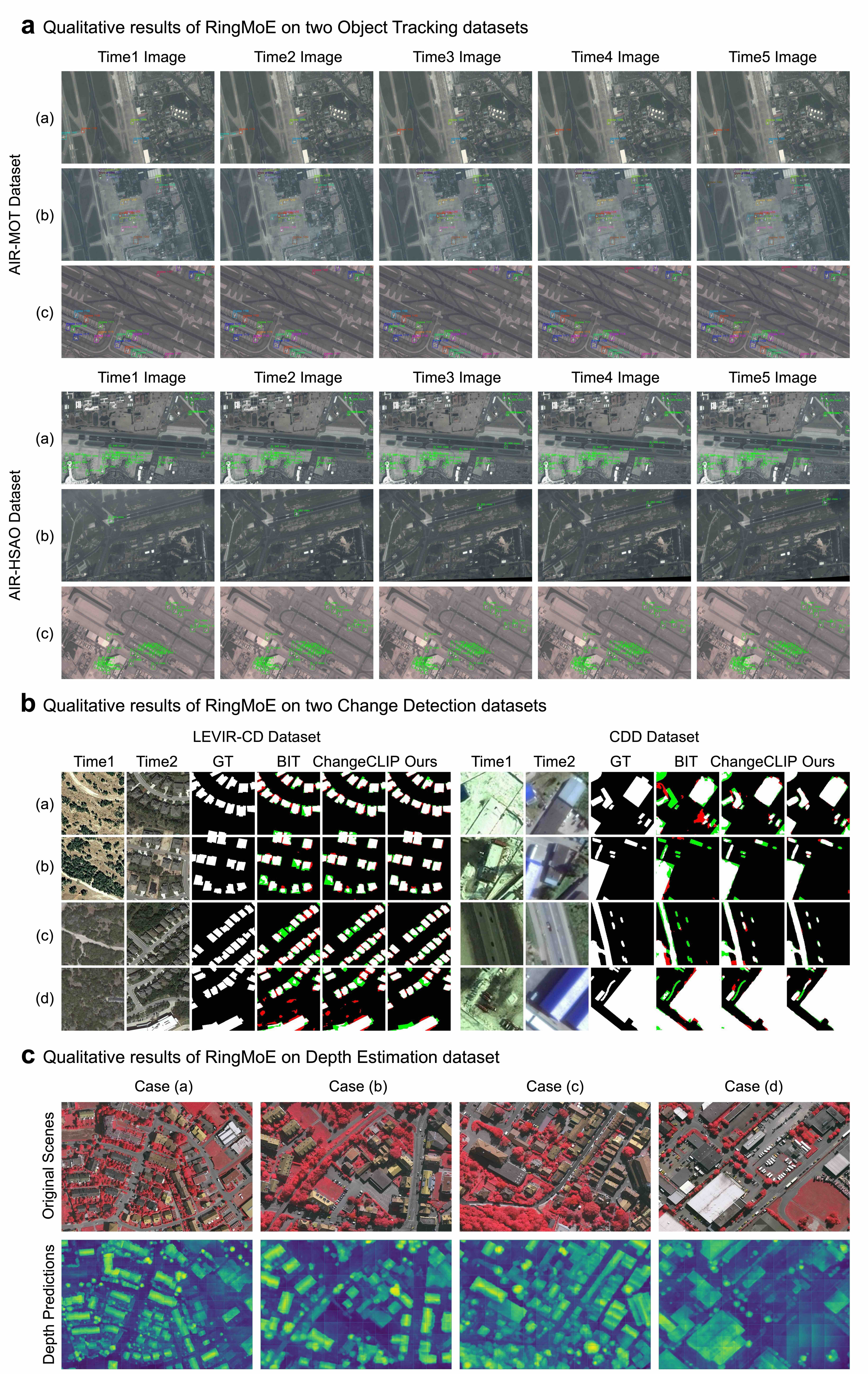}
\caption{\textbf{Qualitative results of RingMoE on object tracking, change detection, and depth estimation datasets.} For object tracking, evaluations are conducted on the AIR-MOT and AIR-HSAO datasets. Change detection performance is showcased using the LEVIR-CD and CDD datasets, while depth estimation results are presented on the Vaihingen dataset.}\label{extend_5}
\end{figure*}

\ifCLASSOPTIONcompsoc

\bibliographystyle{ieeetr}
\bibliography{refs}

\end{document}